\RequirePackage{fix-cm}
\documentclass[11pt,twocolumn]{article}
\usepackage[english]{babel}
\usepackage{url}
\usepackage{microtype}
\usepackage{authblk}
\usepackage{comment}

\usepackage{abstract}
\usepackage{amsmath,amssymb}
\usepackage{mathtools}
\usepackage{graphicx}
\usepackage{caption}
\usepackage{subfig}
\usepackage{floatrow}
\usepackage{rotating}
\newfloatcommand{capbtabbox}{table}[][\FBwidth]
\usepackage{amsthm}
\usepackage{bm}
\usepackage{epstopdf}
\usepackage{float}
\usepackage{xcolor,colortbl}
\definecolor{MyBlueUrl}{rgb}{0.12,0.35,1}
\definecolor{MyredLink}{rgb}{0.9,0.11,0.17}
\definecolor{greenTcite}{rgb}{0.25,0.65,0.38}
\usepackage{enumitem}
\usepackage{hyperref}
\hypersetup{colorlinks=true,
linkcolor=MyredLink,
citecolor=greenTcite,
urlcolor=MyBlueUrl}
\setenumerate[0]{label=\roman*)}
\setenumerate[1]{label=\arabic*.}
\setenumerate[2]{label=(\alph*)}
\usepackage[verbose=true]{geometry}
\AtBeginDocument{
  \newgeometry{
    textheight=9in,
    textwidth=7.4in,
    top=1in,
    headheight=14pt,
    headsep=25pt,
    footskip=30pt
  }
}
\newcommand{\headeright}{}
\usepackage{fancyhdr}
\fancyheadoffset{0pt}
\newcommand{\R}{\mathbb{R}}

\renewcommand{\epsilon}{\varepsilon}

\title{Enabling {\em Local} Neural Operators \\
to perform Equation-Free {\em System-Level} Analysis}
\newcommand{\shorttitle}{Enabling Local Neural Operators to perform Equation-Free System-Level Analysis}
\newcommand{\myand}{$\cdot \text{ }$}
\author[1,2]{Gianluca Fabiani}
\author[2,3]{Hannes Vandecasteele}
\author[4]{Somdatta Goswami}
\author[5,*]{Constantinos Siettos}
\author[2,3,6,*]{Ioannis G.  Kevrekidis}
\affil[1]{Hopkins Extreme Materials Institute, \emph{Johns Hopkins University}, Baltimore, 21218, MD, USA}
\affil[2]{Dept. of Chemical and Biomolecular Engineering, \emph{Johns Hopkins University}, Baltimore, 21218, MD, USA}
\affil[3]{Dept. of Applied Mathematics and Statistics, \emph{Johns Hopkins University}, Baltimore, 21218, MD, USA}
\affil[4]{Dept. of Civil and Systems Engineering, \emph{Johns Hopkins University}, Baltimore, 21218, MD, USA}
\affil[5]{Dipartimento di Matematica e Applicazioni ‘‘Renato Caccioppoli", \emph{Universit\`a degli Studi di Napoli} \emph{Federico II}, Naples, Italy}
\affil[6]{School of Medicine’s Dept. of Urology, \emph{Johns Hopkins University}, Baltimore, 21218, MD, USA}
\affil[*]{Corresponding authors, emails: 
\texttt{constantinos.siettos@unina.it},
\texttt{yannisk@jhu.edu}}

\date{\vspace{-5ex}}

\begin{document}
\twocolumn[
\maketitle
\begin{onecolabstract}
Neural Operators (NOs) provide a powerful framework for computations involving physical laws that can be modelled by (integro-) partial differential equations (PDEs), directly learning maps between infinite-dimensional function spaces that bypass both the explicit equation identification and their subsequent numerical solving.
Still, NOs have so far primarily been employed to explore the dynamical behavior as surrogates of brute-force temporal simulations/predictions. Their potential for systematic rigorous numerical system-level tasks, such as fixed-point, stability, and bifurcation analysis -- crucial for predicting irreversible transitions in real-world phenomena -- remains largely unexplored. Toward this aim, inspired by the Equation-Free multiscale framework, 
we propose and implement a framework that integrates (local) NOs with advanced iterative numerical methods in the Krylov subspace, so as to perform efficient system-level stability and bifurcation analysis of large-scale dynamical systems. 
%
This synergy expands the potential of NOs beyond simulation surrogates, enabling them to tackle fundamental challenges in computer-assisted modeling and risk assessment of complex systems.
Beyond fixed point, stability, and bifurcation analysis enabled by local {\em in time} NOs, we also demonstrate the usefulness of local {\em in space} as well as in {\em space-time} (``patch") NOs in accelerating the computer-aided analysis of spatiotemporal dynamics. 
We illustrate our framework via three nonlinear PDE benchmarks: the 1D Allen-Cahn equation, which undergoes multiple concatenated pitchfork bifurcations; the Liouville-Bratu-Gelfand PDE, which features a saddle-node tipping point; and the FitzHugh-Nagumo (FHN) model, consisting of two coupled PDEs that exhibit both Hopf and saddle-node bifurcations.

\end{onecolabstract}
\hspace{0.5cm} {\small
\textbf{keywords--} Neural Operators (NOs)
\myand Equation-Free (EF) Framework
\myand Krylov Methods
\myand Bifurcation Analysis
\myand Deep Operator Networks (DeepONets)
\myand Random Projection-based Operator Networks (RandONets)
\myand ``Gap-Tooth'' NOs
\myand Numerical Analysis
\myand Scientific Machine Learning (SciML)
}
\vskip 0.05in
\hrule height 1pt
\vskip 0.1in
]

Computer-assisted modeling of dynamical systems is fundamental to understanding, predicting, and analyzing the behavior of real-world processes across science and engineering.
Based on such models, and their system-level analysis -- which extends beyond ``brute-force'' temporal simulations -- we can enable physics-based decision-making, management, and control of the dynamics~\cite{fabiani2024task, kevrekidis2004equation, kevrekidis2003equation, karniadakis2021physics}.
Key objectives include computing long-term steady states / limit cycles (even unstable ones that are unreachable through direct temporal simulations) and constructing bifurcation diagrams to detect and understand the mechanisms that trigger qualitative dynamic transitions, including catastrophic (``hard", tipping point) bifurcations~\cite{kevrekidis2004equation,fabiani2024task, theodoropoulos2000coarse, siettos2014coarse, galaris2022numerical}.
Such computer-aided system-level analysis is crucial to tackle real-world challenges, including runaway reactions, pandemic outbreaks, climate tipping points, economic crises, ecosystem extinctions, cancer progression, and neurological disorders~\cite{fabiani2024task, gnanadesikan2024tipping, scheffer2003catastrophic, hirota2011global, dai2012generic, russo2014detecting}.

Toward this aim, the Equation-Free (EF) approach~\cite{kevrekidis2003equation, kevrekidis2004equation, erban2007variable, theodoropoulos2000coarse, rico2004coarse} was introduced in the early 2000s; its aim was to provide efficient tools for bridging local {\em microscopic/high-fidelity} models of complex systems with system-level tasks for emergent macroscopic dynamics, {\em bypassing the derivation of macroscopic, closed-form surrogate models}; the latter can be notoriously difficult to obtain~\cite{deng2025hilbert}, and may impose biases in the analysis. The main idea is to `wrap around' the local microscopic simulator, matrix-free methods in the Krylov subspace ~\cite{saad1986gmres, kelley1995iterative, arnoldi1951principle, saad2011numerical} and the numerical bifurcation analysis toolkit ~\cite{doedel2007lecture, doedel2012numerical, allgower2012numerical} for the systematic bifurcation analysis and control of the emergent dynamics. 
Since this sidesteps the derivation of the explicit macroscopic evolution equation, the approach was termed ``Equation-Free" (also motivated by matrix-free linear algebra operations). 
%

For many real-world complex systems, high-fidelity simulators are unavailable, and data may be limited to experimental measurements. Furthermore, by its construction, the EF framework provides information only on the numerical quantities required for the system-level analysis and control tasks, without offering any physical, {\em mechanistic} insight about the dynamical model that can effectively describe the emergent dynamics. 
On the other hand, the modern construction of data-driven dynamical models with the aid of scientific machine learning (SciML) has its own inherent challenges. These challenges arise because of the sparsity of available data, the ``curse of dimensionality'', nonlinearities, and stochasticity that escalate the complexity of learning robust representations. 
Traditional SciML methods, such as Deep Neural Networks (DNNs) -- in addition to struggling with the aforementioned challenges -- are also limited by the fact that they provide finite-dimensional representations that do not capture, for example, the infinite-dimensional nature of models such as PDEs. 

Neural Operators (NOs) have been introduced to address the latter problem by learning mappings between function spaces. Various architectures for NOs have been proposed over the past few years: Deep Operator Networks (DeepONets)~\cite{lu2021learning, lu2022comprehensive, goswami2022physics}, Fourier Neural Operators (FNOs)~\cite{li2020fourier}, Graph Neural Operators (GNOs)~\cite{li2020multipole, kovachki2023neural}, Wavelet Neural Operators (WNOs)~\cite{tripura2023wavelet}, Spectral Neural Operators (SNOs)~\cite{fanaskov2023spectral} and more recently Random Projection-based Operator Networks (RandONets)~\cite{fabiani2025randonets}. However, while NOs learn mappings for infinite-dimensional PDEs, they still face the challenges related to the ``curse of dimensionality''. In fact, NOs built using DNNs typically suffer from slow convergence because their optimization is NP-hard~\cite{froese2023training}, and their overparameterization makes training extremely computationally expensive~\cite{murty1987some, blum1988training, almira2021negative, adcock2021gap, fabiani2025random, karumuri2024efficient}. Hence, despite their theoretical promise~\cite{cybenko1989approximation, hornik1989multilayer, chen1995universal}, DNNs and NOs based on them often provide only partially satisfactory numerical approximation accuracy, which in practice hinder their efficiency for tasks requiring high-precision numerical analysis.
Recently, to tackle some of the computational challenges mentioned above, we introduced RandONet~\cite{fabiani2025randonets}, a {\em randomized} neural operator architecture.
RandONets share the same fundamental structure as DeepONets -- both rooted in the Chen and Chen framework~\cite{chen1995universal} -- but replace trainable hidden layers in DNNs with just one hidden layer (as in~\cite{chen1995universal}), using fixed randomized embeddings, drawing on random projections, leveraging the Johnson-Lindenstrauss lemma~\cite{johnson1984extensions} and the universal approximation power of random features~\cite{igelnik1995stochastic, huang2006extreme, rahimi2007random, rahimi2008uniform, gorban2016approximation, galaris2022numerical, fabiani2023parsimonious, fabiani2025randonets}.
This design transforms the original non-convex, gradient-based (and computationally expensive) training used in DNNs into a linear system for the output weights. Although the resulting system may be high-dimensional and ill-conditioned, it can be solved effectively using standard numerical methods such as the Moore–Penrose pseudoinverse, Tikhonov regularization~\cite{golub1999tikhonov}, QR decomposition, or complete orthogonal decomposition (COD)~\cite{hough1997complete}.

Here, for the first time, we integrate NOs with the EF framework, enabling system-level analysis based on data-driven approximations of PDE solution operators. This extends the role of NOs beyond simple simulations to construct bifurcation diagrams and detect tipping points in high-dimensional systems.
The framework is, in principle, applicable to {\em any} NO architecture; its practical effectiveness will, of course, depend on the accuracy and generalization performance of the trained local neural operator it is built around. Our key idea is to train {\em local} NOs - defined over small time intervals, and possibly in small space-time patches - and then use them as modular {\em building blocks} of computational workflows inspired by classical numerical analysis schemes.
\textcolor{black}{Our notion of locality is distinct from recent work on Neural Operators with localized kernels \cite{liu2024neural, li2024local}, where ‘local’ refers to the compactly supported kernels/filters used within the neural operator. 
In contrast, we employ local operators restricted in time or space–time, and compose them modularly into larger workflows.
Related approaches include short-time Neural Operators \cite{wang2023long,li2022learning}, which are applied iteratively to reconstruct long-time predictions. Similarly, we apply local operators iteratively, but also introduce Projective Integration (PI) \cite{gear2003projective} (and possibly, in the future, Telescopic-PI  \cite{gear2003telescopic}) to further accelerate long-time simulations.
This modularity echoes existing work on patch dynamics and space-time multiscale computation \cite{kevrekidis2003equation, samaey2006patch, runborg2002effective, xiu2005equation}, as well as recent developments on learning homogenized and effective dynamics from microscopic models \cite{arbabi2020linking, chatterjee2021robust}. Related advances include the enhancement of neural operators through finite element methods and Schwarz-type domain decomposition techniques \cite{wang2025accelerating, ouyang2025neural, jiao2021one}.
We envision that combining such ideas with local training can significantly broaden the range of NOs applicable to coarse-grained, equation-free modeling in complex multiscale systems.}
To illustrate the proposed framework, we use three representative examples: (a) the Allen-Cahn phase-field PDE with homogeneous Neumann boundary conditions~\cite{zhao2024bifurcation}, a system that undergoes consecutive pitchfork bifurcations; (b) the Liouville-Bratu-Gelfand PDE~\cite{fabiani2021numerical} which exhibits a saddle-node bifurcation; and (c) the FitzHugh-Nagumo model~\cite{theodoropoulos2000coarse, lee2020coarse, galaris2022numerical}, a pair of coupled reaction-diffusion PDEs that have Hopf and saddle-node bifurcations. We show that the proposed approach results in high-accuracy predictions in identifying critical points and constructing solution branches, even unstable ones, robustly confirming the correct stability. 
%
\begin{figure*}[ht!]
    {
    \includegraphics[trim={1.3cm 5.9cm 1.3cm 4.2cm},clip, width=0.9\textwidth]{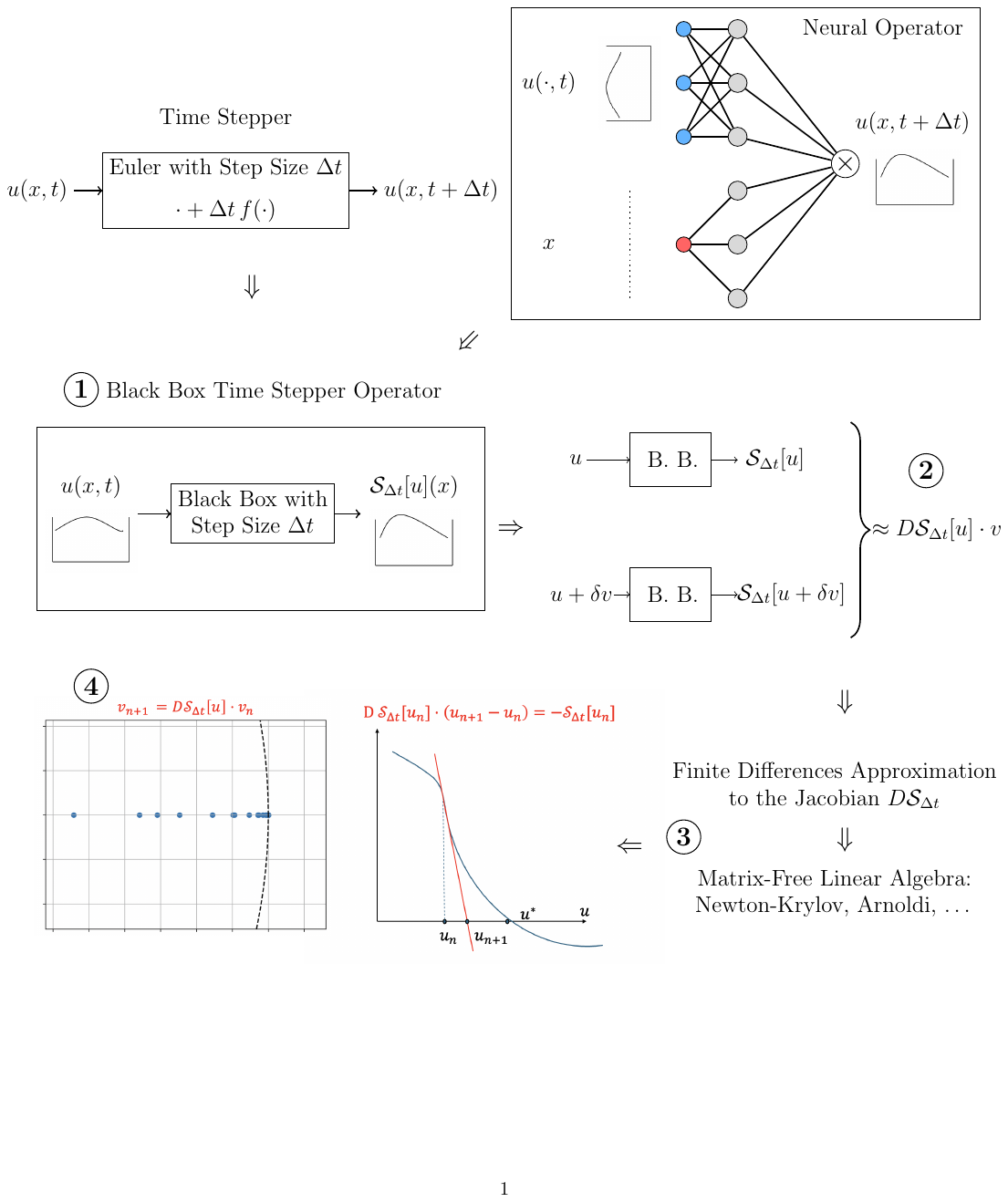}
    \label{fig:schematic_bifurcation_JFNK}
    }
    \caption{Schematic representation of Equation-Free fixed point and stability computations using data-driven (local in time) Neural Solution Operator, i.e. neural {\em timesteppers} (following the arrows) (1.) Neural operator taking the current solution $u(x,t)$ profile as input and outputting the profile at the next time step $u(x,t+\Delta t)=\mathcal{S}_{\Delta t}[u](x)$: a neural {\em timestepper}. (2.) Numerical directional derivative approximation of the {\em action} of the Jacobian; (3.) The need for matrix-free Krylov methods; (4.) Jacobian-Free Newton-Krylov (resp. Arnoldi) computations for steady-state (resp. eigenvalue) computations.}
    \label{fig:schematic}
\end{figure*}
\subsection*{Learning parametric operators of PDEs with NOs}
\label{sec:problem}
We address the challenging problem of learning, using data-driven NOs, nonlinear parametric operators $\mathcal{F}_\lambda:\mathcal{U}\times\R^p \rightarrow \mathcal{V}$ of PDEs, particularly when working with limited and sparse data and enable them to perform system-level tasks with high accuracy and low computational cost. 

These data-driven operators define mappings between infinite-dimensional function spaces $\mathcal{U}$ and $\mathcal{V}$ for each value of the parameter(s) $\lambda \in \mathbb{R}^p$. For simplicity, we assume that both $\mathcal{U}$ and $\mathcal{V}$ are subsets of ${C^1}(\R^d)$, the space of continuously differentiable functions on $\R^d$. The elements of $\mathcal{U}$ are functions $u:\mathsf{X}\subseteq \R^d\rightarrow \R$, which are transformed by the operator $\mathcal{F}_\lambda$ into functions $v=\mathcal{F}_\lambda[u]:\mathsf{Y}\subseteq \R^d\rightarrow \R$. We use the following notation for an operator evaluated at location $\bm{y} \in \mathsf{Y}\subseteq \mathbb{R}^d$: 
\begin{equation}
     v(\bm{y})=\mathcal{F}_\lambda[u](\bm{y}).
\end{equation}
Here, we initially focus on learning a data-driven approximation of the so-called {\em ``time-stepper''/solution operator} of PDEs using NOs.
The learned time-stepper provides a global solver-free surrogate of the PDE, mapping input functions (e.g., initial conditions and parameters) to the system’s state at a future time. 
Then the learned time-stepper, can be used iteratively for system-level computations (see e.g.~\cite{li2022learning, wang2023long}, and more below about Projective Integration). 
In the standard multiscale EF framework, this timestepper is computed {\em on demand}, every time its application is required, and then discarded; in our EF Local NO framework, the same trained neural timestepper is used with different initial/boundary condition inputs as necessary.

In particular, let's assume the \textit{evolution} operator of a PDE:
\begin{equation}
\frac{\partial u(\bm{x},t)}{\partial t}=\mathcal{L}[u;\lambda](\bm{x},t), \qquad \bm{x}\in\Omega, \quad t\in[0,T],
\end{equation}
where $u:{\Omega} \times [0,T] \subseteq \R^d \rightarrow \R$ is the unknown solution of the PDE and $\mathcal{L}: \mathcal{U} \times \Lambda \to \mathcal{U}$, where $\mathcal{U}$ is an appropriate functional space, such as $H^1(\Omega)$, and $\Lambda$ represents the parameter space. Then given an initial condition $u_0(\bm{x}) \equiv u(\bm{x}, t=0)$, the solution operator, $\mathcal{S}_{\Delta t}$, outputs the state profile $u(\cdot,t):{\Omega}\rightarrow \R$ after a certain time horizon $\Delta t$:
\begin{equation}
u(\bm{x},\Delta t)=\mathcal{S}_{\Delta t}[u_0;\lambda](\bm{x}), \quad \mathcal{S}_{\Delta t}: \mathcal{U} \times \Lambda \to \mathcal{U}.
\end{equation}
Note that the boundary conditions are encoded in $\mathcal{S}_{\Delta t}$ as part of the solution. 

Although the objective is to learn solution operators of PDEs as mappings between infinite-dimensional function spaces, practical implementation requires discretizing these spaces to enable effective learning via neural networks. An approach, implemented in DeepONets (and in RandONets), is to use the function values ($u(\bm{x}_j)$) at a sufficient, but finite, number of locations ${\bm{x}_1, \bm{x}_2, \dots , \bm{x}_m}$, where $\bm{x}_j \in {X}\subseteq\R^d$. These locations are often referred to as `sensors'.
Other methods to represent functions are Fourier coefficients~\cite{li2020fourier}, wavelets~\cite{tripura2023wavelet}, spectral Chebyshev basis functions~\cite{fanaskov2023spectral} and graphs~\cite{li2020multipole}. 

Regarding the availability of data for the output function $v$, there are two scenarios. In the first scenario, the value ($v(\bm{y}_i)$) of functions in the output are known at a fixed grid ${\bm{y}_1, \bm{y}_2,\dots,\bm{y}_{n}}$, where $y_i \in Y$. This case is termed as `aligned' data. In the second scenario, the output grid may be sparse and limited, vary randomly, or come from multiple experiments with different grid sizes, leading to what is known as `unaligned' data~\cite{bahmani2024resolution}. 
In both cases, the ``trunk'' network is responsible for processing the spatial locations of the output data. This is a case of particular interest, and we will focus on this by introducing ``Gap-Tooth'' NOs, a novel approach tailored to this context.

\subsection*{Equation-Free computations with NOs: fixed-points, stability and bifurcation analysis}
Performing accurate bifurcation analysis with NOs, including the reliable detection of critical points, is a particularly challenging task. A simple approach that is currently in use is the construction of bifurcation diagrams via long temporal simulations of the NOs. Toward this aim, as noted in~\cite{li2022learning, wang2023long}, it is often more efficient to learn the system's evolution, $\mathcal{S}_{\Delta t}$ over shorter time steps $\Delta t$, and then to autoregressively apply the learned time stepper via
\begin{equation*}
    u(\bm{x},T) = \mathcal{S}_T[u,\lambda]=\underbrace{\mathcal{S}_{\Delta t} \circ \mathcal{S}_{\Delta t} \circ \cdots \mathcal{S}_{\Delta t}}_{\frac{T}{\Delta t} \text{times}}[u,\lambda]
\end{equation*}
rather than train a long-time solver.
Such an autoregressive approach can suffer from successive error accumulation, specifically in the presence of high-frequency components, which NOs often do not capture accurately, as highlighted in the HINTS framework~\cite{zhang2024blending}. 

To mitigate this issue, an effective approach is to employ a fixed-point iteration method, such as Newton-Raphson,  to {\em directly find the steady-state} solutions as fixed points of the short time solution operator, bypassing the need for long-time simulations. This approach mitigates error accumulation in steady-state predictions and, by constructing appropriate initial conditions, enables the convergence to unstable steady states that are unreachable via simple temporal simulations.
A schematic overview of this Equation-Free strategy, including the Neural Operator time-stepper, finite-difference Jacobian approximation, and matrix-free Newton-Krylov and Arnoldi iterative methods, is shown in Figure~\ref{fig:schematic}.

Specifically, for a solution operator $\mathcal{S}_{T}$, a steady state $u^*$ is a fixed point
\begin{equation}
    u^* = \mathcal{S}_{T}[u^*,\lambda],
    \label{eq:NO_timestepper}
\end{equation}
and the zeros of the operator $\psi(u;\lambda)$ are defined as
\begin{equation} \label{eq:psi}
    \psi(u;\lambda) = u - \mathcal{S}_{T}[u,\lambda].
\end{equation}

One can then compute the zeros of $\psi$ for a fixed parameter value $\lambda$ by applying e.g., Newton's method: starting with a current estimation $u^{(k)}$, the next estimate is calculated by solving the linear system
\begin{equation} \label{eq:newton}
\nabla\psi(u^{(k)}; \lambda)\delta^{(k)}=- \psi(u^{(k)}; \lambda), \quad  \delta^{(k)}=u^{(k+1)} - u^{(k)}, 
\end{equation}
where $\nabla \psi$ is the Jacobian of $\psi$ with respect to $u$. 
When dealing with large-scale systems, one can resort to matrix-free iterative methods in the Krylov subspace, which explicitly avoid computing the Jacobian matrix and its inverse. In this context, here, we use the Newton-Generalized Minimal Residual (GMRES) method, but other algorithms, such as Newton-Conjugate Gradients, BiCGStab, etc. can also be applied. Briefly, the linear system in Eq.~\eqref{eq:newton} is solved by minimizing the GMRES problem:
\begin{equation}
    \delta^{(k)} = \arg\min_{\delta^{(k)} \in \mathcal{K}_m} \| \psi(u^{(k)}; \lambda) + \nabla\psi(u^{(k)}; \lambda) \delta^{(k)} \|_2,
\end{equation}
where the Krylov subspace $ \mathcal{K}_m $ is defined as:
\begin{equation}
\mathcal{K}_m = \text{span}\{r_0, \nabla\psi(u^{(k)}; \lambda) r_0, \dots, \nabla\psi(u^{(k)}; \lambda)^{m-1} r_0\};
\end{equation}
$r_0$ can be a random vector or $r_0 = -\psi(u^{(k)}; \lambda) - \nabla\psi(u^{(k)}; \lambda) \delta_{0}$, where $ \delta_{0} $ is the initial guess for GMRES. After $m$ iterations, GMRES returns $ \delta^{(k)} = \delta_{m}$.
Note that the GMRES method avoids the explicit construction and storage of the full Jacobian matrix by estimating the desired directional derivatives (the action of the Jacobian on a desired direction $r_0$, $\nabla \psi(u^{(k)}; \lambda)  r_0$), using finite differences:
\begin{equation} \label{eq:fd}
   \nabla\psi(u^{(k)}; \lambda) r_0 \approx \frac{\psi(u^{(k)}+ \epsilon r_0; \lambda) - \psi(u^{(k)}; \lambda)}{\epsilon},
\end{equation}
where $\epsilon$ is a small number, typically on the order of the square root of the machine precision or the inherent noise level in the data.

A full bifurcation diagram as a function of the parameter $\lambda$ can then be computed using pseudo-arclength continuation~\cite{allgower2012numerical}.  
In matrix-free numerical analysis, the stability of steady states can be assessed using the Arnoldi algorithm~\cite{saad2011numerical, arnoldi1951principle}, which approximates the dominant eigenpairs relevant to stability. We refer to Appendix section~\ref{App:numerical_analysis} for more details.
At this point, we note that, as highlighted in~\cite{kelley2004newton}, the discrete-time solution operator \(\mathcal{S}_{\Delta t}\) has favorable numerical properties: 
its discrete nature inherently leads to a more compact and controlled spectrum, compared to the continuous counterpart, where the eigenvalues are often confined to a region that is easier to analyze and manage. In particular, the spectrum of the Jacobian matrix of the time-stepper tends to be more tightly grouped, leading to better conditioning and more stable numerical behavior.
\begin{figure*}[ht!]
    \centering
    \includegraphics[angle=0,trim={2cm 4.5cm 2.5cm 3.5cm},clip,width=1\textwidth]{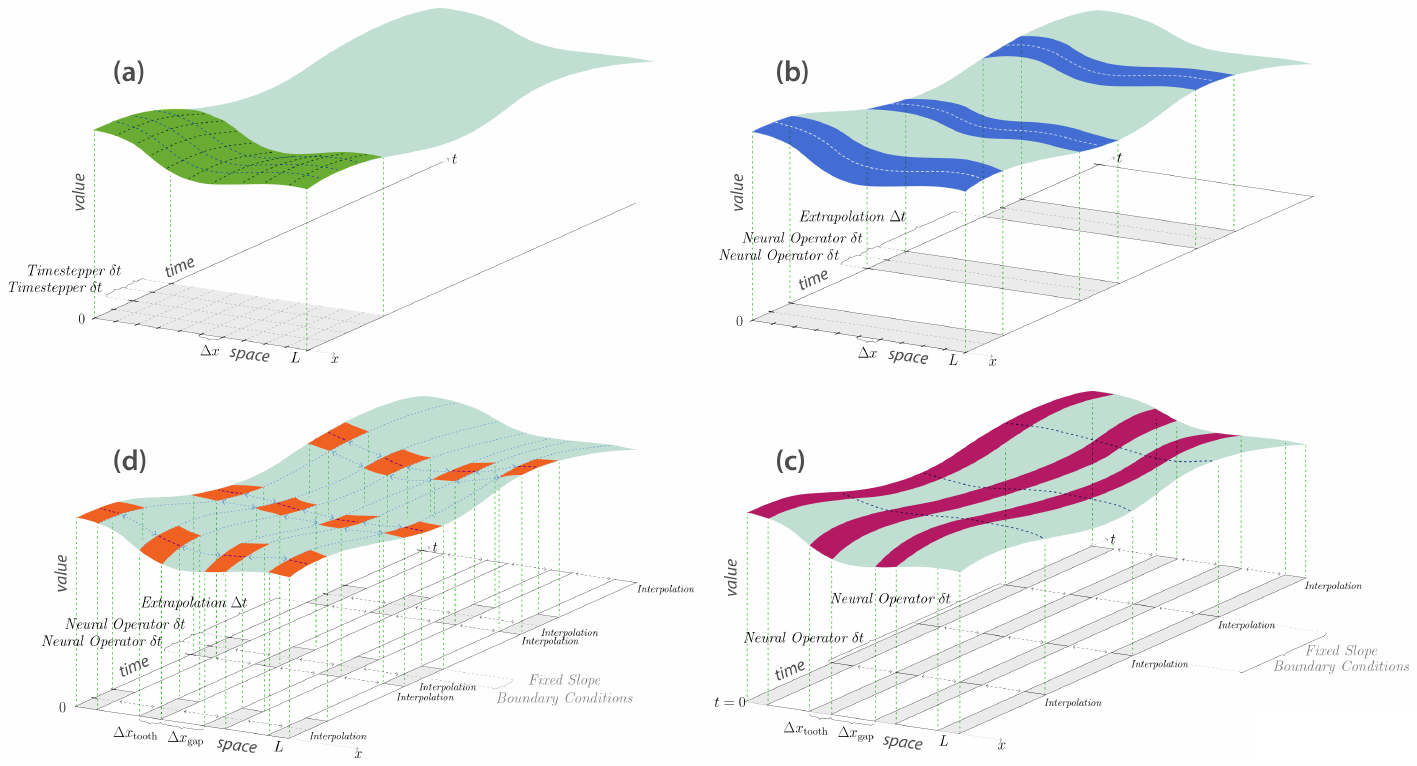}
    \label{fig:schematic_patches_a}
    \caption{Schematic representation of (a) The Euler timestepper;  (b) Projective Integration applied for NOs; (c) The Gap-Tooth scheme applied to an NO; and (d) The full patches scheme for NOs. Regions shaded in gray indicate where the timestepper or neural operator is active, white regions is where they are not. Dashed arrows indicate communication of boundary conditions through local polynomial interpolation in the Gap-Tooth and Patches schemes. \label{fig:schematic_patches}}
\end{figure*}
This property allows for improved convergence rates in linear iterations, such as those used in pseudo-arclength continuation, and can facilitate the derivation of more accurate bounds on convergence. This makes stability and bifurcation analysis in discrete-time operators more robust~\cite{kelley2004newton}.

\subsection*{From Local NOs to Global computations: Gap-Tooth NOs and Projective Integration}
Although our primary focus is on performing accurate bifurcation and stability analysis, multiscale computations can enhance the applicability of NOs for accelerating computations and learning more efficiently from sparse and limited data.

Two notable techniques in this context are (Telescopic) Projective Integration (PI) and the `Gap-Tooth` Scheme, initially developed in the multiscale EF framework~\cite{kevrekidis2003equation, gear2003projective, gear2003telescopic, gear2003gap}, that we briefly discuss here. Both exploit the smoothness of PDE solutions in time (PI) and in space (Gap-Tooth). The combination of PI and Gap-Tooth leads to the so-called \emph{patch dynamics} \cite{samaey2006patch}, for which we have also proposed \emph{patch Neural Operators} (patch NOs). A schematic representation of PI, Gap-Tooth, and their combination in patch dynamics is shown in Figure~\ref{fig:schematic_patches}.

Projective Integration (PI) is useful when the learned NO, \(\mathcal{S}_{\Delta t}\), solves over short time intervals, yet long-time predictions are desired. Examples include systems that exhibit long-term periodicity. Multiscale Projective Integration consists of alternating between \emph{short bursts} of fine-scale evolution with \emph{extrapolation} over a longer time interval. Any errors induced by extrapolation are subsequently damped by the fine-scale evolution.
In our case, PI consists of a few repeated applications of a trained short-time NO timestepper, followed by an {\em extrapolation} over a longer time interval. This approach is appropriate for problems characterized by multiple (fast-slow) time scales; the trained NO operates over fast times, and the extrapolation allows evolution over slow times.

Given the current state $u_n$, at the $n$-th projective integration step, we first apply the time-stepper (the short time NO) for $ k $ small steps,
\begin{equation}
u_{n,j} = \mathcal{S}_{\Delta t}[u_{n,j-1}], \quad j = 1, \dots, k,
\end{equation}
with $u_{n,0} = u_n$. With these states, one can approximate the slow time-evolution derivative using e.g., finite differences, and then apply a forward Euler scheme (with {\em estimated, slow} time derivative), to perform a \emph{long time step projection} using extrapolation, e.g.:
\begin{equation}
u_{n+1} = u_{n,k} + \Delta \tau \frac{u_{n,k} - u_{n,k-1}}{\Delta t},  
\end{equation}
where \(\Delta \tau \gg \Delta t\) is a larger time step. After this extrapolation, short fine-scale steps are resumed to relax back onto the slow manifold. 
%
{\em Telescopic} projective integration~\cite{gear2003telescopic} uses short-time timesteppers to systematically ``jump" over multiple time scales (multiple eigenvalue gaps in the system linearization). 

The ``Gap-Tooth'' scheme is based on the smoothness of the PDE solution {\em in space}. 
This implies that it is theoretically sufficient to learn the (short time) solution operator in {\em local regions in space} (`teeth'), with possibly large gaps in between, rather than the full domain. Within each `tooth', a local solution operator \(\mathcal{S}_{\Delta t}^{\text{local}}\) is learned:
\begin{equation}
u_{t+\Delta t} = \mathcal{S}_{\Delta t}^{\text{local}}[u_t].
\end{equation}
Importantly, the boundary conditions for these local-in-space operators are typically prescribed boundary spatial derivatives (corresponding to physical fluxes). 
Once the ``Gap-Tooth'' NO is trained for various boundary conditions, we can now combine them to evolve a {\em large spatial domain} over short times. 
After each step, the solutions are interpolated to update the boundary conditions for the next iteration, using standard numerical schemes. Repeating this process yields a {\em large space-time evolution} constructed from short-space, short-time neural operators.

In particular, we partition the domain $ \Omega $ into $ N $ small intervals (teeth) centered at $ \{x_i\}_{i=1}^N $, each of width $ \Delta x$, defined as:
\begin{equation}
    T_i = \left[ x_i - \frac{\Delta x}{2}, x_i + \frac{\Delta x}{2} \right],
\end{equation}
which are separated by $N-1$ inactive intervals (gaps) of size typically $\Delta \xi \geq \Delta x$.
At each time step $ t = 0, \Delta t, 2\Delta t, \dots, T $, we interpolate the boundary values, $u\big( x_i - \frac{\Delta x}{2}, s \big)$ and $u\big( x_i + \frac{\Delta x}{2}, s \big)$, for each tooth $ T_i $, using computed solutions in neighboring teeth $T_{i-1}$ and $T_{i+1}$, for $s \in [t, t+\Delta t] $, estimating the values of the right and left flux (boundary) $F^{right}_{i}$ and $F^{left}_{i}$. 
Then we learn (and compute) the local-tooth ($T_i$) solution operator $\mathcal{S}_{\Delta t}^{T_i}[u_s, F^{right}_{i}, F^{left}_{i}]$ within each tooth $T_i $ over $ s \in [t, t+\Delta t] $ using these interpolated boundary conditions ($F^{right}_{i}, F^{left}_{i}$), as additional branch-inputs for the Gap-tooth NOs. 
When there is no explicit spatial dependence, the learned local NOs are identical (same weights and biases) for each tooth, that is, $\mathcal{S}_{\Delta t}^{T_i}\equiv \mathcal{S}_{\Delta t}^{T_j}$, $\forall i,j=1,\dots, N$. 
This enables a ``patch'' learning process that reduces computational complexity.

Both PI and Gap-Tooth methods offer promising ways to enhance the efficiency of NO-based solvers, particularly for multiscale systems where fine-scale resolution is computationally expensive.
Most importantly, the two methods can be combined -- we can have local-in-space {\em and} local-in-time NOs; these trained NOs will be used over only part of the spatial domain (taking advantage of the solution smoothness in space-time). This is the so-called ``patch dynamics" EF scheme \cite{samaey2006patch}, which now becomes a ``patch NO" scheme for large space-time, even geometry agnostic domains. 

\section*{Results}
We present numerical results for three benchmark problems focused on identifying the solution operator of parameter-dependent PDEs: (a) the Allen-Cahn PDE; (b) a parabolic extension of the Liouville–Bratu–Gelfand PDE; (c) a Lattice-Boltzmann implementation of the FitzHugh-Nagumo (FHN) PDE. In particular, as described in the methodology, we assessed the performance of our local NOs to reconstruct the corresponding bifurcation diagrams, using Newton-GMRES and pseudo-arclength continuation to converge on steady states (stable or unstable) and track unstable solution branches past turning points. We also attempted to approximate the correct dominant eigenspectrum using a local NO-based Arnoldi method. Furthermore, we explored the use of the ``Gap-Tooth'' NOs: local {\em in space} solution operators, aiming for greater data efficiency and for increased adaptability to changing domains and boundary conditions.
For our first (the Allen-Cahn) example, we perform a direct comparison between two training strategies for NOs: the standard gradient-based approach for the fully-trained DeepONet variant, and the random projection-based RandONets variant.
While both models share architectural structure features, their different training algorithms lead to distinct generalization behavior. 
%
In our own experiments, the ``standard" DeepONet was sometimes only partially successful in reconstructing the full bifurcation diagram, occasionally missing solution branches, or producing spurious ones. The RandONet formulation was more consistently accurate, recovering both solution branches as well as their dominant eigenspectrum. For the other two case studies we present the results obtained just with RandONets (DeepONet computations were also performed, consistent with the above observations). The important issue here, in our opinion, is {\em not} to focus on contrasting the relative performance of various NO implementations in various examples; it is rather to demonstrate practically that {\em all} implementations can benefit from our Local NO framework.
\begin{figure*}[ht!]
    \subfloat[sine-type, iteration]{
\includegraphics[width=0.31\textwidth]{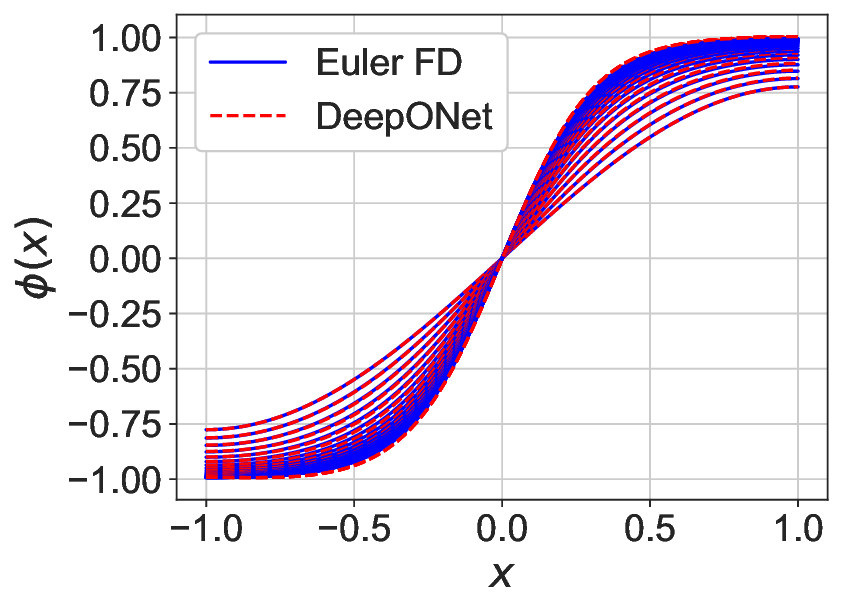}
    \label{fig:AC_DON_iter}
    }
    \subfloat[sine-type, error]{
    \includegraphics[ width=0.31\textwidth]{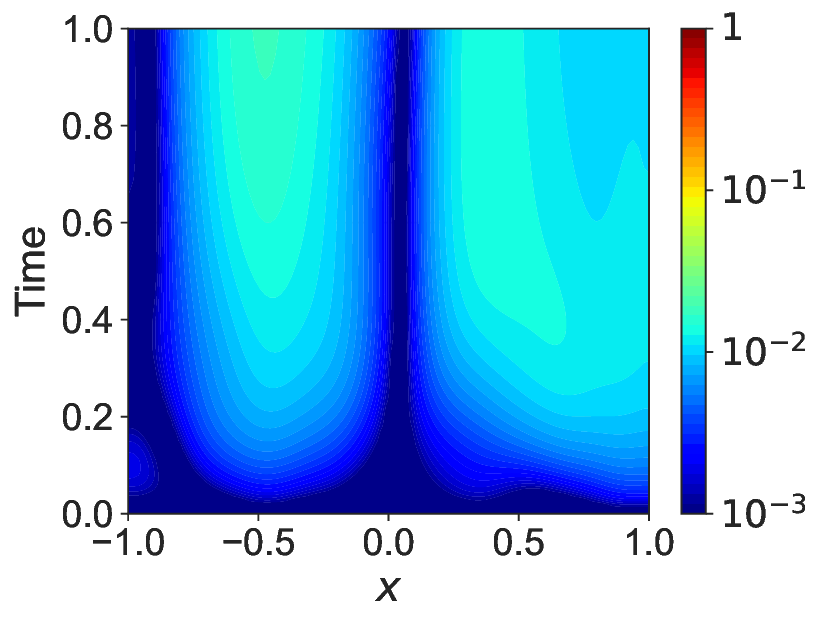}
    \label{fig:AC_DON_error}
    }
    \subfloat[sine-type, steady-state]{
    \includegraphics[ width=0.31\textwidth]{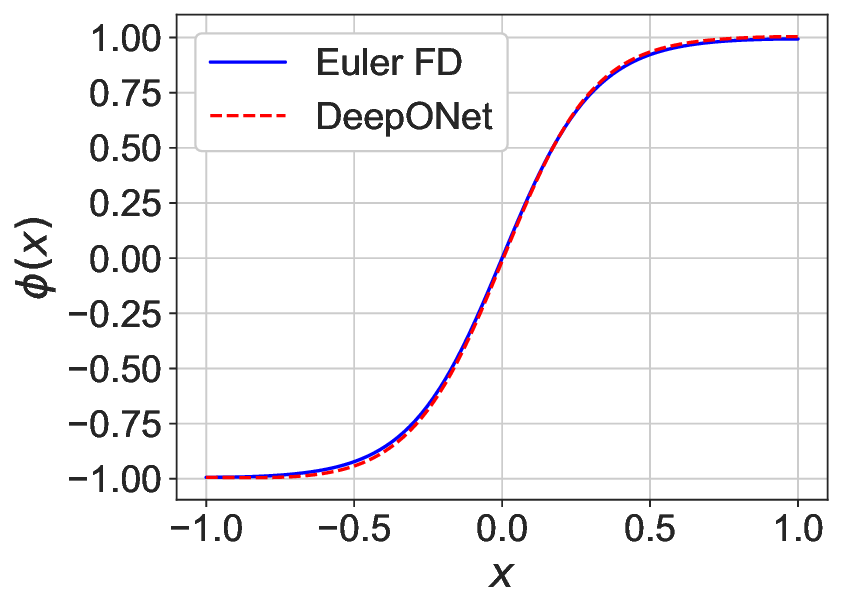}
    \label{fig:AC_DON_ss}
    }
    \\
    \subfloat[sine-type, full spectrum]{
    \includegraphics[ width=0.4\textwidth]{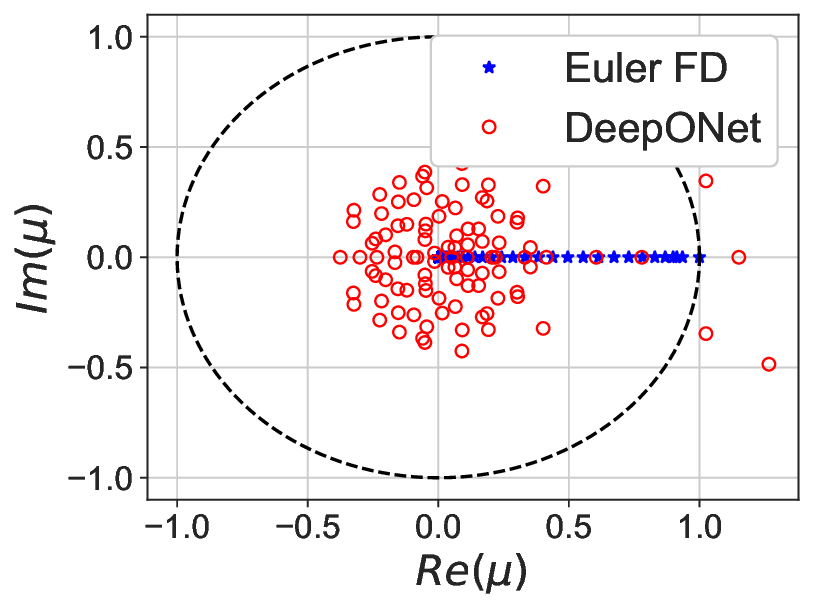}
    \label{fig:AC_DON_spectrum}
    }
    \subfloat[partial and spurious bifurcation diagram]{
    \includegraphics[ width=0.4\textwidth]{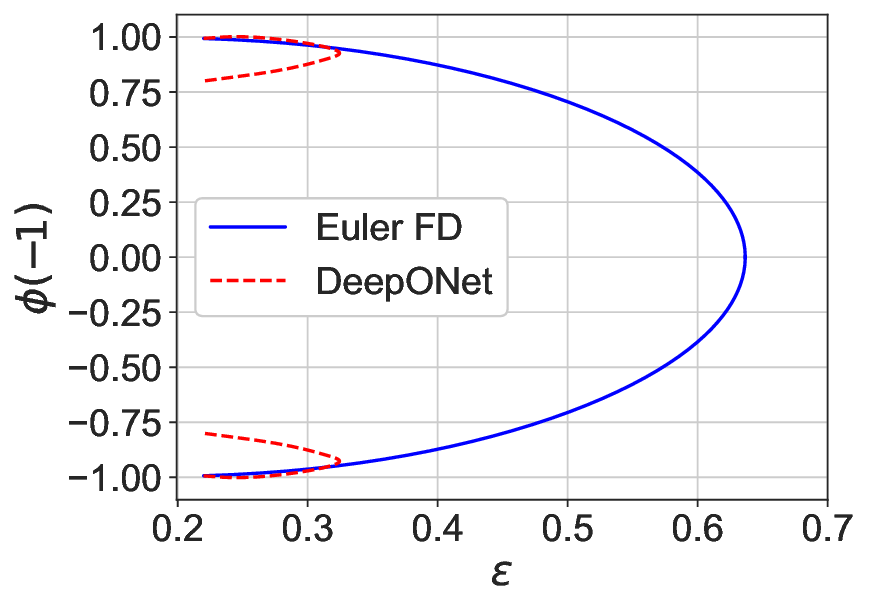}
    \label{fig:AC_DON_BD}
    }
    \caption{DeepONet-based timestepper for the Allen-Cahn PDE in Eq.~\eqref{eq:AC_PDE}, trained with data with multiple parameters in the range $\varepsilon \in [0.22, 0.7]$. The results are compared with the approximated solutions, obtained using Finite-Difference (FD) discretization in space and forward Euler in time of the \emph{known} PDE. (a) Time evolution for a sine-type initial condition for $\varepsilon=0.22$;
    (b) corresponding point-wise absolute error;
    (c) sine-type steady-state solution for $\varepsilon=0.22$;
    (d) inaccurate eigenvalue spectrum of Jacobian computed for the steady-state;
    (e) arc-length continuation of DeepONet steady-states give rise to an incomplete and spurious bifurcation diagram.
    \label{fig:AC_deepONet}}
\end{figure*}
\begin{figure*}[ht!]
    \centering
    %
    \subfloat[sine-type, steady-state]{
    \includegraphics[ width=0.31\linewidth]{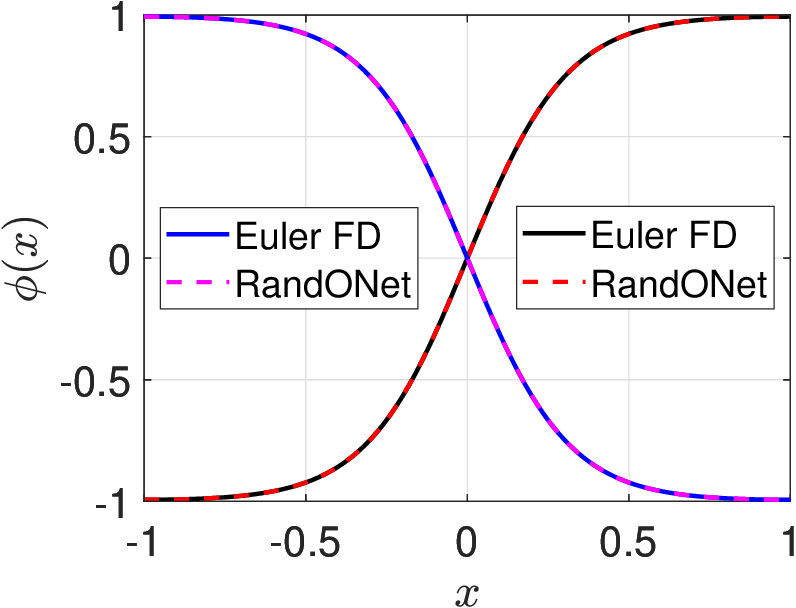}
    \label{fig:AC_steady_sin}
    }
    \subfloat[sine-type, spectrum \hspace{2cm} leading eigenvalues (Arnoldi)]{
    \includegraphics[trim={0 0 2cm 0.2cm},clip,width=0.67\textwidth]{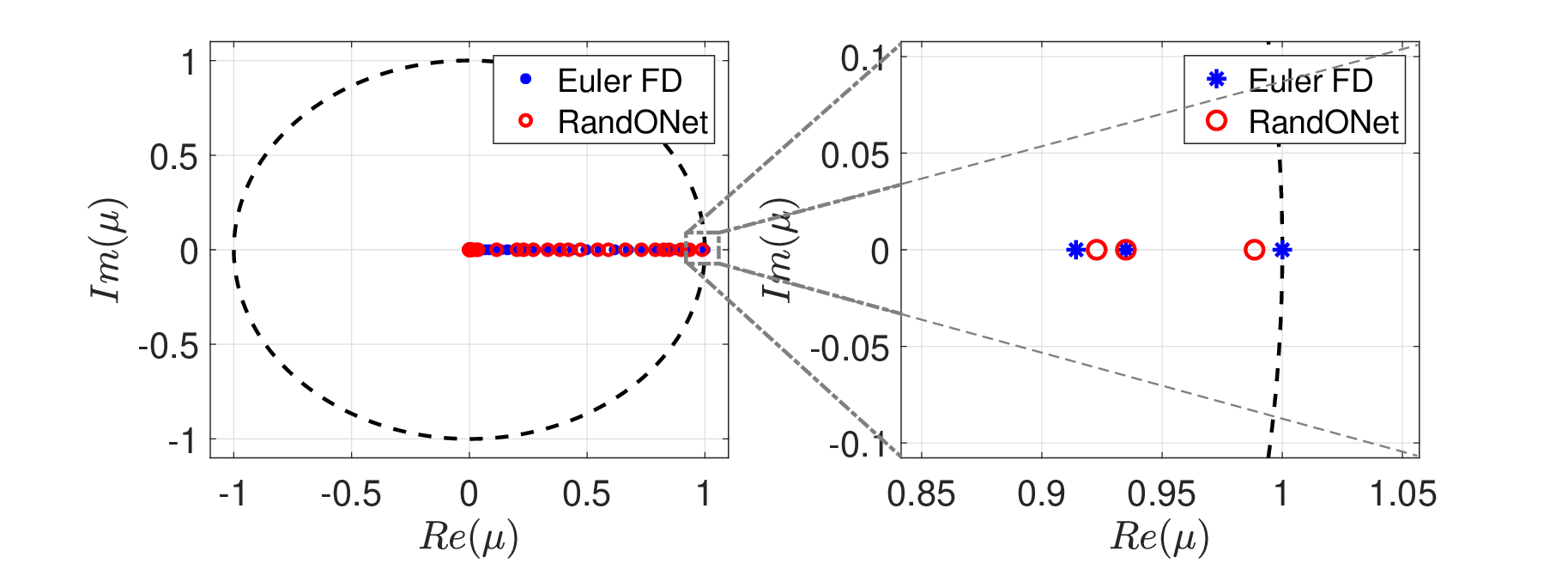}
    \label{fig:AC_spectrum_sin}
    } \\
    %
    \subfloat[cosine-type, steady-state]{
    \includegraphics[width=0.31\linewidth]{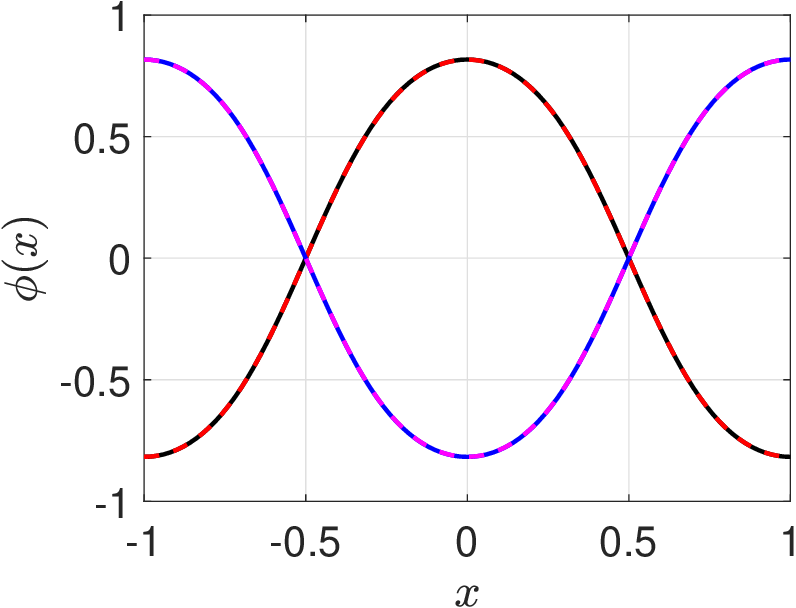}
    \label{fig:AC_steady_cos}
    }
    \subfloat[cosine-type, spectrum \hspace{2cm} leading eigenvalues (Arnoldi)]{
    \includegraphics[trim={0 0 2cm 0.2cm},clip,width=0.67\textwidth]{figures_RandONets_AC/fig_AC_single_par_RandONet_spectrum_3lead_merge_sin.eps}
    \label{fig:AC_spectrum_cos}
    }
    \caption{RandONet-based timestepper for the Allen-Cahn PDE in Eq.~\eqref{eq:AC_PDE}, trained using sub-sampled trajectories at a fixed value of parameter ($\varepsilon=0.22$) . Results are compared with the approximated solutions, obtained using Finite-Difference (FD) discretization in space and forward Euler in time of the \emph{actual} PDE. Top row: sine-type solution.
    Bottom row: cosine-type solution.
    (a) and (c): Steady state with Newton-GMRES (and symmetric counterpart), (b) and (d): Full eigenvalue spectrum at the steady-state of the full Jacobian matrix; in the zoom in of (b) and (d): The 3 leading eigenvalues of the RandONet computed with Arnoldi algorithm.}
    \label{fig:AC_results}
\end{figure*}

\subsection*{Case study 1: Allen-Cahn PDE}
Our first illustrative case is the one-dimensional Allen-Cahn PDE~\cite{zhao2024bifurcation}, a model widely used to describe phase transitions and interface dynamics. The Allen-Cahn equation is given by
\begin{equation}
\frac{\partial \phi}{\partial t}(x, t) = \epsilon \Delta \phi(x, t) - \frac{1}{\epsilon} W'(\phi(x, t)), \quad x \in \Omega, \, t > 0,
\label{eq:AC_PDE}
\end{equation}
with \(\Omega = [-1, 1]\) and \(0 < \epsilon \ll 1\) is a parameter controlling the interface width. The phase field variable \(\phi(x, t)\) represents two distinct phases (a value of $+1$ representing one phase, and $-1$ representing the other). 
The function \(W(\phi) = \frac{1}{4} \left(\phi^2 - 1\right)^2\) is a double-well potential forcing \(\phi\) to assume values close to \(\pm 1\) to clearly differentiate the phases and to make the boundaries between them narrow. The Allen-Cahn PDE arises naturally as the \(L^2\)-gradient flow of the Ginzburg-Landau free energy functional,  
\begin{equation}
E(\phi) = \int_{-1}^1 \left( \frac{\epsilon}{2} \phi_x^2 + \frac{1}{4\epsilon} \left(\phi^2 - 1\right)^2 \right) dx,
\end{equation}  
and the associated steady-state profile $\phi(x)$ is also a solution of the stationary Euler-Lagrange equation
\begin{equation} \label{eq:ac_ss}
-\epsilon \phi_{xx} + \frac{1}{\epsilon} \left(\phi^3 - \phi\right) = 0, \quad -1 < x < 1,
\end{equation}  
with Neumann boundary conditions: \(\phi_x(-1) = \phi_x(1) = 0\). 

The bifurcation diagram of the 1D Allen-Cahn PDE has been extensively studied in~\cite{zhao2024bifurcation}. In particular, it has been shown, that the bifurcations from the trivial steady state can be explained by Crandall-Rabinowitz theorem~\cite{crandall1971bifurcation}. For the completeness of the presentation, we summarize these results here. It is straightforward to verify that \(\phi(x) = -1, 0, 1\) are three trivial solutions of Eq.~\eqref{eq:ac_ss}. They always satisfy $\psi[\phi, \varepsilon] = 0$ where $\psi$ is the time-stepping map defined in Eq.~\eqref{eq:psi}. Beyond these, there are a number of bifurcation points along $\phi(x)=0$ at specific values of $\varepsilon$.
\begin{itemize}
    \item For each integer \(n \geq 0\), the parameter values \(\epsilon^{(1)}_n = \frac{1}{\pi/2 + n\pi}\) are bifurcation point. Local solutions $ (\phi_n(x, s), \epsilon_n(s)) $ around these points are well approximated by 
    \begin{equation} \label{eq:sin_bf}
    \begin{split}
    \epsilon_n(s) = \epsilon^{(1)}_n + s, \quad \quad |s| \leq 1, \\
    \phi_n(x, s) = s \sin\left(\frac{\pi}{2} + n\pi x\right) + O(s^2).
    \end{split}
    \end{equation}
    \item For each integer \(n \geq 0\), the parameter values \(\epsilon^{(2)}_n = \frac{1}{\pi/2 + n\pi}\) are bifurcation points. Local solutions $ (\phi_n(x, s), \epsilon_n(s)) $ around these points are well approximated by
\begin{equation} \label{eq:cos_bf}
\begin{split}
\epsilon_n(s) = \epsilon^{(2)}_n + s, \quad |s| \leq 1\\
\phi_n(x, s) = s \cos(n\pi x) + O(s^2);   
\end{split}
\end{equation}
\end{itemize}
$s$ is the arclength parameterization.
%
%
To generate accurate training and test datasets closely approximating the true dynamics, we solved the Allen–Cahn equation for various values of the parameter $ \varepsilon $. We used a pseudo-spectral Chebyshev method for spatial discretization and an implicit, variable-order, variable-step BDF method for time integration. For production run results, however, we compared the predictions of the trained DeepONet and RandONet models against solutions of the true equations obtained with a simpler numerical scheme -- finite differences in space and forward Euler in time. 

Initial conditions, to generate data, are sampled from a family of functions that capture the different types of transient dynamics, eventually converging to the different multiple steady-state solutions. These include sine, cosine, and flat states, randomized by cosine-modulated Gaussian profiles. The datasets are further structured to ensure symmetry. For training and short-time predictions, we used a local time step of $\Delta t = 0.01$. For longer horizon predictions, the neural operator was applied autoregressively. Full details on data generation and preprocessing are provided in Appendix section~\ref{App:AC_data_generation}.

\paragraph{Bifurcation and stability analysis using DeepONet and RandONet timesteppers.}  We evaluated the performance of {\bf local in time} RandONet and DeepONet timesteppers, both trained on the same dataset, comprising the multi-parameter case with \(\epsilon \in [0.22, 0.7]\), thus allowing for the construction of the full bifurcation diagram. 
\begin{figure*}[ht!]
   \centering
   \subfloat[bifurcation diagram]{   \includegraphics[ width=0.4\linewidth]{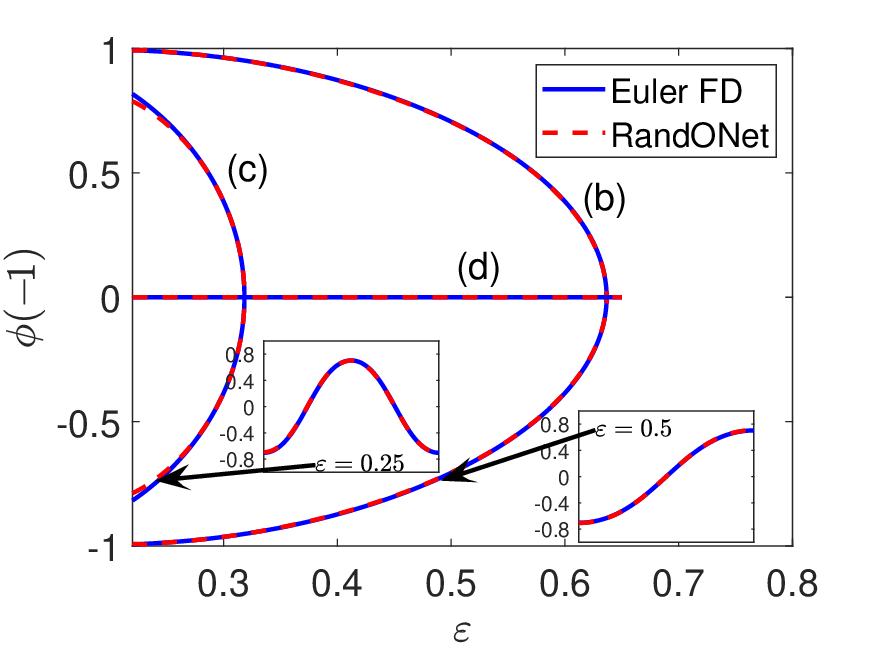}
   \label{fig:AC_BD}
   }
   \subfloat[sine-type, leading eigenvalues]{
   \includegraphics[ width=0.4\linewidth]{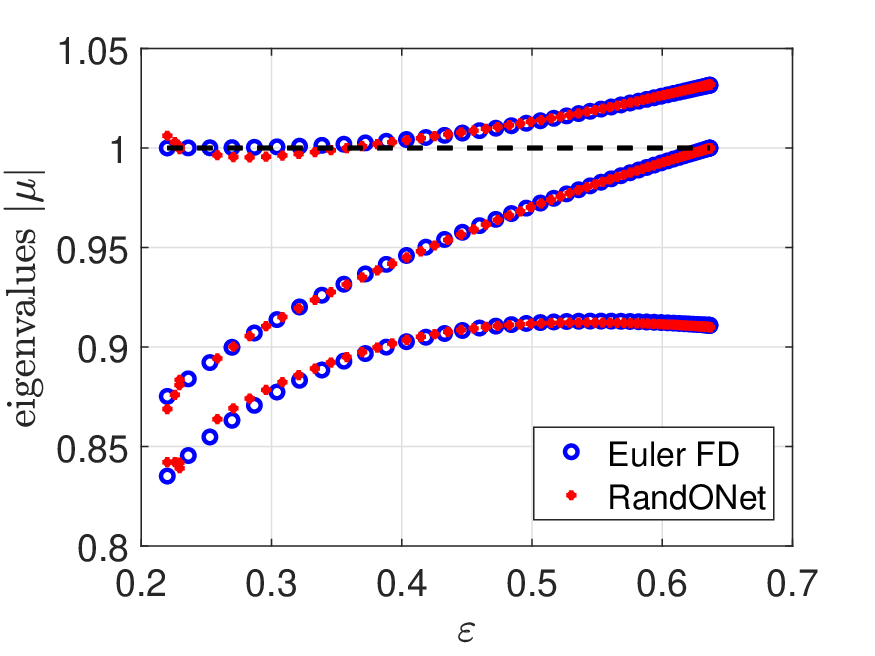}
   \label{fig:AC_BD_3lead_sin}
   }
   \\
   \subfloat[cosine-type, leading eigenvalues]{
   \includegraphics[ width=0.4\linewidth]{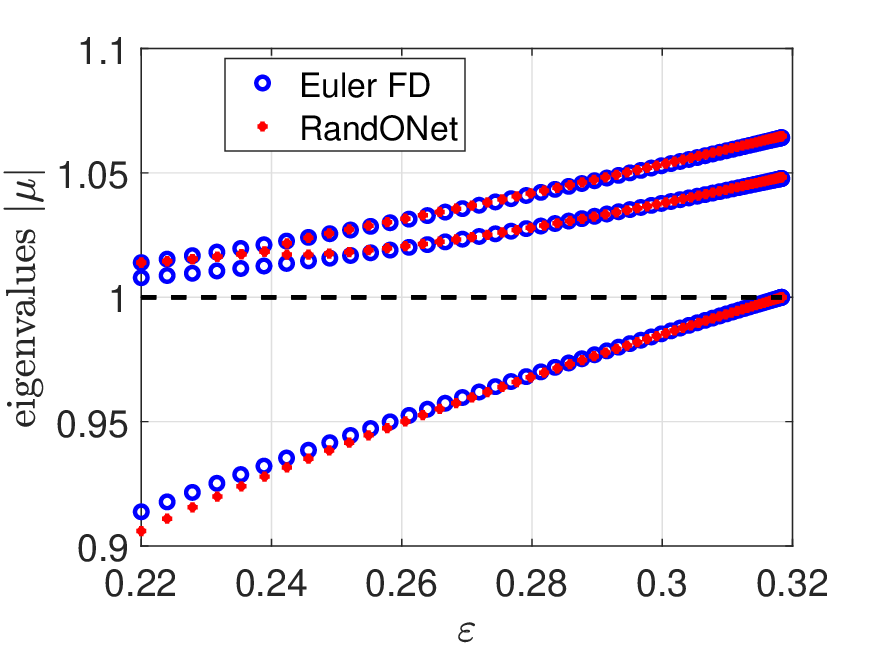}
   \label{fig:AC_BD_3lead_cos}
   }
   \subfloat[zero-type, leading eigenvalues]{
   \includegraphics[ width=0.4\linewidth]{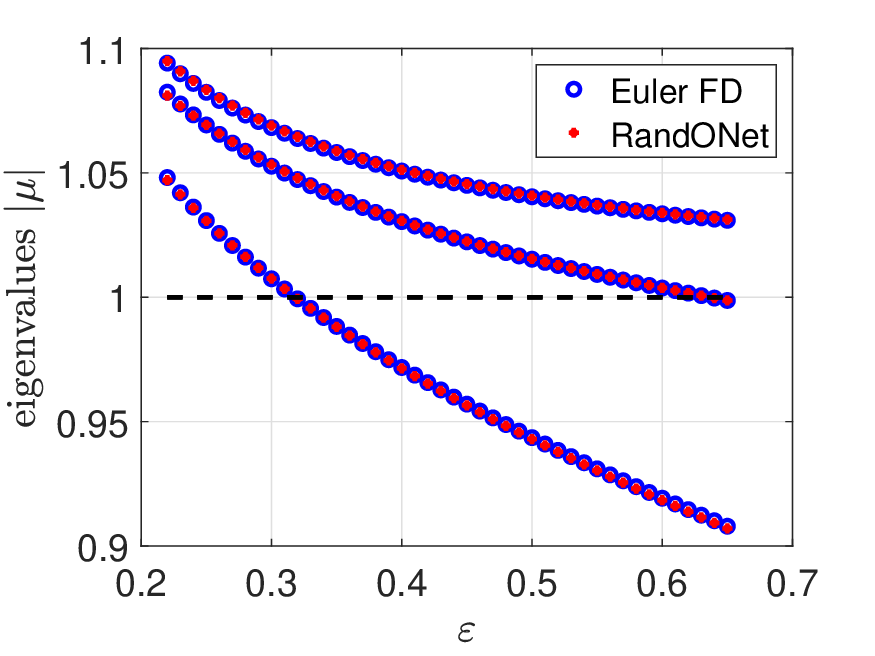}
   \label{fig:AC_BD_3lead_flat}
   }
   \caption{Bifurcation diagram obtained via the  homotopy-based RandONet  for the Allen-Cahn PDE in Eq.~\eqref{eq:AC_PDE}, trained with data in $\epsilon \in (0.22,0.7)$ using Newton-GMRES and pseudo-arclength continuation; a comparison with the reference bifurcation diagram computed based on the FD discretization of the \emph{actual} PDE is also provided. 
   (b)-(c)-(d) Tracking real part of first 3 leading eigenvalues using the Arnoldi method upon convergence. 
   (b) sine-type solution (external branch). (c) cosine-type solution (internal branch). (d) zero-type solution (middle line).}
   \label{fig:AC_BifDiag}
\end{figure*}

For the RandONet, we used random Fourier features in the branch network and sigmoidal activation functions in the trunk network. 
When learning the dynamics in the full parameter range \(\epsilon \in (0.22,0.7)\), we used a special homotopy-based RandONet with $1500$ neurons in the branch and $300$ in the trunk.
This constructs homotopy-based embeddings across the parameter space using a convex combination of multiple random bases, as detailed in the Methods.
This homotopy-based RandONet obtained an MSE of \(10^{-9}\), a median \(L^2\) error of \(10^{-5}\), and a maximum absolute error of \(10^{-3}\) for the test set.
Notably, for the training of the RandONet, only $5.22$ seconds were necessary, using a single NVIDIA GeForce RTX 2060 GPU and with a \texttt{MATLAB} implementation.
Further details on the training process and convergence analysis are provided in Appendix section~\ref{App:AC_training}.

In our DeepONet implementation, the neural operator consists of $6$ hidden layers of $100$ neurons in both the branch and trunk nets, followed by an additional fully connected layer with $300$ neurons, yielding 7 hidden layers in total for both trunk and branch nets. Thus, the training process consists of optimizing more than $200,000$ trainable parameters. The implementation is carried out in Python using the \textit{JAX} backend, ensuring compatibility with automatic differentiation and GPU acceleration.
Training is performed using the Adam optimizer with a learning rate of \(10^{-3}\), for 100,000 epochs. The training process takes approximately \(3.91 \times 10^3\) seconds (about 1 hour) on a high-performance GPU cluster (ARCH core facility at Johns Hopkins University and the Rockfish cluster). The resulting model achieves a mean squared error (MSE) of \(1.60 \times 10^{-7}\) on the test dataset.
Finally, after the training of DeepONet and RandONets, we attempted to reconstruct the full bifurcation diagram, using Newton-GMRES and pseudo-arclength continuation. Both the DeepONet and RandONet time steppers are compared against a reference solver based on finite differences in space and a forward Euler method in time. For numerical stability, the Euler solver uses a fixed time step of \( dt = 0.00005 \). In contrast, DeepONet and RandONet were trained to approximate dynamics with a larger time step of \( dt = 0.01 \), which is $200$ times larger. As a result, despite the initial training cost, DeepONet and RandONet enable significantly faster simulations.
\color{black}
Finally, to enforce the symmetry required in pitchfork bifurcation, we used an explicit transformation of the DeepONet/RandONet operator
\begin{equation}
    u(t+\Delta t)=\mathcal{S}_{\Delta t }^*[u(t),\epsilon]=\frac{\mathcal{S}_{\Delta t }[u(t),\epsilon]-\mathcal{S}_{\Delta t }[-u(t),\epsilon]}{2},
\end{equation}
which symmetrizes the solution operator. This is necessary, as perturbations break the symmetry and lead to a separation of the `fork' into saddle-node branches~\cite{evangelou2024machine}.

\paragraph{DeepONet performance evaluation.} We assessed the accuracy of the DeepONet surrogate model in approximating the time evolution, steady states, spectrum, and bifurcation diagram of the Allen–Cahn PDE. 
The results for the EF computations, using the DeepONet timestepper, are reported in Figure~\ref{fig:AC_deepONet}.
To begin with, we evaluate the DeepONet model at a representative parameter value, \( \varepsilon = 0.22 \). At this setting, DeepONet accurately reproduces the short-time dynamics (see Figures~\ref{fig:AC_DON_iter}–\ref{fig:AC_DON_error}) and identifies the sine-type steady state (see Figure~\ref{fig:AC_DON_ss}), but fails to recover the cosine-type steady state. Additionally, while the sine-type solution is captured, the model does not accurately represent the spectral properties near the steady state (see Figure~\ref{fig:AC_DON_spectrum}), with a notable discrepancy in the eigenvalue spectrum compared to the reference Euler FD method, suggesting that the linearized dynamics around the sine-type steady state are not well-learned.
Furthermore, as shown in Figure~\ref{fig:AC_DON_BD}, DeepONet provides a rather poor approximation of the full bifurcation diagram, with missing branches and resulting in spurious solutions. DeepONet follows the expected solution branch up to approximately \( \varepsilon = 0.32 \), after which it deviates along a spurious path that appears to mix characteristics of sine- and cosine-type solutions. Since the cosine-type steady state is not captured by the DeepONet surrogate, continuation in that direction fails.
This behavior indicates a partial learning of the true dynamics of the system.
This inaccurate approximation of the solution branches and stability can also be explained from matrix perturbation theory~\cite{stewart1990matrix}. In particular, the Bauer-Fike theorem states that if a perturbation (the numerical approximation error introduced by the NO), say $E$, to a diagonalizable matrix $A$ (the linearized DeepONet model around steady states) with respect to a matrix $V$ is small, then each eigenvalue $\tilde{\lambda}_i$ of the perturbed matrix $A+E$ is bounded as~\cite{stewart1990matrix}:
 \begin{equation}
 |\tilde{\lambda}_i - \lambda_i| \leq \|V\|\|V^{-1}\| \|E\|,
 \end{equation}
that is, the distance in the approximated eigenvalues varies linearly with the norm of the perturbation matrix $E$, with the linear gain being the condition number of $V$. On the other hand, the sensitivity of the eigenvectors, which reflect the solution per se near steady states, is inversely proportional to the spectral gap, between its associated eigenvalue $\lambda_i$, to all others, defined as $\min_{j \neq i} |\lambda_i - \lambda_j|$. Hence, the smaller the spectral gap, the more sensitive the eigenvectors are to perturbations (i.e., to the numerical approximation error of the Jacobian matrix)~\cite{stewart1990matrix,anderson1999lapack}.
While the DeepONet architecture employed is sufficiently expressive in principle, its training (for several tested configurations, not all shown here) proved challenging. On the other hand, we observed that training -- for this example -- RandONets,  significantly reduced training time and improved approximation accuracy in our experiments, making the process more robust and accessible to tuning. 

\paragraph{RandONet performance evaluation.}
To assess the performance of the RandONet timesteppers, we first report the results obtained when RandONets were trained at a fixed value of the parameter (for our illustrations for $\epsilon=0.22$). Then, we report the results obtained when the homotopy-based RandONets were trained to learn the dynamics in the parameter range $\epsilon \in (0,22,0.7)$ and the corresponding bifurcation diagram obtained. Figure~\ref{fig:AC_results} shows the results obtained at the fixed value of the bifurcation parameter. The top row focuses on the sine-type solutions, whereas the bottom row shows the cosine-type solutions. Figures~\ref{fig:AC_steady_sin} and~\ref{fig:AC_steady_cos} display the steady-state solutions of the respective methods obtained through Newton-GMRES iterations. In Figures~\ref{fig:AC_spectrum_sin} and~\ref{fig:AC_spectrum_cos}, we present the full eigenvalue spectrum of the steady-state Jacobian of the \emph{actual} PDE time-stepper. Finally, the zoom-in of Figures~\ref{fig:AC_spectrum_sin} and~\ref{fig:AC_spectrum_cos} shows the three leading eigenvalues calculated by the Arnoldi method. All these figures highlight a close agreement between the RandONet and the reference solver. These results demonstrate the ability of RandONets to effectively capture both steady-state solutions and critical spectral features of the Allen-Cahn equation.
Finally, to construct the bifurcation diagram, including the first two of the many pitchfork bifurcations given by equations~\eqref{eq:sin_bf} and~\eqref{eq:cos_bf}, we integrated the RandONet timestepper within the pseudo-arclength continuation algorithm coupled with Newton-GMRES. The two right-most pitchfork bifurcations computed for both timesteppers are shown in Figure~\ref{fig:AC_BD}. 
As can be seen, there is excellent visual agreement between the bifurcation diagrams of the \emph{actual} PDE and the one constructed via the RandONet.
We also computed the absolute values of the three leading eigenvalues with respect to the bifurcation parameter. The sine-type solutions are shown in Figure~\ref{fig:AC_BD_3lead_sin}, the cosine-type in Figure~\ref{fig:AC_BD_3lead_cos}, and the zero-type in Figure~\ref{fig:AC_BD_3lead_flat}. 
These results demonstrate that RandONets also accurately approximate the actual key eigenvalues, i.e., those governing the dominant dynamics and bifurcation behavior, thus accurately estimating the correct stability of the associated solution branches. 

\subsection*{Case study 2: The parabolic Liouville–Bratu–Gelfand PDE}
Our second illustrative example is the parabolic Liouville–Bratu–Gelfand PDE~\cite{boyd1986analytical}, that arises in modeling several physical and chemical systems, given by:
\begin{figure*}[ht!]
    \centering
    %
    \subfloat[steady-state]{
    \includegraphics[ width=0.31\linewidth]{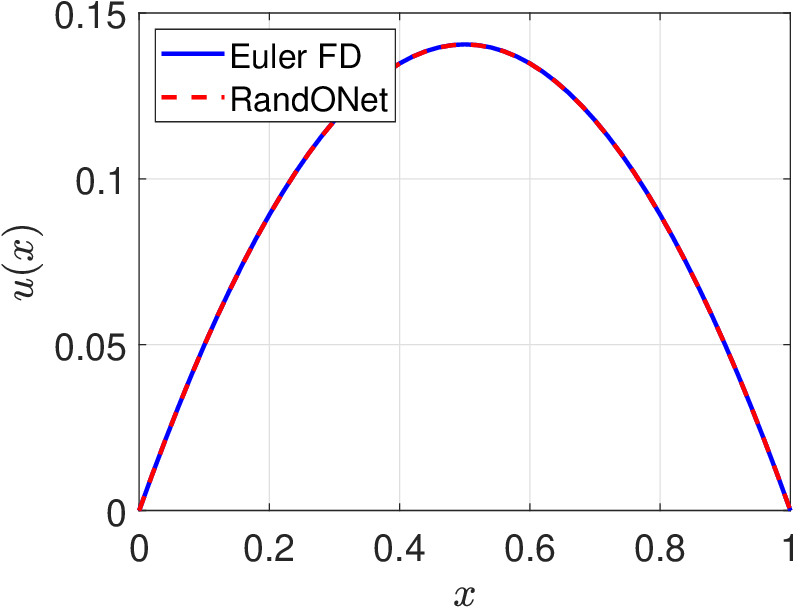}
    \label{fig:bratu_steady_NKGMRES}}
    \subfloat[full spectrum \hspace{2.5cm} leading eigenvalues (Arnoldi)]{
    \includegraphics[trim={0 0 2cm 0.2cm},clip,width=0.67\textwidth]{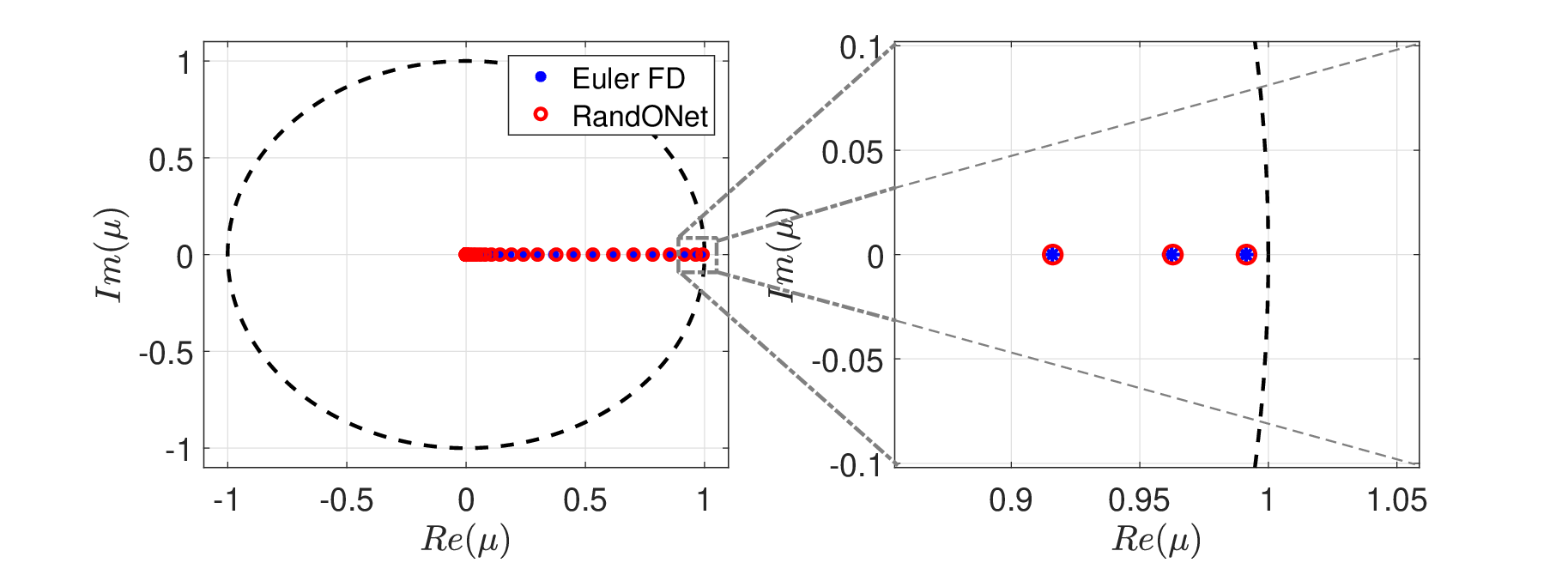}
    \label{fig:bratu_full_spectrum}
    }
    \\
    \subfloat[]{
   \includegraphics[ width=0.4\linewidth]{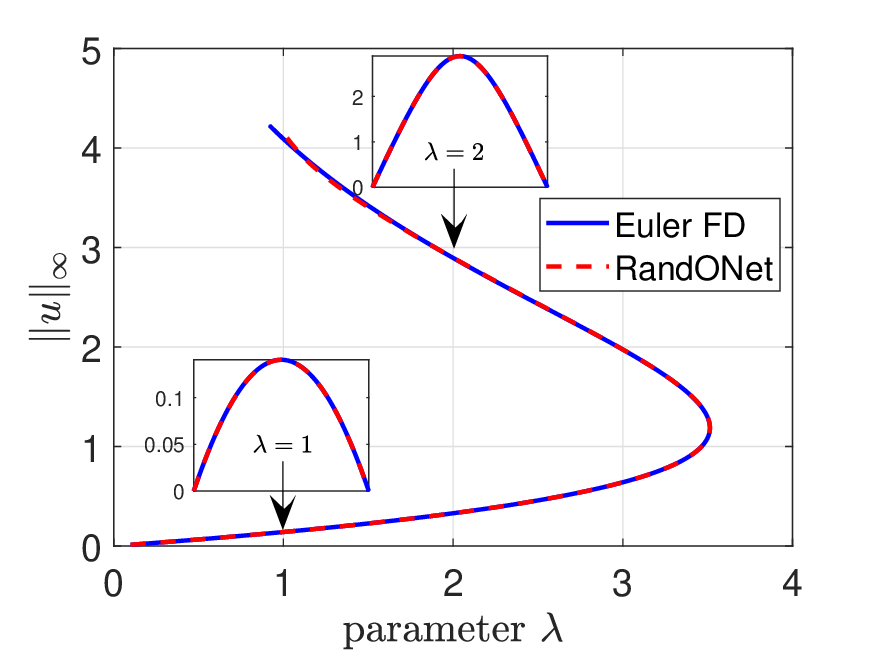}
   \label{fig:bratu_BD}
   }
   \subfloat[]{
   \includegraphics[width=0.4\linewidth]{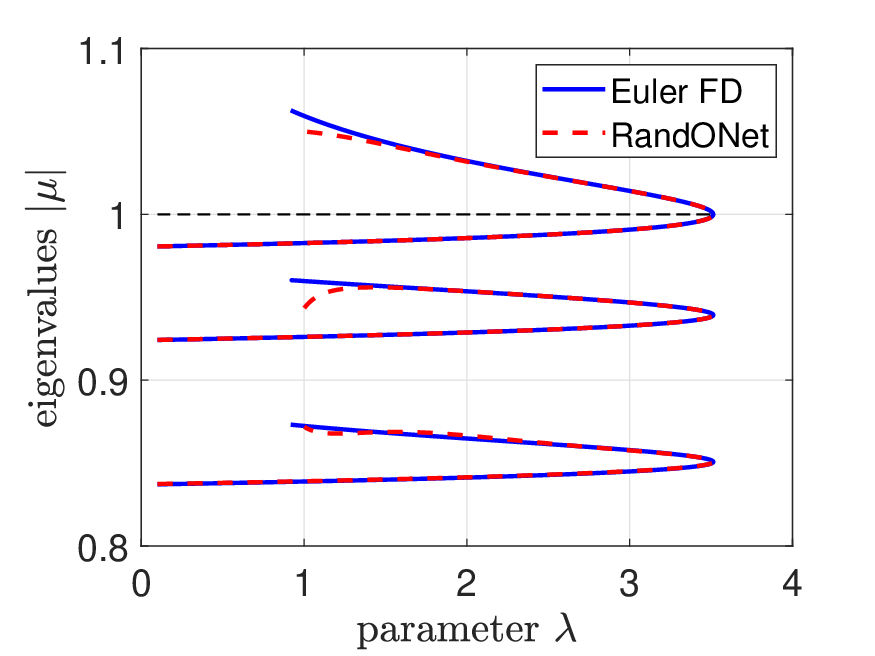}
   \label{fig:bratu_BD_3leading}
   }
    \caption{RandONet for the parabolic Liouville–Bratu–Gelfand PDE in Eq.~\eqref{eq:Bratu_parabolic}, trained using trajectories at a fixed value of the bifurcation parameter ($\lambda=1$) (in (a)-(b)) and trained using the full dataset of the dynamics in \(\lambda \in [0, 3.8]\) (in (c)-(d)). 
    Results are compared with a Finite Difference (FD) based time-stepper.
    (a) Steady-state solutions obtained with Newton-GMRES; (b) Full spectrum of the steady-state Jacobian of the \emph{actual} PDE time-stepper; in the blow up panel, the three leading eigenvalues as computed using the Arnoldi method.
    (c) Reconstructed bifurcation diagram with the RandONets, compared with the true FD bifurcation diagram. In the two insets, we show two representative solutions associated with $\lambda=1$ (stable branch) and $\lambda=2$ (unstable branch).
   (d) Tracking the absolute value of first 3 leading eigenvalues.}
    \label{fig:bratu_single_par}
\end{figure*}
   

\begin{equation}
    \frac{\partial u}{\partial t}=  \frac{ \partial^2 u}{\partial x^2}+\lambda \exp(u), \qquad x\in \Omega=[0,1],
    \label{eq:Bratu_parabolic}
\end{equation}
with homogeneous Dirichlet boundary conditions
$u(0,t) = u(1,t) =0$.
The one-dimensional steady state problem admits an analytical solution~\cite{mohsen2014simple}, reading
\begin{equation}
\begin{split}
    u(x)=2\ln\frac{\cosh{\theta}}{\cosh{\theta (1-2x)}},\\
    \text{such that } \cosh{\theta} = \frac{4\theta}{\sqrt{2 \lambda}}.
\end{split}
\label{eq:sys1}
\end{equation}
It can be shown that when $0<\lambda <\lambda_c$, the problem admits two solution branches that meet in the saddle-node point $\lambda_c \sim 3.513830719$ where the stability changes, while beyond $\lambda_c$ no solutions exist~\cite{boyd1986analytical}.

To generate accurate training and test dataset closely approximating the true dynamics, we solved the parabolic Liouville-Bratu-Gelfand PDE for various values of the parameter \(\lambda \in [0, 3.8]\). We used a pseudo-spectral Chebyshev method for spatial discretization and an implicit, variable-order, variable-step BDF method for time integration~\cite{platte2010chebfun,driscoll2014chebfun, shampine1997matlab}.

As above, we compared the trained RandONet timestepper models against approximate solutions obtained with a simpler
numerical scheme – finite differences in space and forward Euler in time.
The initial conditions to generate data are sampled from a parametrized family of functions that satisfy the Dirichlet boundary conditions. Full details of the discretization, solver, and dataset structure are provided in  Appendix section~\ref{App:Bratu_data_generation}.

For numerical stability, the Euler solver uses a fixed
time step of $dt = 0.0001$. In contrast,
RandONets are trained to approximate dynamics with a ten times 
larger time step of $dt = 0.001$.
As a result, despite the initial training cost, RandONet enabled slightly faster simulations. More details on the convergence results are provided in Appendix section~\ref{App:Bratu_training}.
\paragraph{Bifurcation and stability analysis}
We first evaluated the accuracy of RandONets in approximating the time evolution, steady states, and spectral properties of the Liouville–Bratu–Gelfand PDE at a fixed value of the bifurcation parameter (here, at \(\lambda = 1\)). For the fixed parameter case, we used a RandONet with $200$ neurons in the branch and $150$ neurons in the trunk.  In the branch network, we used cosine activation functions combined with Gaussian-distributed weights. This forms a Random Fourier Feature Network (RFFN) representation of the function space, following the methodology of~\cite{rahimi2007random,fabiani2025randonets}. Full details of the training setup are provided in Appendix section~\ref{App:Bratu_training}. 
We report the results in Figure~\ref{fig:bratu_single_par}. 
We compared the steady-state solutions obtained using Newton-GMRES on both the RandONet timestepper and the reference time-stepper. These results are shown in Figure~\ref{fig:bratu_steady_NKGMRES}. The steady-state profile computed by the RandONet aligns accurately with the one obtained by the reference FD solver. This result confirms that the learned model reliably captures the long-term behavior.
We further assess the stability analysis efficiency of the RandONet. Figure~\ref{fig:bratu_full_spectrum} shows the full eigenvalue spectrum obtained with the RandONet and the reference time-stepper at steady states, showing a nearly-exact match up to an accuracy of \(10^{-4}\). These eigenvalues of the full Jacobian matrix were computed using a direct differentiation approach. Furthermore, as shown, the three leading eigenvalues computed using the Arnoldi method also exhibit an accuracy of \(10^{-4}\).  These results confirm that a RandONet not only accurately replicates short-term dynamics but also provides a robust approximation of the system’s fundamental characteristics such as steady-state computations and spectral properties.

Finally, we trained a homotopy-based RandONet to learn the dynamics in the full range of the bifurcation parameter in \(\lambda \in [0, 3.8]\). For that purpose, we used $1500$ neurons in the branch and $150$ neurons in the trunk.
In the trunk network, which processes low-dimensional spatial coordinates, we applied a sigmoidal transformation with randomized weights. These weights are sampled following the approach described in~\cite{fabiani2025randonets, fabiani2021numerical, fabiani2023parsimonious, fabiani2025random}. Figure~\ref{fig:bratu_BD} shows the bifurcation diagram obtained wrapping around the RandONet, Newton-GMRES with pseudo-arclength continuation. As shown, the bifurcation diagram obtained with a RandONet is, for any practical means, identical up to the tolerance of $10^{-5}$, to the reference one, accurately capturing both stable and unstable branches.  The saddle-node bifurcation was identified at $\lambda_c=3.51381$, a value that corresponds to the theoretical one with an error of the order $10^{-5}$. Figure~\ref{fig:bratu_BD_3leading} shows the approximation of the real part of the three leading eigenvalues as a function of $ \lambda$ as obtained by RandONets by applying the Arnoldi method. As shown, RandONets approximate the actual eigenvalues and stability of the solution with high accuracy.

\paragraph{Projective integration for accelerating time simulations}
Here, we illustrate how we can further accelerate time simulations by applying projective integration (PI) using the RandONet time-stepper with a small time step (\(dt = 0.001\)). Projective integration (for problems with time-scale separation) alternates between short relaxation steps and large projective steps to advance efficiently along the slow time-evolution manifold.
\begin{figure*}[ht!]
    \centering
    \subfloat[Projective-integration]{
    \includegraphics[trim={0 0 0cm 0.2cm},clip,width=0.4\textwidth]{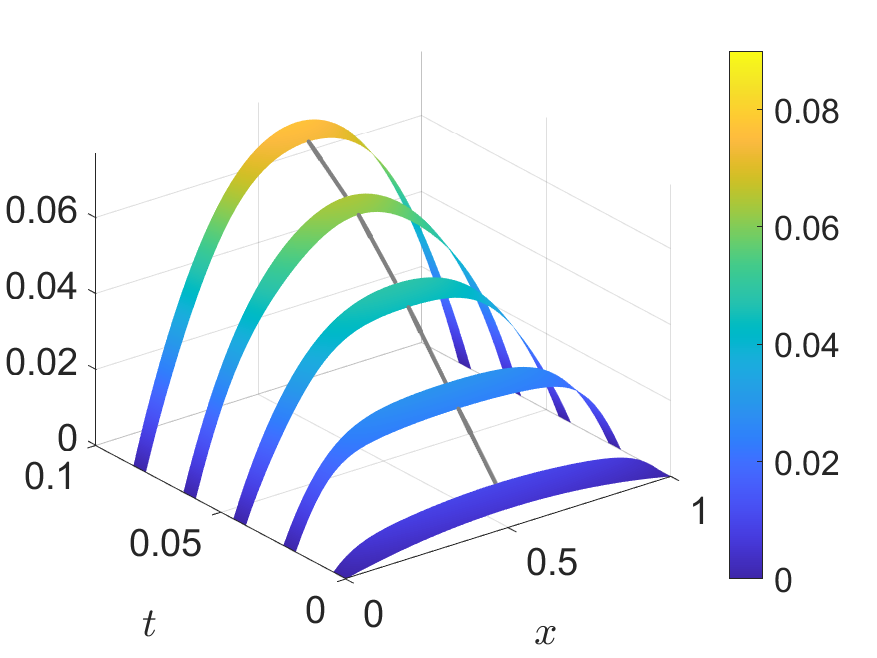}
    \label{fig:CPI_solution}
    }
    \subfloat[error]{
    \includegraphics[width=0.4\textwidth]{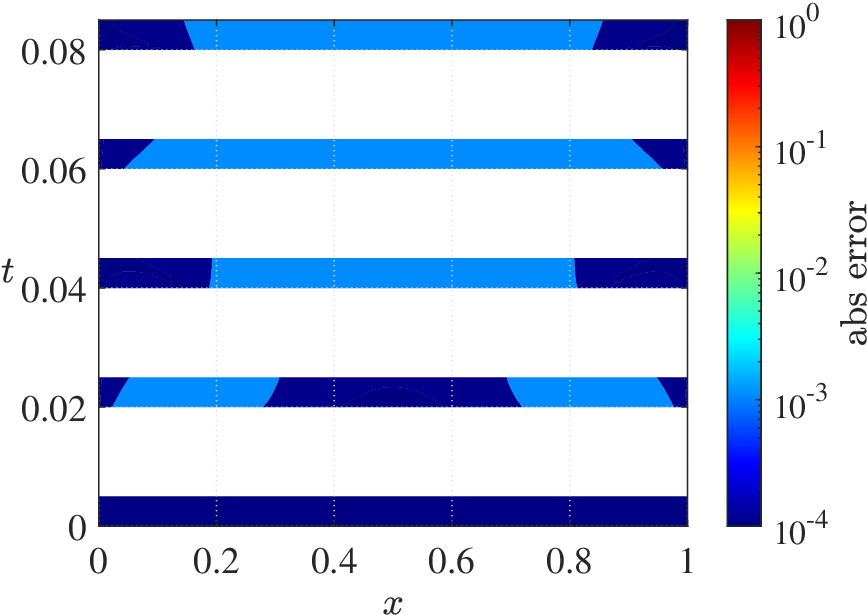}
    \label{fig:cpi_error}
    }\\
    \subfloat[trajectory of $\| u\|_{\infty}$]{
    \includegraphics[ width=0.31\textwidth]{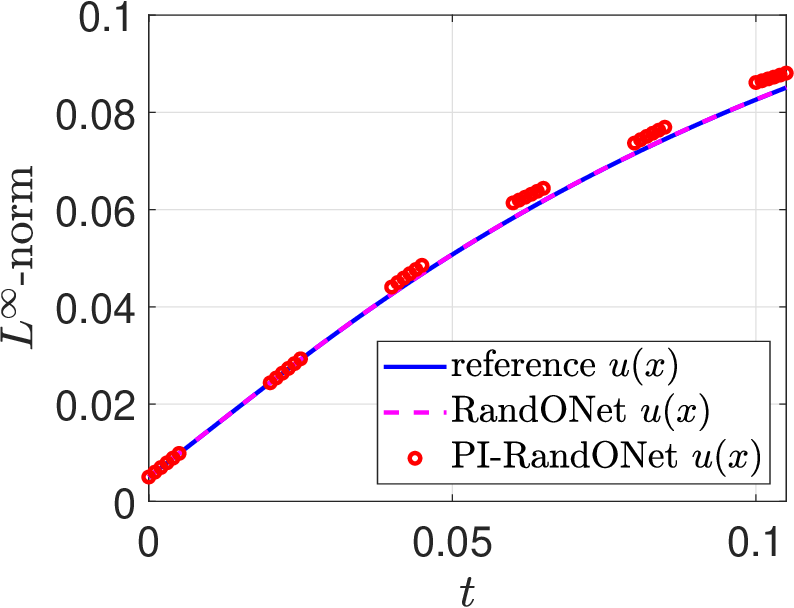}
    \label{fig:cpi_linf}
    }
    \subfloat[trajectory of $\| u_x\|_{\infty}$]{
    \includegraphics[ width=0.31\textwidth]{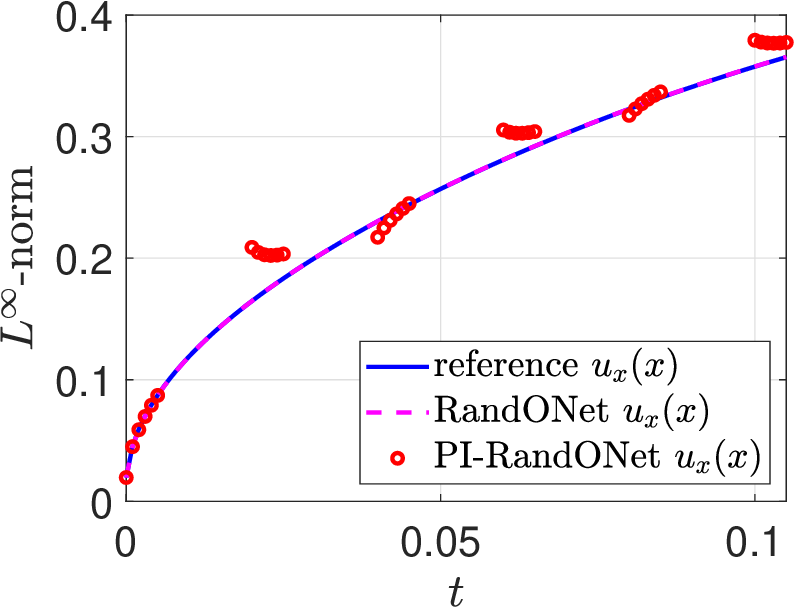}
    \label{fig:cpi_variation}
    }
    \subfloat[trajectory of $\| u_{xx}\|_{\infty}$]{
    \includegraphics[width=0.31\textwidth]{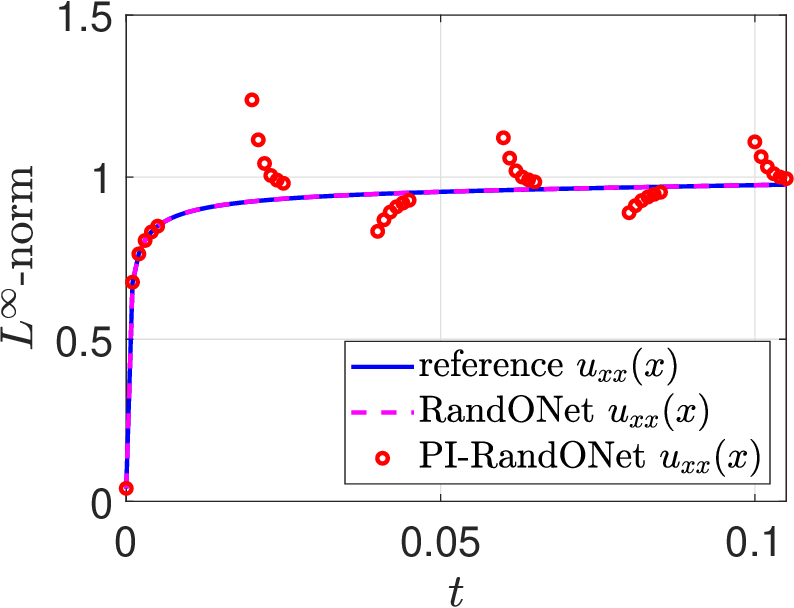}
    \label{fig:cpi_curvature}
    }
    \caption{Projective Integration (PI) of the RandONet time-stepper for the Liouville-Bratu-Gelfand PDE in Eq.~\eqref{eq:Bratu_parabolic}.
    (a) PI solution trajectory. (b) Space-time error compared to the reference solution.  
    (c) Evolution of the \(L^\infty\)-error in the solution over time.  
    (d) Evolution of the total variation (maximum error in the first spatial derivative).  
    (e) Evolution of the total curvature (maximum error in the second spatial derivative), highlighting relaxation to the slow manifold. \label{fig:CPI_results}}
\end{figure*}

Specifically, we performed five short relaxation steps with \(\Delta t = 0.001\) (total relaxation time \(0.005\)), followed by a large projective step of size \(\Delta T = 0.015\) using a backward finite difference estimation of the solution time derivative. Figure~\ref{fig:CPI_results} summarizes the performance of the RandONet forward Euler projective integrator. Figure~\ref{fig:CPI_solution} shows the evolution of the PI solution $u(x,t)$, and Figure~\ref{fig:cpi_error} displays the related space-time error. This error remains confined within \(10^{-3}\), demonstrating excellent agreement with the reference solver. Figures~\ref{fig:cpi_linf}-\ref{fig:cpi_curvature} track the evolution of the \(L^\infty\)-error in the solution and its first and second spatial derivatives over time. The first derivative error, representing total variation, and the second derivative error, representing total curvature, highlight the relaxation process. Notably, as shown in Figure~\ref{fig:cpi_curvature} there is a rapid decay in total curvature, confirming the fast convergence to the slow manifold. This behavior reflects the typical separation of timescales in dissipative PDEs, where higher spatial derivatives correspond to higher frequency modes, which tend to relax more rapidly than the field component $u$.

These results demonstrate the effectiveness of projective integration in reducing computational cost while maintaining accuracy. This effectiveness extends to RandONets as well, further proving that they are efficient surrogates for equation-free multiscale computations.

\begin{figure*}[ht!]
    \centering
    \subfloat[Gap-Tooth NO timestepper]{
    \includegraphics[ width=0.31\linewidth]{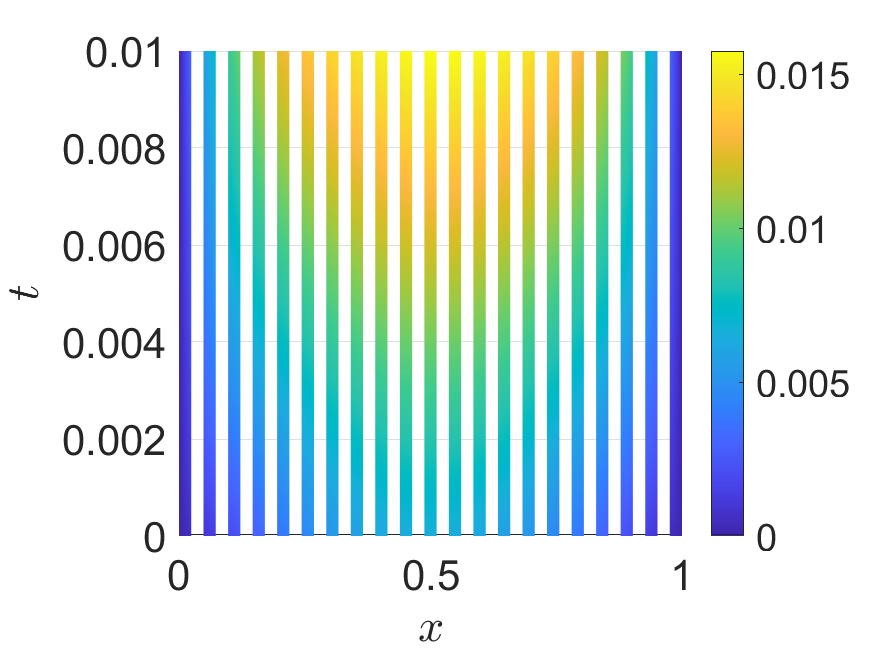}
     \label{fig:gap_tooth_solution}
    }
    \subfloat[Gap-Tooth NO error]{
    \includegraphics[ width=0.31\linewidth]{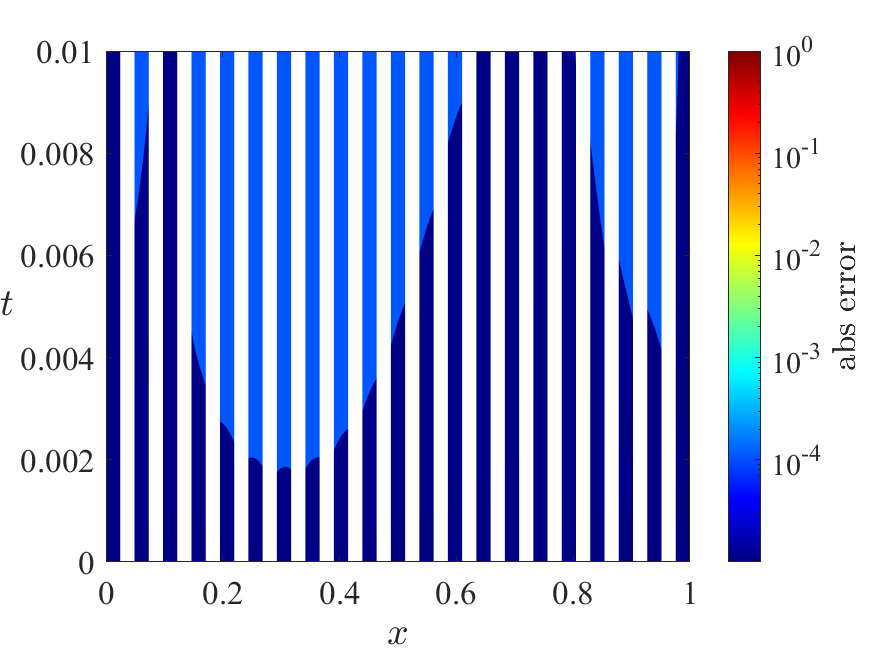}
     \label{fig:gap_tooth_error}
    }
    \subfloat[steady-state]{
    \includegraphics[width=0.31\linewidth]{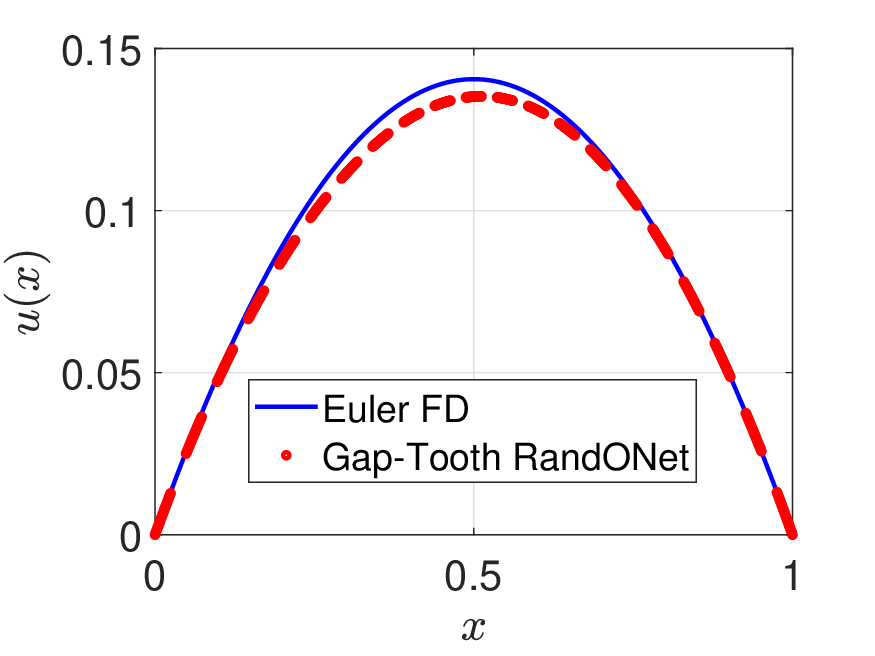}
    \label{fig:gap_tooth_steady}
    }\\
    \subfloat[Patch NO timestepper]{
    \includegraphics[trim={0 0 0cm 0.2cm},clip,width=0.4\linewidth]{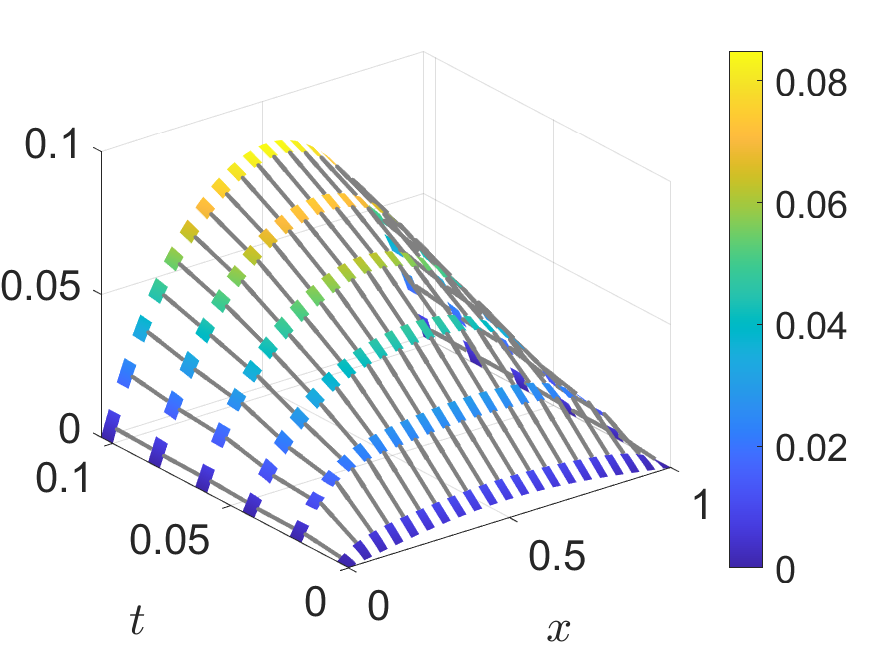}
     \label{fig:patches_solution}
    }
    \subfloat[Patch NO error]{
    \includegraphics[width=0.36\linewidth]{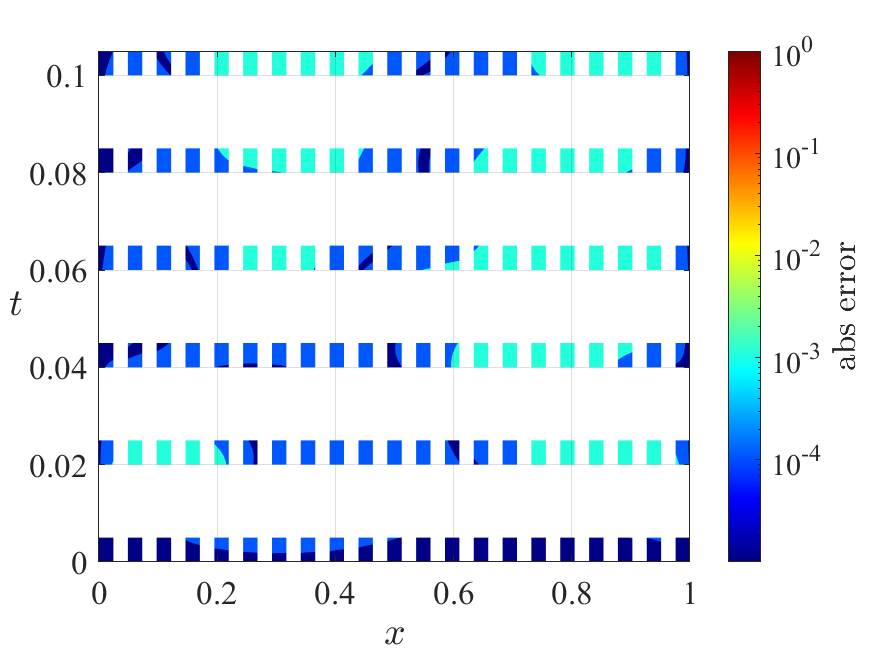}
     \label{fig:patches_error}
    }
    \\
    \subfloat[full spectrum \hspace{3cm} leading eigenvalues (Arnoldi)]{
    \includegraphics[trim={0 0 2cm 0.2cm},clip,width=0.85\textwidth]{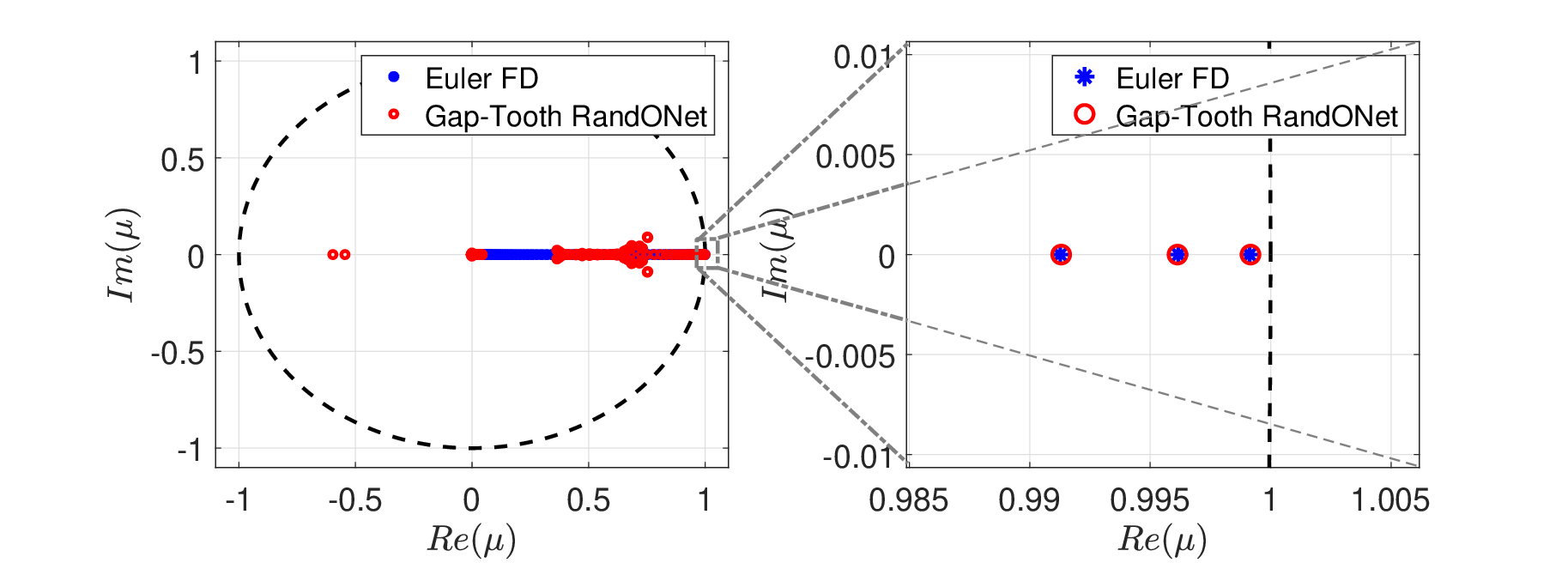}
    \label{fig:gap_tooth_full_spectrum}
    }
    \caption{Gap-Tooth RandONet timestepper results for the Liouville-Bratu-Gelfand PDE in Eq.~\eqref{eq:Bratu_parabolic}, with $\lambda=1$. (a) the Gap-Tooth RandONet timestepper; (b) Corresponding spatio-temporal error w.r.t the reference solution;
    (c) Patch RandONet timestepper (Combination of Gap-Tooth NO and PI); (d) correspponding spatio-temporal error w.r.t the reference solution; (e) Steady states of both time steppers computed with Jacobian-Free Newton-Krylov; (f) Full eigenvalue spectrum of the RandONet and reference Jacobians in their respective steady states; in the zoom in, the 3 leading eigenvalues computed with the Arnoldi method. (e)-(f) results are compared with a Finite Difference (FD) based time-stepper.}
    \label{fig:gap_tooth}
\end{figure*}

\paragraph{The ``Gap-Tooth'' NO} 
Finally, we demonstrate how a NO can be used to learn and exploit a local {\em in space} solution operator within the ``Gap-Tooth`` framework. 
In particular, we discretized spatial domain of the Liouville–Bratu–Gelfand PDE into $N_{\text{teeth}}=21$ teeth and $N_{\text{gaps}}=20$ gaps of equal width, so that the input functions are defined “inside the tooth,” evolving over smaller spatial domains. 
Thus, they require smoother, less variegated families of initial conditions.
This local‐in‐space setup avoids full‐domain coverage (here, coupled local NOs only evolve over half of the domain, and the other half --the gaps-- are recovered by smoothness/interpolation); this enables more data-efficient training, and accelerates EF computations.
For training and short-time predictions, we used sampled data with a local time step of $\Delta t = 0.0001$.
For approximating the Gap-Tooth solution operator, we used a RandONet with $300$ neurons in the branch and $100$ in the trunk.
We obtain results in a mean squared error (MSE)
of $1.37E$-$11$, a median L2 error of $5.13E$-$06$, and a maximum absolute error of $9.18E$-$05$. Notably, the training of the single-parameter ``Gap-Tooth'' RandONet took $0.15$ seconds. We provide further details on the data generation and training process in Appendix sections~\ref{App:Bratu_gap_tooth_data_generation} and~\ref{App:Bratu_gap_tooth_training}. We used the forward Euler method to integrate the ``Gap-Tooth'' RandONet. Figure~\ref{fig:gap_tooth} compares the time evolution between the ``Gap-Tooth'' RandONet time stepper and the reference ``Gap-Tooth'' solver. Figure~\ref{fig:gap_tooth_solution} shows the predicted evolution. As shown in Figure~\ref{fig:gap_tooth_error}, this evolution matches well with the reference trajectories. Figure~\ref{fig:gap_tooth_steady} compares both steady-state solutions, again demonstrating that the ``Gap-Tooth'' RandONet captures the steady-state behavior accurately.\par   
Spectral analysis of the ``Gap-Tooth'' scheme shown in Figures~\ref{fig:gap_tooth_full_spectrum} reveals that the three leading eigenvalues match very closely with the full reference problem solution ones.

\paragraph{The ``patch" NO}
We further extend the Gap-Tooth NO approach by incorporating Projective Integration, resulting in a Patch Neural Operator (Patch NO). This combines local-in-space and local-in-time evolution, enhancing both computational efficiency.
Specifically, we performed fifty short relaxation steps with \(\Delta t = 0.0001\) (total relaxation time \(0.005\)), followed by a large projective step of size \(\Delta T = 0.015\) using a backward finite-difference estimation of the solution time derivative.
Figures~\ref{fig:patches_solution} and~\ref{fig:patches_error} show the Patch NO timestepper and the corresponding error.
This error remains confined within $10^{-3}$, demonstrating excellent agreement with the reference solver.
As in the Gap-Tooth case, the Patch NO accurately reproduces the reference dynamics while reducing temporal resolution requirements through projective updates.


\subsection*{Case study 3: The Fitzhugh-Nagumo equations}
The Fitzhugh-Nagumo equations were first introduced in~\cite{fitzhugh1961impulses} to describe the propagation of an action potential as a traveling wave. 
Here, we considered the following system of PDEs,
describing the evolution of activator $u$ and inhibitor $v$ neurons:
\begin{equation}
\label{eq:FHN_PDE}
\begin{aligned}
&\frac{\partial u(x,t)}{\partial t}=D^{u}\frac{\partial^{2}u(x,t)}{\partial x^2}+u(x,t)-u(x,t)^3-v(x,t),\\ 
&\frac{\partial v(x,t)}{\partial t}=D^{v}\frac{\partial^{2}v(x,t)}{\partial x^2}+\varepsilon(u(x,t)-\alpha_1v(x,t)-\alpha_0),
\end{aligned}
\end{equation}
where, here, $\Omega = [0,L]$ with $L=20$; $D^u$ and $D^v$, $\alpha_0$, $\alpha_1$ are the respective constant diffusion and reaction parameters, and $\varepsilon$ is our bifurcation parameter. This problem is naturally endowed with homogeneous Neumann boundary conditions for both variables.
The bifurcation diagram is known to have a turning/tipping point and a supercritical Hopf bifurcation~\cite{galaris2022numerical, theodoropoulos2000coarse}.

To generate the training data for RandONet, we employed the $D1Q3$-Lattice-Boltzmann method (LBM)~\cite{galaris2022numerical,theodoropoulos2000coarse,bhatnagar1954model,qian1995scalings} to simulate the spatiotemporal dynamics through particles with discrete velocities, as detailed in Appendix section~\ref{App:FHN_LBM}.
In particular, a local time step of $\Delta t = 0.01$ was used during the simulation, and data snapshots were collected every $\Delta t = 0.1$ for training and short-time predictions of the NO. For longer time predictions, the NO was applied autoregressively.
We used random sigmoid-like initial conditions, and we considered bifurcation parameter values in the range $\varepsilon \in [0.005, 0.955]$. The lower value represents stable oscillating solutions beyond the Hopf bifurcation, while the upper value is the location of the saddle-node bifurcation.

\begin{figure*}[ht!]
    \centering
    \subfloat[error in $u(t,x)$]{
    \includegraphics[ width=0.31\linewidth]{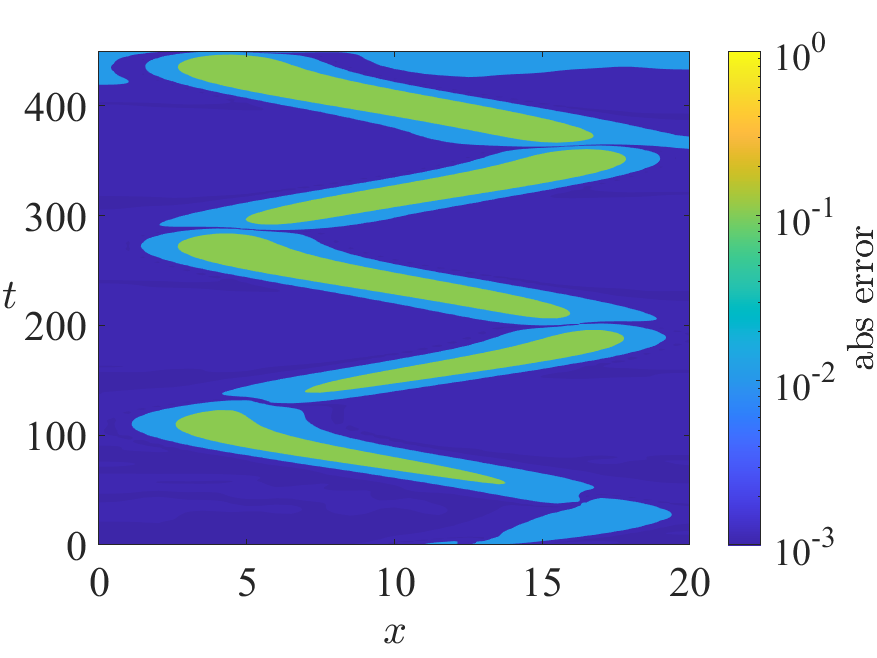}
    \label{fig:FHN_timestepper_error_u1}
    }
    \subfloat[error in $v(t,x)$]{
    \includegraphics[ width=0.31\linewidth]{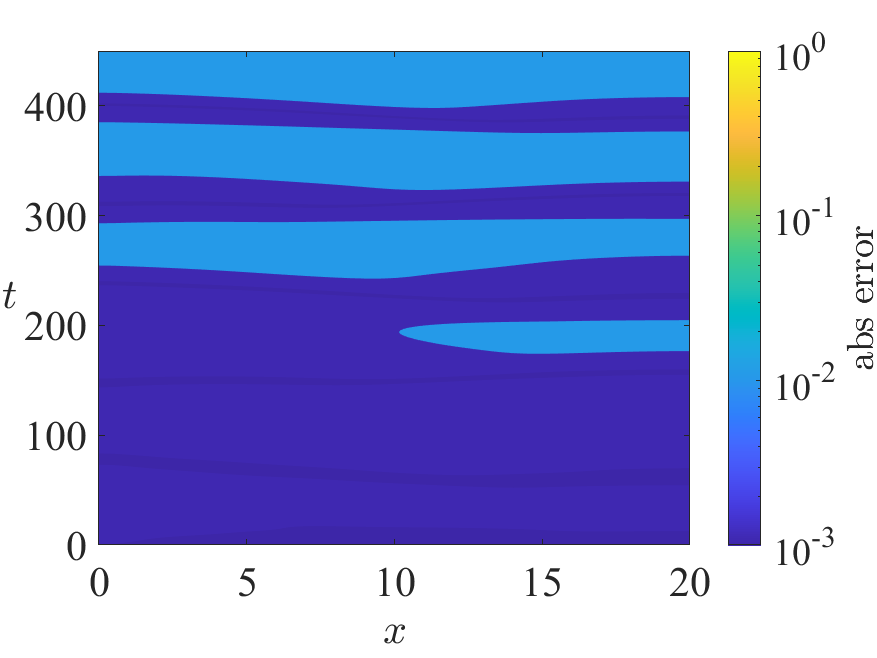}
    \label{fig:FHN_timestepper_error_v2}
    }
    \subfloat[Steady-state]{
    \includegraphics[ width=0.3\linewidth]{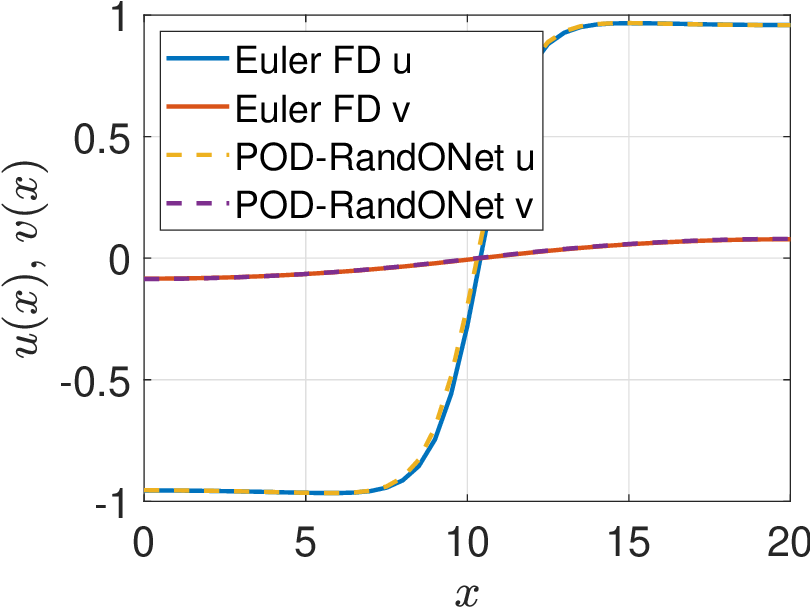}
    \label{fig:FHN_steady_NKGMRES}}\\
    \subfloat[full spectrum \hspace{3cm} leading eigenvalues (Arnoldi)]{
    \includegraphics[trim={0 0 2cm, 0.2cm},clip,width=0.8\textwidth]{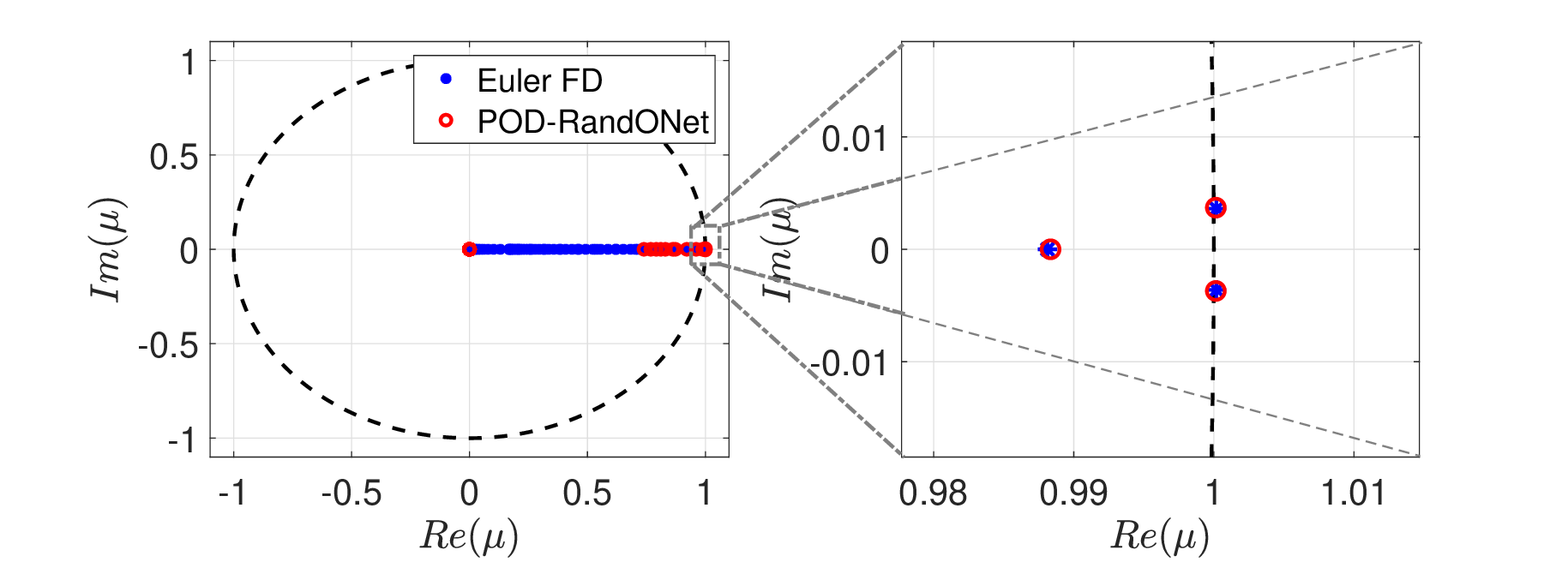}
\label{fig:FHN_full_spectrum}
    }
    \\
    \subfloat[bifurcation diagram]{
   \includegraphics[trim={0cm 0 1cm 0.2cm},clip, width=0.31\linewidth]{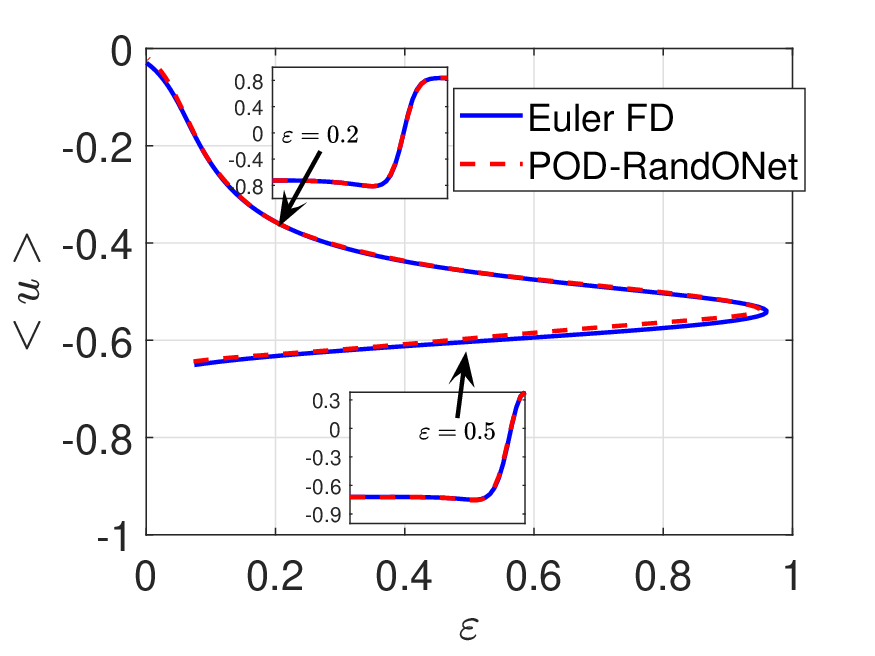}
   \label{fig:FHN_BD}
   }
   \subfloat[tracking leading eigenvalues]{
   \includegraphics[trim={0cm 0 2cm 0.2cm},clip,width=0.64\linewidth]{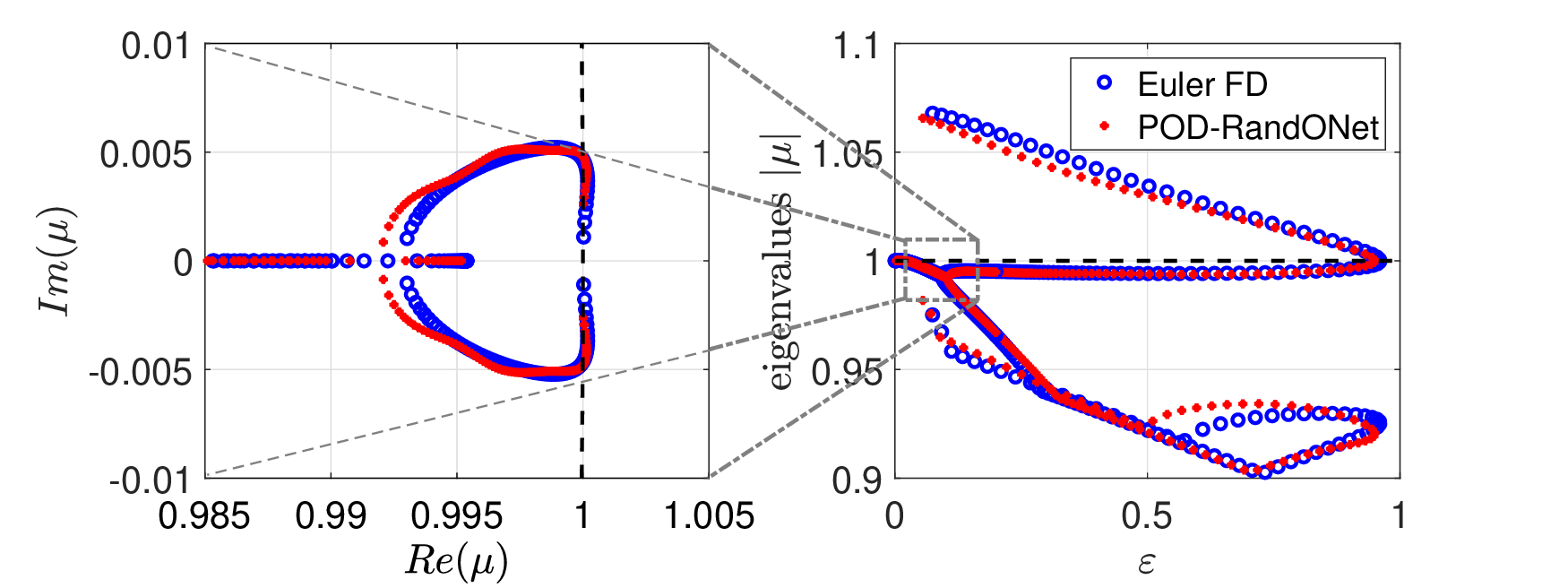}
\label{fig:FHN_BD_3leading_hopf}
   }
    \caption{POD-RandONet timestepper for the FHN PDEs in Eq.~\eqref{eq:FHN_PDE}, trained for $\varepsilon \in [0.005,0.995]$.
    (a)-(b) Space-time error of the RandONet vs. the ground-truth integration of the actual FHN PDEs, indicatively for $\epsilon=0.008$. (c-f) Comparison of the RandONet-based vs. a Finite Difference (FD)-based time-stepper of the actual FHN PDEs:
    (c) Steady-states with Newton-GMRES;
    (d) Full eigenvalue spectrum via the full Jacobian; (in blow up) the three leading eigenvalues obtained with the Arnoldi method.
    (e) Reconstructed bifurcation diagrams;
   (f) Tracking the real part of the two leading eigenvalues via the Arnoldi method;  zoom on their real and imaginary parts near the Hopf bifurcation. \label{fig:FHN_single_par}}
\end{figure*}

\paragraph{Bifurcation and Stability analysis}
As in the other problems, we first evaluated the accuracy of RandONets in approximating the time evolution, steady states, and spectral properties, at a fixed value of the bifurcation parameter (here at \(\epsilon = 0.008\)).
For this task, the RandONet consisted of $1000$ neurons in the branch with a random Fourier feature embedding, and $250$ in the trunk with sigmoidal activation functions. 
%
We compare the learned POD-RandONet time stepper against the reference Euler-based time stepper. In Figure~\ref{fig:FHN_single_par}, we report the performance of the POD-RandONets.
In Figures~\ref{fig:FHN_timestepper_error_u1} and~\ref{fig:FHN_timestepper_error_v2}, we show the space-time pointwise errors of the POD-RandONet, for a single representative trajectory.
Figure~\ref{fig:FHN_steady_NKGMRES} depicts the steady-state solutions obtained by the Jacobian-Free Newton-Krylov, on POD-RandONet and the Euler time stepper. In Figure~\ref{fig:FHN_full_spectrum}, we present the full eigenvalue spectrum of the POD-RandONet and compare it to the Euler time stepper.
By construction, the POD-RandONets only retain $12$ active eigenvalues and eigenvectors, with the remaining modes being mapped to zero due to the inherent dimensionality reduction in POD modes. While this limits the ability to precisely reproduce the full spectrum, the leading modes still provide a coherent and accurate view of the dynamical behavior.
Finally, the zoom-in of Figure~\ref{fig:FHN_full_spectrum} shows the three leading eigenvalues calculated with the Arnoldi method. The eigenvalues computed with the RandONet were accurate within an error of the order $10^{-3}$.
Full details on the training setup and convergence analysis are provided in Appendix section~\ref{App:FHN_training}.

Finally, we assessed the performance of the RandONet timesteppers to perform bifurcation and stability analysis. To learn the full dynamics in $\epsilon \in [-0.005,0.995]$, we trained a homotopy-based RandONet enhanced with POD modes. 
In the first layer, we applied a linear transformation using the leading eigenvectors, mapping the stacked function fields \([u(x); v(x)]\) to their corresponding POD modes. These modes are computed from all training data and retain $99.9\%$ of the total eigenvalue score, eventually reducing the dimensionality of the branch input from $82$ to $12$. The transformed POD modes are then fed into the homotopy-based RandONets. To construct the bifurcation diagram, we wrapped around the RandONet time stepper Newton-GMRES with the pseudo-arclength continuation technique. 
The bifurcation diagrams are displayed on top of each other in Figure~\ref{fig:FHN_BD}. The RandONet timestepper-based solutions are shown as a dashed red line, whereas the Euler solutions are shown as a solid black line. As shown, RandONets result in an accurate reconstruction of the solution branches for the whole range of the bifurcation parameter. 
We finally also traced the real parts of the first three leading eigenvalues (two real and one pair of complex conjugated eigenvalues) as computed by the RandONets upon steady states using the Arnoldi method. The diagram with the real parts vs. t bifurcation parameter is shown in Figure~\ref{fig:FHN_BD_3leading_hopf}. A zoom in Figure~\ref{fig:FHN_BD_3leading_hopf} depicts the real and imaginary parts of the leading eigenvalues for $\epsilon$ values close to the Hopf bifurcation. These images demonstrate that POD-RandONets accurately capture the key eigenvalues with a high degree of accuracy. As a result, we rediscover the saddle-node and Hopf points to a high precision.

\section*{Discussion}
In this work, we have demonstrated the application of {\em local} NOs -- local in time, over short-time windows, and possibly even local in space, over small spatial patches --  to perform system-level analysis, with a particular focus on long-time steady-state, stability, and bifurcation analysis. Through the use of advanced methods such as homotopy-based embedding NOs (for our illustrations RandONets) and the exploration of multiscale techniques like Projective Integration (PI) and ``Gap-Tooth'' schemes, we have shown how scientific machine learning can be integrated with traditional numerical analysis methods (in the absence of an explicit, closed-form equation) to offer efficient, scalable solutions to complex, spatiotemporally varying dynamical systems.

In this work, we performed some illustrative comparison between ``vanilla'' DeepONets and their similar but shallow and randomized counterparts -- ``vanilla'' RandONets -- specifically on the Allen-Cahn equation. For the Liouville–Bratu–Gelfand PDE and the FitzHugh–Nagumo equations, only RandONets were used, as they exhibited favorable performance and training efficiency in the first benchmark.
A detailed comparison with alternative NO architectures -- such as DeepONets and Fourier neural operators -- for learning PDE solution operators (including higher-dimensional PDEs and hyperbolic systems with conservation laws) remains an important direction for future research. While such architectures have demonstrated promise, they also present notable challenges. DeepONets and FNOs, in particular, often involve millions of parameters and require costly, high-dimensional non-convex optimization, which can hinder both training efficiency and generalization. In our experiments, these difficulties manifested, for example, in the Allen–Cahn benchmark, where the DeepONet-based timestepper struggled to yield reliable results for stability analysis. To mitigate such issues, potential improvements can include Sobolev training to better capture higher-frequency information induced by derivatives \cite{li2022learning, yu2022gradient, o2024derivative}, and the enforcement of dissipation to steer out-of-distribution predictions toward attractors \cite{li2022learning}. These directions remain exploratory, but they highlight the need for physically informed constraints in complex operator learning. More broadly, our key message is that every variant of neural operator -- including DeepONet and FNO -- can benefit from the combination of local training and classical numerical analysis workflows, regardless of the specific implementation or example.

It has already been argued (in~\cite{wang2023long}) that iterating short time neural operators (trained in their case not on data, but in an entirely self-supervised manner as {\em physics informed DeepONets}) exhibits advantages over long-time PINN training.
In our case, we argue that the use of such {\em local} in time (and possibly in space) NOs, trained either on data or in a physics-informed manner, opens up exciting algorithmic possibilities.
We anticipate that an important benefit - in the case of data-driven NOs - is the potential to ``get away with" much more small and parsimonious training data sets. Indeed, we expect that it is sufficient to train ``patch" NOs with relatively low-dimensional families of initial and boundary conditions (constant, linear, possibly quadratic) - much less variegated than large-space initial conditions and long-time boundary conditions (and, for that matter, different domain shapes). We also expect, in the spirit of~\cite{wang2023long}), that entirely self-supervised physics-informed NOs will also be easier to train if they are local-in-space $\times$ time. 

In this spirit, we provided an initial glimpse of the potential for further extending the use of {\em spatially localized} NOs in a so-called ``Gap-Tooth" space-time scheme; it will be straightforward to combine spatial {\em and temporal} local NOs in what equation-free terminology names {\em patch dynamics}~\cite{samaey2006patch}.

Finally, we have shown that our neural operators can also be used ``inside"  Projective Integration (PI) (as well as telescopic projective integration~\cite{gear2003telescopic}) schemes. PI can significantly accelerate transient simulations as long as the ``inner"  time-stepper can correctly dampen the fast transient modes. We demonstrated that a Local Neural Operator can be seamlessly included in such a framework. 

In current (and future) work, we extend our implementation to include adaptive time stepping, and to combine Projective Integration with the ``Gap-Tooth'' scheme to obtain solutions accelerated by {\em doubly} local (in space and in time) NOs. 
It is important to note that, with local NOs playing the role of {\em discretization points} in traditional numerical schemes -- with the ``right" flux communication between them, as prescribed by numerical analysis -- 
our approach can also be ``geometry-agnostic". Indeed, to the extent that different macroscale domains can be usefully modeled through point, or finite volume, discretizations, local patch NOs hold the promise of liberating a class of NO computations from needing to be trained on data from specific geometric domains.

\section*{Methods}
In this section, we present the Neural Operators, specifically RandONets and Homotopy-based RandONets that we have used in our framework. These approaches leverage random feature embedding architectures to efficiently approximate complex operators.

\subsection*{Neural Operator and Methodologies}
\subsubsection*{Preliminaries}
\label{sec:pre_deepOnets}
Here, we will briefly outline the general framework introduced in Chen and Chen work~\cite{chen1995universal}, which forms the basis for both DeepONets and RandONets. Consider the problem of approximating an operator $\mathcal{F}$ acting on functions $u \in \mathcal{U}$, sampled at $m$ fixed locations $x_j$ in the domain $\mathcal{K}_1$. The input vector $\bm{U} = (u(\bm{x}_1), \dots, u(\bm{x}_m)) \in \mathbb{R}^{m \times 1}$ is processed by a branch network, a single (or multi) hidden layer FNN with $M$ neurons, employing an activation function $\varphi^{br}$ and internal weights $\alpha^{br}_{ij} \in \R$ and biases $\beta^{br}_{i} \in \R$. While the trunk network, again a single (or multi) hidden layer FNN with $N$ neurons, with activation function $\varphi^{tr}$, weights $\bm{\alpha}^{tr}_k \in \R^d$ and biases $\beta^{tr}_k \in \R$, processes the new location $\bm{y} \in K_2 \subset \mathbb{R}^{1 \times d}$ to evaluate $\mathcal{F}[u]$. The hidden values of the branch and trunk networks are represented by $\bm{B} = (B_1, \dots, B_M) \in \mathbb{R}^{M \times 1}$ and $\bm{T} = (T_1, \dots, T_N) \in \mathbb{R}^{1 \times N}$, respectively:
\begin{equation}
    B_i(\bm{U}) = \varphi^{br} \left( \sum_{j=1}^{m} \alpha^{br}_{ij} u(\bm{x}_j) + \beta^{br}_{i} \right), \quad i = 1, \dots, M,
    \label{eq:branch}
\end{equation}
\begin{equation}
    T_k(\bm{y}) = \varphi_{tr} \left(\bm{y} \cdot \bm{\alpha}^{tr}_k + \beta^{br}_k \right), \quad k = 1, \dots, N.
    \label{eq:trunk}
\end{equation}
The network output is then the weighted inner product of the branch and trunk outputs:
\begin{equation}
    \mathcal{F}[u](\bm{y}) \approx \sum_{k=1}^{N} \sum_{i=1}^{M} w_{ki} B_i(\bm{U}) T_k(\bm{y}) = \bm{T} W \bm{B} = \langle \bm{T}, \bm{B} \rangle_W,
    \label{eq:branch_trunk}
\end{equation}
where $W \in \mathbb{R}^{N \times M}$ contains the weights $w_{ki}$.

Although Chen and Chen's original result uses shallow networks, DeepONet extends this by employing deep networks or other types of NNs. 
Training such large networks is computationally intensive and requires parallel computing and GPUs. To circumvent this, we used RandONets, that we introduced recently in~\cite{fabiani2025random} and for the completeness of the presentation we briefly describe below.

\subsubsection*{Random Projection-based Operator Networks (RandONets)}
\label{sec:RandONets}
In this section, we present the architecture of RandONets, that has been recently introduced in~\cite{fabiani2025randonets}. Also, in~\cite{fabiani2025randonets}, we have proven that RandONets are universal approximators of nonlinear operators. As proposed in the original Chen and Chen Theorem~\cite{chen1995universal}, RandONets use two single hidden layer FNNs with appropriate random bases as activation functions. 

Specifically, we propose using (parsimoniously chosen) randomized hyperbolic tangent bases to efficiently embed the space of spatial locations ($y$) ~\cite{fabiani2023parsimonious}. Thus, $\bm{\alpha}^{tr}_k\in\R^d$, $\beta^{tr}_k$, $k=1,\dots,N$ are i.i.d. randomly sampled from a continuous probability distribution function, as explained in~\cite{fabiani2025randonets, fabiani2025random, fabiani2021numerical, fabiani2023parsimonious}. 
%
For the branch network, here we have implemented the following Nonlinear random Fourier feature network (RFFN) embeddings~\cite{rahimi2007random}: 
\begin{equation}
    \begin{split}
        \bm{B}&=\bm{\varphi}_M^{br}(\bm{U};\bm{\alpha}^{br},\bm{\beta}^{br})=\\ 
    &=\sqrt{\frac{2}{M}}[\cos(\bm{\alpha}^{br}_1\cdot \bm{U}+\beta^{br}_1),\dots,\cos(\bm{\alpha}^{br}_M\cdot \bm{U}+\beta^{br}_M)],
    \end{split}
    \label{eq:RFFN_embedding}
\end{equation}
where we have two vectors of random variables $\bm{\alpha}^{br}$ and $\bm{\beta}^{br}$, 
from which we sample $M$ realizations. The weights $\bm{\alpha}^{br}_i$  are i.i.d. sampled from a standard Gaussian distribution, and the biases $\bm{\beta}^{br}_i$ are uniformly distributed in $\mathcal{U}[0,2\pi]$. This explicit random lifting has a low distortion for a Gaussian shift-invariant kernel distance.
RandONets training reduces to the solution of a linear least-squares problem in the unknowns $W$, that is, the external weights of the weighted inner product as in Eq.~\eqref{eq:branch_trunk}.

In what follows, we will discuss the particular treatment of the parameter-dependent embeddings we adopted. Specifically, we construct a homotopy across different values of the parameter. For a comprehensive description of the RandONet architecture and implementation, please refer to the original RandONets paper~\cite{fabiani2025randonets} and the additional implementation notes in Appendix section~\ref{App:randonets}, where we also describe the POD-enhanced RandONets used for treating and preconditioning the training of multi-field neural operators.

\subsubsection*{Handling parameter-dependent operators via Homotopy-Based Embedding}
\label{sec:homotopy}
In operator learning, handling parameter-dependent operators is particularly challenging. The `standard" approach consists in appending the parameter(s) directly to the branch-side of the network and mixing it within the function space representation. This approach can lead to a lack of distinction between the influence of the parameter and the function space. When the operator domain involves high-dimensional function inputs and low-dimensional parameters (e.g., a single scalar), naive embeddings may fail to capture the parameter's significance effectively.
Furthermore, when using randomized embeddings, the parameter information may be entangled or diluted within the randomized features, worsening the ability of the network to separate variations induced by the function input from those due to the parameter.
This problem also occurs in generic dual-branch architectures: when the parameter and function are processed separately but fused through naive operations (e.g., concatenation or addition), the resulting representation can lack the expressivity needed to capture strong or nonlinear parameter effects, especially in regimes characterized by bifurcations or critical transitions.

To mitigate these challenges, we introduce a novel parameter-dependent embedding strategy based on a continuous homotopy of branch hidden features $\bm{B}(\lambda)$, where $\lambda \in \R$ is the parameter of interest. 
For simplicity, assume that the parameter is rescaled to $\tilde{\lambda}\in[0,1]$. Our proposed homotopy-based embedding $\bm{B}_{\tilde{\lambda}}$ is constructed as a convex combination of two independent branch networks $\bm{B}_0$ and $\bm{B}_1$  (or, in the case of RandONets, two randomized matrices $R_0$ and $R_1$) corresponding to the extreme parameter values $\tilde{\lambda}=0$  and $\tilde{\lambda}=1$:
\begin{equation}
    \bm{B}_{\tilde{\lambda}}(\bm{U},\lambda)=(1-\tilde{\lambda})\bm{B}_0(\bm{U},\lambda)+\tilde{\lambda} \bm{B}_1(\bm{U},\lambda).
\end{equation}

Heuristically, the homotopy-based embedding improves parameter sensitivity and enhances the ability of NOs to capture interactions between the parameter and the input function, leading to more accurate and generalizable operator learning.

\section*{Data and code availability}
Data and code used to produce the findings of this study are publicly available on the GitHub repository \href{https://github.com/Centrum-IntelliPhysics/local-neural-operator-time-stepper-instead-of-just-time-stepper}{\texttt{local-neural-operator-time-stepper}}.

\bibliographystyle{naturemag-doi.bst}
\bibliography{AA_references.bib}

\section*{Acknowledgments}
G.F., H.V., S.G., and I.G.K. acknowledge the Department of Energy (DOE) support under Grant No. DE-SC0024162; I.G.K.also acknowledges partial support by the National Science Foundation under Grants No. CPS2223987 and FDT2436738. The authors would like to acknowledge computing support provided by the Advanced Research Computing at Hopkins (ARCH) core facility at Johns Hopkins University and the Rockfish cluster. ARCH core facility (\url{rockfish.jhu.edu}) is supported by the National Science Foundation (NSF) grant number OAC1920103.
C.S. acknowledges partial support by the PNRR MUR projects PE0000013-Future Artificial Intelligence Research-FAIR \& CN0000013 CN HPC - National Centre for HPC, Big Data and Quantum Computing, Gruppo Nazionale Calcolo Scientifico-Istituto Nazionale di Alta Matematica (GNCS-INdAM).
Any opinions, findings, conclusions, or recommendations expressed in this material are those of the author(s) and do not necessarily reflect the views of the funding organizations.

\section*{Author contributions statement}
\textbf{G.F.}: Conceptualization, Methodology, Software, Validation, Formal analysis, Investigation, Writing - Original Draft, Writing - Review \& Editing, Visualization. \\
\textbf{H.V.} Conceptualization, Methodology, Software, Validation, Formal analysis, Investigation, Writing - Original Draft, Writing - Review \& Editing, Visualization.\\
\textbf{S.G.} Methodology, Software, Validation, Formal analysis, Writing - Original Draft, Writing - Review \& Editing, Supervision.\\
\textbf{C.S.}: Conceptualization, Methodology, Validation, Formal analysis, Writing - Review \& Editing, Visualization, Supervision.\\
\textbf{I.G.K.}: Conceptualization, Methodology, Validation, Formal analysis, Resources, Writing - Review \& Editing, Visualization, Supervision.

\section*{Competing interest statement}
The authors declare that there are no known competing financial interests or personal relationships that could have appeared to influence the work reported in this manuscript.

\vspace{1cm}
\newpage
\textbf{\LARGE Appendix}
\appendix
\renewcommand{\thefigure}{S\arabic{figure}}
\setcounter{figure}{0}
\makeatletter
\renewcommand{\fnum@figure}{Supplementary Figure \thefigure}
\makeatother


\section{Further Details about RandONets implementation}
\label{App:randonets}

\paragraph{RandONets implementation for ``aligned" data.}
Let us assume that the training dataset consists of $s$ sampled input functions at $m$ collocation/grid points. Thus, the input is included in a matrix $U \in \R^{m\times s}$. Let us also assume that the output function can be evaluated on a fixed grid of $n$ points $\bm{y}_k \in \R^d$, which are stored in a matrix  $Y\in \R^{n\times d}$ (row-vector); $d$ is the dimension of the domain. In this case, we assume that for each input function $u$, we have function evaluations $v$ on the grid $Y$ stored in matrix $V=\mathcal{F}[U] \in \R^{n\times s}$.

Although this assumption may appear restrictive at first glance (as, for example, some values in the matrix $V$ could be missing, or $Y$ can be nonuniform and may change in time), nonetheless, for many problems in dynamical systems and numerical analysis, such as the numerical solution of PDEs, involves employing a fixed grid where the solution is sought. This is clearly the case for methods like FD-based or FEM-based numerical schemes without mesh adaptation.
Additionally, even in cases where the grid is random or adaptive, there is still the opportunity to construct a ``regular" output matrix $V$ through ``routine" numerical interpolation of outputs on a fixed regular grid.
Now, given that the data are aligned, following Eq.~\eqref{eq:branch_trunk}, we can solve the following linear system (double-sided) of $n\times s$ algebraic equations in $N\times M$ unknowns:
\begin{equation}
    V=\mathcal{F}[U]= \bm{\varphi}_n^{tr}(Y;\bm{\alpha}^{tr}) \,  W  \, \bm{\varphi}_m^{br}(U;\bm{\alpha}^{br})=T(Y) \, W \, B(U).
    \label{eq:RandONets}
\end{equation}
Let us observe that --differently from a classical system of equations (e.g., $Ax=b$), here we have two matrices from the trunk and the branch features, that multiply the readout weights $W$ on both sides.\par
Although the number of unknowns and equations appears large due to the product, the convenient alignment of the data allows for effective operations that involve separate and independent (pseudo-) inversion of the trunk/branch matrices $T(Y)\in \R^{n\times N}$ and $B(U)\in\R^{M\times s}$.
Thus, the solution weights of Eq.~\eqref{eq:RandONets}, can be found by employing methods such as the Tikhonov regularization~\cite{golub1999tikhonov}, tSVD, QR/LQ decomposition and COD\cite{hough1997complete} of the two matrices, as we will detail later, obtaining:
\begin{equation}
    W=\big(T(Y)\big)^{\dagger} \, V \, \big(B(U)\big)^{\dagger}.
    \label{eq:RandONets_solve}
\end{equation}
As one might expect, the trunk matrix typically features smaller dimensions compared to the branch matrix. This is because the branch matrix can involve numerous samples $s$ of functions (usually exceeding the number $n$ of points in the output grid), along with a larger number of neurons $M$ required to represent the high-dimensional function-input, as opposed to the $N$ neurons of the trunk net, which embeds the input space.
In the end, the computational cost associated with the training (that is, the solution of the linear least squares problem) of RandONets is of the order $O(M^2 s+s^2M)$.
Here, we use the SVD-based Moore-Penrose pseudo-inverse for the inversion of both $T$ and $B$ matrices.

From a numerical point of view, the resulting randomized trunk $T(Y)$ and branch $B(U)$ matrices tend to be ill-conditioned. Therefore, in practice, we suggest solving Eq.~\eqref{eq:RandONets} via a truncated SVD (tSVD), Tikhonov regularization, QR decomposition, or regularized COD~\cite{hough1997complete, fabiani2025random, fabiani2025randonets}. Here, we describe the procedure for the branch network. For the trunk matrix, the procedure is similar.\par 

The regularized pseudo-inverse $(B(U))^{\dagger}$, for the solution of the problem in Eq.~\eqref{eq:RandONets_solve} is computed as:
\begin{equation}
\begin{split}
&B(U)=U\Sigma V^T=[U_r \quad \tilde{U}]
    \begin{bmatrix}
        \Sigma_k & 0\\
        0 & \tilde{\Sigma}
    \end{bmatrix}
     [V_k \quad \tilde{V}]^T
    , \\
    &(B(U))^{\dagger}=V_k\Sigma_k^{-1}U_r^T,
\end{split}
\label{eq:pseudo_inverse_RandONet}
\end{equation}
where the matrices $U=[U_k \quad \tilde{U}]\in \R^{k\times n}\oplus\R^{(n-k)\times n}$ and $V=[V_k \quad \tilde{V}] \in \R^{k\times s}\oplus\R^{(s-k)\times s}$ are orthogonal and $\Sigma \in \R^{n\times s}$ is a diagonal matrix containing the singular values $\sigma_i=\Sigma_{(i,i)}$. Here, we select the $k$ largest singular values that exceed a specified tolerance $0<\epsilon\ll 1$, i.e., $\sigma_1,\dots,\sigma_k>\epsilon$, effectively filtering out insignificant contributions and improving numerical stability. For implementation, we use the \texttt{Matlab} built-in function \texttt{pinv}, enabled with GPU computations by using \texttt{GpuArray} format.\par

For the ``unaligned" data case, the interested reader can refer to~\cite{fabiani2025randonets, lu2021learning}.

\paragraph{POD enanched RandONets: Preconditioned Random Projections for Multi-Field Systems}
When extending the RandONet regression framework to approximate the dynamics to systems of PDEs, involving multiple fields -- each possibly evolving on different amplitude scales -- randomized feature construction becomes more challenging.
In such settings, the embedding process may efficiently capture the discretization structure of a single field, but identifying an effective joint representation across several fields often requires a much larger embedding dimension. Specifically, the number of neurons in the branch network may need to grow significantly, potentially even exponentially, with the number of fields involved.

For example, in single-field cases, such as the Allen–Cahn and Bratu–Gelfand problems, we observed no major difficulty in scaling up to $2000$ neurons in the branch layer. Obtaining smooth convergence as the width increased, as detailed in Appendix sections~\ref{App:AC_training} and~\ref{App:Bratu_training}. However, in the two-field FitzHugh–Nagumo (FHN) system, we did not observe a similar behavior. Increasing the number of neurons far beyond $2000$ was not feasible in our setting due to memory constraints during the matrix pseudo-inversion step in RandONets, which exceeded the capacity of our available NVIDIA GPU.

To address this, we introduced a dimensionality reduction step before applying the random projection. Specifically, we computed a low-rank representation of the input data using Proper Orthogonal Decomposition (POD). Given a data matrix \( U \in \R^{m \times s}\) of the input function, comprising the stacking of multiple discretized fields in space, we formed the covariance matrix \( UU^T \) and computed its spectral decomposition \( UU^T = V \Lambda V^T \). We then projected the data onto the first \( k \) POD modes, using the matrix \( V_k \in \R^{m\times k}\) formed by the first \( k \) POD eigenfunctions, with $k<<m$.

After obtaining the reduced features \( \tilde{U} = V_k^T U 
 \in \R^{k\times s}\), we applied a random projection \( R \in \R^{N\times k}\) to \( \tilde{U} \). This two-step procedure can be viewed as a composed linear random projection \( \tilde{R} = R V_k^T \in \R^{N\times m}\), where the POD serves as a preconditioner to the randomized projection. This approach allows the random features to operate efficiently on a lower-dimensional and more structured representation of the data, improving scalability and robustness in multi-field systems.

\section{Further Details on the Allen-Cahn PDE}
\subsection{Data generation for the Allen-Cahn Equation}
\label{App:AC_data_generation}
The Allen-Cahn PDE in Eq.~\eqref{eq:AC_PDE} is solved numerically to generate training and test data for different values of the bifurcation parameter $\epsilon$.
We simulate the PDE using the \texttt{chebfun}~\cite{platte2010chebfun, driscoll2014chebfun} package in \texttt{Matlab}, which uses a pseudo-spectral Chebyshev method for spatial discretization of the PDE~\cite{platte2010chebfun,driscoll2014chebfun}, combined with a method of lines approach. We further use implicit time stepping of variable order and variable step size as implemented in the chebfun suite \texttt{pde15s} solver, an expansion of the well-known \texttt{ode15s}~\cite{shampine1997matlab}. The boundary conditions are set to Neumann-type, imposing $\partial u/\partial x = 0$ at $x = \pm 1$. We vary the parameter $\epsilon$ between $0.22$ and $0.7$ across $N_{\text{par}} = 25$ evenly spaced values.

To maintain data efficiency, we subsample the dense output obtained by the Chebyshev spectral method on an equidistant grid in both space and time. Physically, we placed ``sensors" to collect the output of the system at specific space-time locations. The spatial domain $x \in [-1, 1]$ is discretized at $N_x = 100$ grid points, and the temporal domain $t \in [0, T]$ is sampled at $N_t = 41$ time steps, corresponding to a time step size $\Delta t = 0.01$.

The initial conditions reflect various types of transient solutions. They are based on the bifurcation structure, e.g., the sine-type and cosine-type steady states. Specifically, transient solutions are generated as follows:
\begin{itemize}
    \item \textbf{Sine and Cosine Modes:} Initial conditions are sine and cosine functions based on the bifurcation points derived in equations~\eqref{eq:sin_bf} and~\eqref{eq:cos_bf}. For each $\epsilon$, we perturb the solutions by random amplitudes to account for slight deviations from the exact bifurcation solutions. The sine modes are given by $\sin((n-1)\pi x + \frac{\pi}{2})$, where $n$ is determined by the bifurcation point, while the cosine modes are of the form $\cos(n \pi x)$.
    \item \textbf{Flat and Zero Solutions:} In addition to the sine and cosine modes, initial conditions also include zero solutions, which represent the unstable steady state, and flat solutions with random perturbations in the range $[-1.5, 1.5]$ to explore the solution space more broadly.
    \item \textbf{Random Gaussian Profiles with Cosine Modulation:} We create additional initial conditions by combining Gaussian functions with randomized centers, widths, and amplitudes, multiplied by cosine waves to enforce the boundary conditions. These Gaussian profiles are of the form 
    \begin{equation}
    u_0(x) = \sum_{i=1}^{10} w_i \exp\left(-\text{s}_i (x - \text{c}_i)^2\right) (1+\cos(\pi x))/2,
    \end{equation}
    where $w_i$, $\text{s}_i$, and $\text{c}_i$ are randomly chosen parameters. This form ensures that the boundary conditions $\frac{\partial u}{\partial x} = 0$ at $x = \pm 1$ are satisfied.
\end{itemize}

For each $\epsilon$ and initial condition, the solution $u(x, t)$ is computed and stored at all time steps. We duplicate the dataset with symmetric solutions $-u(x, t)$ due to the symmetric structure of the Allen-Cahn equation. The dataset is then organized into training and test sets, with:
\begin{itemize}
    \item \textbf{Branch Inputs:} encoded as $\{u(x, t), \epsilon\}$ at a given time.
    \item \textbf{Output Labels:} corresponding to $u(x, t + \Delta t)$ at the next time step.
\end{itemize}
A healing time of $0.03 s$, is introduced to remove initial sharp transient effects.
The final training dataset consists of 22,000 pairs of input functions and the parameter $\epsilon$, denoted as $U = [u_i(t); \varepsilon_i] \in \mathbb{R}^{101 \times 22259}$, where the functions $u$ are discretized over a 100-point grid. The corresponding outputs are stored in $V = u_i(t + \Delta t) \in \mathbb{R}^{100 \times 22259}$. For testing purposes, we used 5,565 different input/output function samples.

\begin{figure*}[ht!]
    \centering
    \subfloat[$\epsilon=0.22$]{
    \includegraphics[ width=0.4\linewidth]{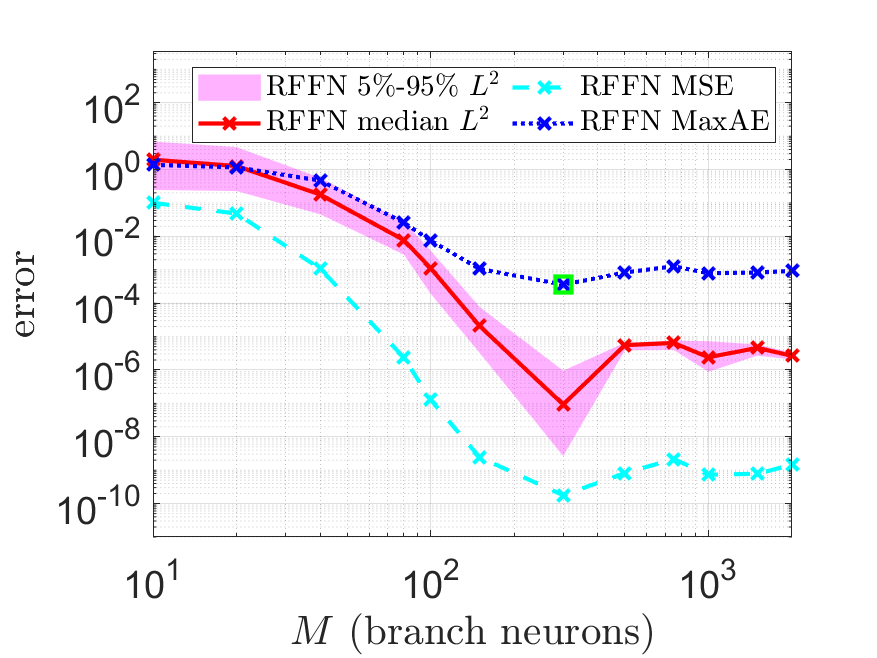}
    \label{fig:AC_training_single_neuron}
    }
    \subfloat[$\epsilon=0.22$]{
    \includegraphics[ width=0.4\linewidth]{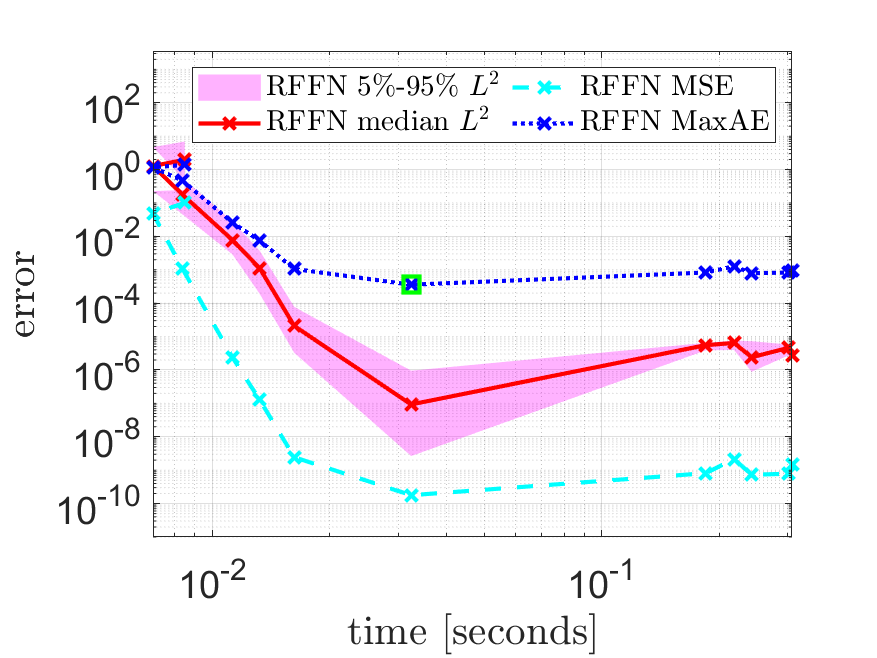}
    \label{fig:AC_training_single_time}
    }\\
    \subfloat[ $\epsilon \in(0.22,0.7)$]{
    \includegraphics[ width=0.4\linewidth]{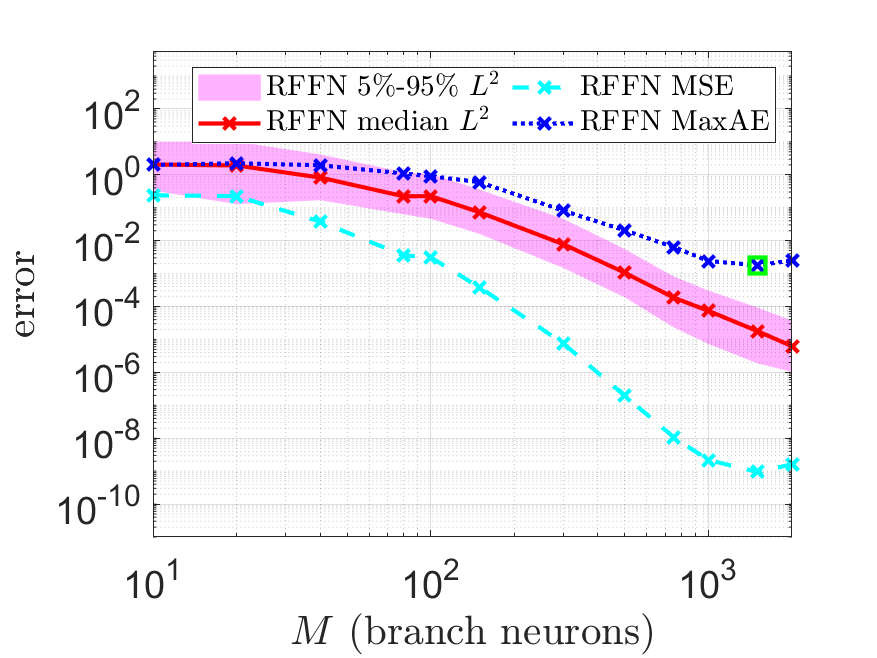}
    \label{fig:AC_training_multi_neuron}
    }
    \subfloat[ $\epsilon \in(0.22,0.7)$]{
    \includegraphics[ width=0.4\linewidth]{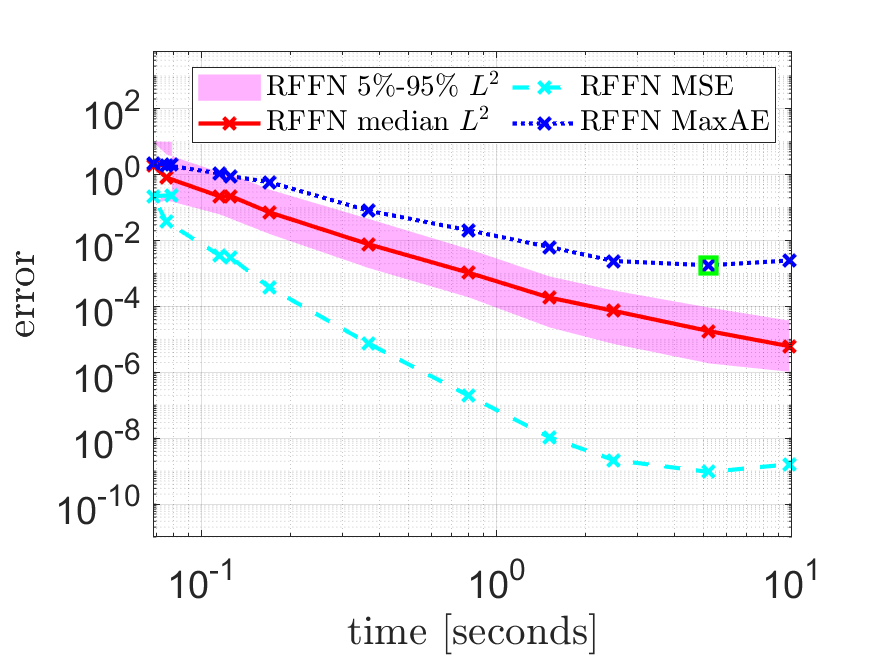}
    \label{fig:AC_training_multi_time}
    }
    \caption{Training convergence of the RandONets for the Allen-Cahn PDE in Eq.~\eqref{eq:AC_PDE}.
    (a)-(b) Single value $\epsilon=0.22$ with vanilla RandONets. (c)-(d) Multi paramater $\epsilon \in (0.22,0.7)$ with Homotopy-based RandONets.
    Convergence of the RandONets w.r.t (a)-(c) the number of neurons in the branch and (b)-(d) the computational time. We report the Maximum absolute error (MaxAE), the Mean Square Error (MSE) and the median of the $L^2$ error and the range between $5\%$ and $95\%$ percentiles.}
    \label{fig:RandONets_AC_training}
\end{figure*}

\subsection{Training of the RandONet Neural Operators}
\label{App:AC_training}

In \texttt{Matlab}, we train RandONet architectures on the set of generated data. In particular, the number of neurons in the trunk layer is set to 200. We vary the number of neurons in the hidden branch layer: $\{10, 20, 40, 80, 100, 150, 300, 500, 750, 1000, 1500, 2000\}$. Our code investigates the convergence of the method with respect to these neuron counts.

We consider two scenarios: the single-parameter case ($\varepsilon=0.22$) and the multi-parameter case.
For the first simpler scenario, we used a RandONet with $2000$ neurons in the branch and $300$ in the trunk, achieving a mean-squared error (MSE) of \(10^{-14}\), a median \(L^2\) error of \(10^{-7}\), and a maximum absolute error of \(10^{-5}\), for the test set.
In the second scenario, we employ the homotopy-based RandONets as described in the Methods section and results are reported in the main text. 
We train over $10$ iterations, and evaluate the error on a separate test dataset, which is subsampled from the training data, to select the best model. The \texttt{MATLAB} code utilizes the \texttt{Parallel Computing Toolbox}, which supports GPU acceleration, to accelerate the training process. We specifically report the following metrics:
\begin{itemize}
\item Maximum Error (MaxAE): The maximum absolute error between the predicted and true outputs.
\item MSE (Mean Squared Error), i.e., the test loss
\item L2 Error, we report the median across the set of functions and the range between $5\%$ and $95\%$ percentiles.
\end{itemize}

In Supplementary Figure~\ref{fig:AC_training_single_neuron}, we report the results for the database when only the parameter $\epsilon=0.22$ is considered. We also report the corresponding execution times in Supplementary Figure~\ref{fig:AC_training_single_time}.

In the Supplementary Figure~\ref{fig:AC_training_multi_neuron}, we present the convergence error in terms of MaxAE, L2, and MSE across the test dataset. The corresponding execution times are shown in the Supplementary Figure~\ref{fig:AC_training_multi_time}.
Please note that each point in the figures represents a different architecture, with the given number of neurons in the branch-net. Each architecture was trained in a single step, without multiple training epochs, using only the two pseudo-inverse computations of the trunk and branch collocation matrices, as described in Methods.
For the other equation-free computations presented in the main text, we selected the best-performing RandONets, based on the minimum MaxAE, which are highlighted by a green square in the figures.

\section{Further Details on the Liouville-Bratu-Gelfand PDE}
\subsection{Data generation for the Liouville-Bratu-Gelfand PDE}
\label{App:Bratu_data_generation}
We discretize the spatial domain $[0, 1]$ in $N_x = 51$ grid points, and the temporal domain $t \in [0, T]$ in $N_t = 31$ time steps, corresponding to a step size of $\Delta t = 0.001$. We vary the parameter $\lambda$ between 0 and 3.8 across $N_{\text{par}} = 39$ evenly spaced values.

We simulate the PDE using the \texttt{chebfun}~\cite{platte2010chebfun, driscoll2014chebfun} package in \texttt{Matlab}, which uses a pseudo-spectral Chebyshev method for spatial discretization of the PDE~\cite{platte2010chebfun,driscoll2014chebfun}, combined with a method of lines approach. We further use implicit time stepping of variable order and variable step size as implemented in the \texttt{chebfun} suite \texttt{pde15s} solver, an expansion of the well-known \texttt{ode15s}~\cite{shampine1997matlab}.

The initial conditions are drawn from the following parametrized family of functions: 
\begin{equation}
\begin{split}
    u_0(x) = l\left(x-x^2\right)  \left(\exp(-s_g(x-c_g))+\frac{1}{2}\right)\\ \left(A_c \cos (2\pi n_c x)+1\right) \left(A_s \sin (2\pi n_s x)+1\right),
\end{split}
\end{equation}  

where the parameters are randomly sampled from uniform distributions:  
\begin{equation}
\begin{split}
    n_c, \, n_s, s_g \sim \mathcal{U}(0,10), \quad A_c, \, A_s \sim \mathcal{U}(0,0.2), \\ l \sim \mathcal{U}(0,3), \quad c_g \sim \mathcal{U}(1/4,3/4).
\end{split}
\end{equation}  

This choice of random functions ensures that the Dirichlet boundary conditions are satisfied because the \((x-x^2)\) term vanishes at $ x=0 $ and $ x=1 $. The exponential function introduces a skewed Gaussian-like modification centered at $ c_g $, while the sine and cosine terms act as random small perturbations.

We compute the transient solution $u(x,t)$ for each initial condition and value of $\lambda$, and store it at all time steps. The inputs to the neural operator then become
\begin{itemize}
    \item \textbf{Branch Inputs:} Encoded as $\{u(x, t), \lambda\}$ at a given time.
    \item \textbf{Output Labels:} Corresponding to $u(x, t + \Delta t)$ at the next time step.
\end{itemize}
We also introduce a healing time of $0.001 s$ to remove sharp transient effects from the initial evolution because they complicate training.

\begin{figure*}[ht!]
    \centering
    \subfloat[ $\lambda=1$]{
    \includegraphics[ width=0.4\linewidth]{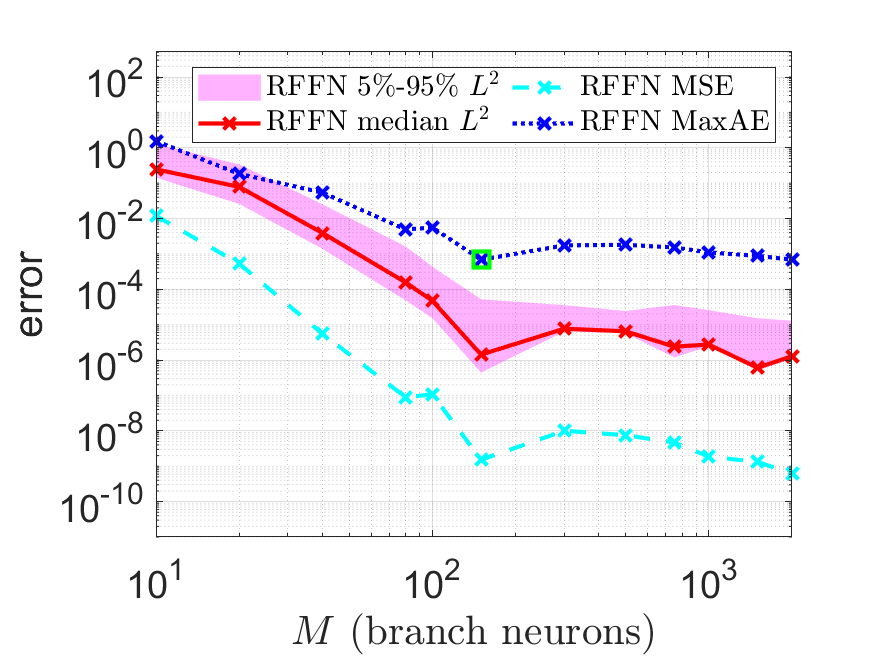}
    \label{fig:bratu_training_single_neurons}
    }
    \subfloat[ $\lambda=1$]{
    \includegraphics[ width=0.4\linewidth]{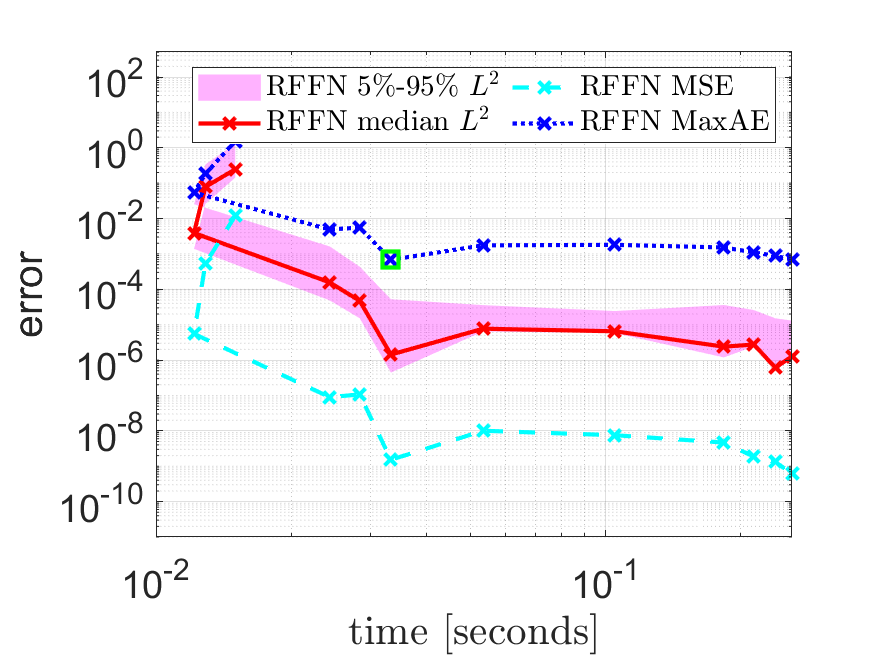}
    \label{fig:bratu_training_single_time}
    }\\
    \subfloat[ $\lambda\in (0,3.8)$]{
    \includegraphics[ width=0.4\linewidth]{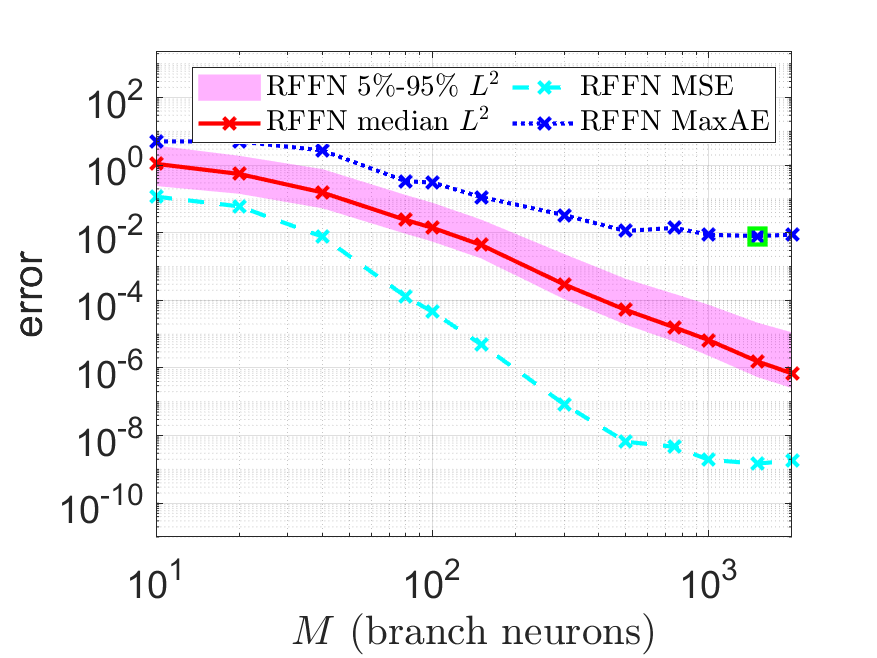}
    \label{fig:bratu_training_multi_neurons}
    }
    \subfloat[ $\lambda\in(0,3.8)$]{
    \includegraphics[ width=0.4\linewidth]{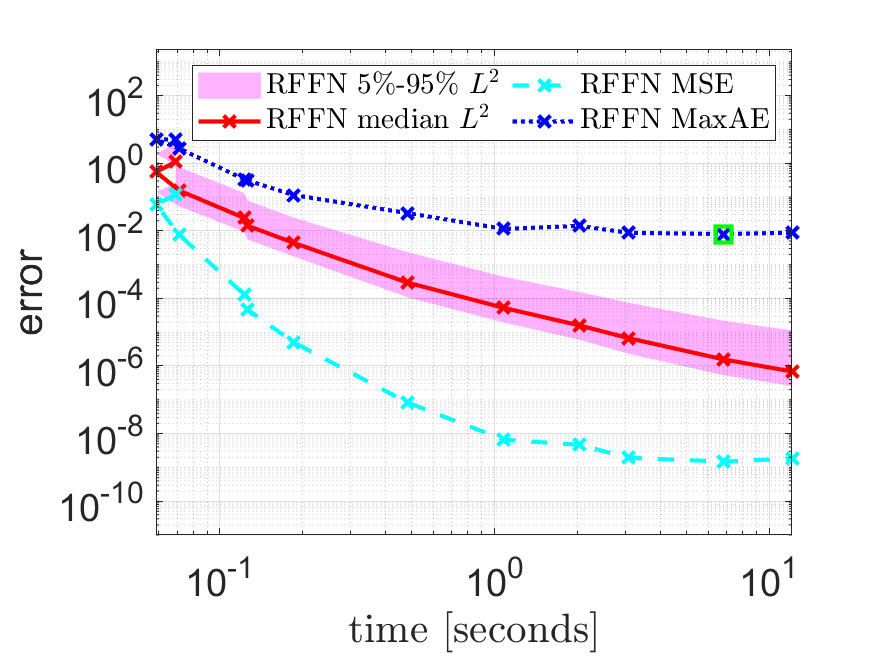}
    \label{fig:bratu_training_multi_time}
    }
    \caption{Training convergence results of the RandONets for the Liouville-Bratu-Gelfand PDE in Eq.~\eqref{eq:Bratu_parabolic}. 
    (a)-(b) RandONet for Single value $\lambda=1$ of the Liouville-Bratu-Gelfand PDE. (c)-(d) Homotopy-based RandONets for the parametric Liouville-Bratu-Gelfand PDE.
    Convergence of the RandONets w.r.t ((a)-(c)) the number of neurons in the branch and ((b)-(d)) the computational time. We report the Maximum absolute error (MaxAE), the Mean Square Error (MSE) and the median of the $L^2$ error and the range between $5\%$ and $95\%$ percentiles.
    \label{fig:RandONets_bratu_training}
    }
\end{figure*}

\begin{figure*}[ht!]
    \centering
    \subfloat[ $\lambda=1$]{
    \includegraphics[ width=0.4\linewidth]{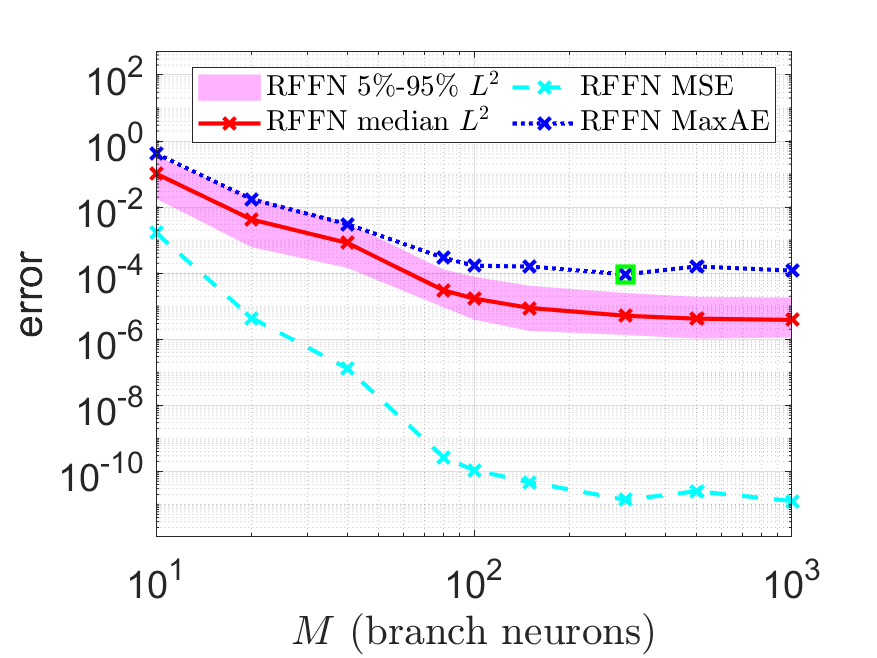}
    \label{fig:bratu_gap_tooth_training_single_neurons}
    }
    \subfloat[ $\lambda=1$]{
    \includegraphics[ width=0.4\linewidth]{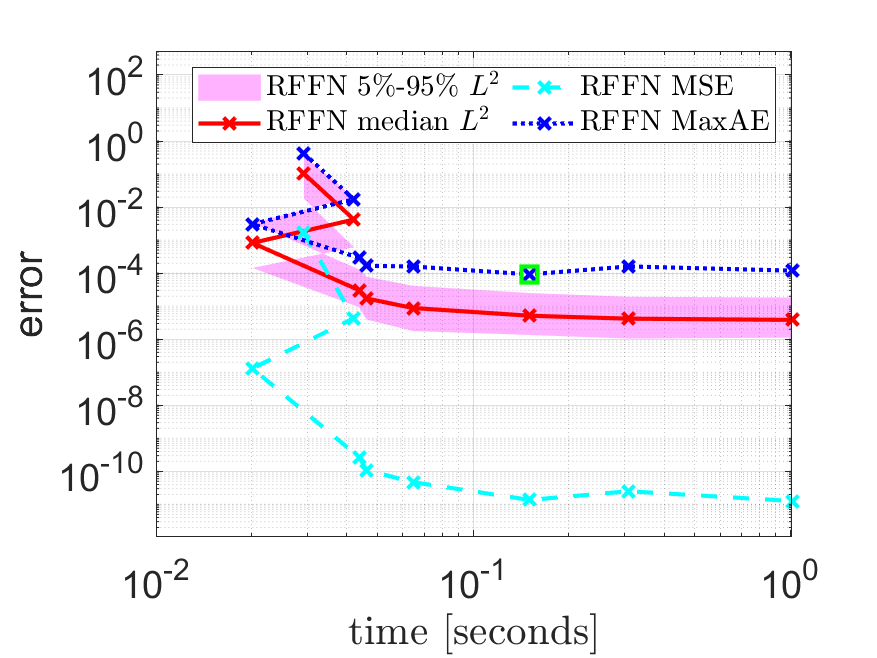}
    \label{fig:bratu_gap_tooth_training_single_time}
    }
    \caption{Training convergence results of the RandONets for the Gap-Tooth Liouville-Bratu-Gelfand PDE in Eq.~\eqref{eq:Bratu_parabolic}. 
    (a)-(b) RandONets for Single value $\lambda=1$. Convergence of the RandONets w.r.t (a) the number of neurons in the branch and (b) the computational time. We report the Maximum absolute error (MaxAE), the Mean Square Error (MSE) and the median of the $L^2$ error and the range between $5\%$ and $95\%$ percentiles.
    \label{ fig:RandONets_Gap_tooth_bratu_training}}
\end{figure*}

\subsection{Training of the RandONet Neural Operators for the Liouville-Bratu-Gelfand PDE}
\label{App:Bratu_training}
In \texttt{Matlab}, we train RandONet architectures on the training data for the Liouville-Bratu-Gelfand PDE and we evaluate its performance. The process is set up for both the parametric and non-parametric scenarios. In particular, the number of neurons in the trunk layer is set to $150$. We train RandONets for a different number of neurons in the branch hidden layer: $[10, 20, 40, 80, 100, 150, 300, 500, 750, 1000, 1500, 2000]$. Our code investigates the convergence of the method with respect to these neuron counts in the branch net.

For the multi-parameter case, we employ the homotopic RandONets as described before in Methods section. The training process is repeated for $10$ iterations, and we compare the $L^\infty$-error to select the best RandONet model among all iterations and different branch net size. The code supports GPU acceleration, which speeds up the training process, with the use of \texttt{gpuArray} of the \texttt{parallel computing toolbox} of \texttt{Matlab}.

We report the following test metrics:
\begin{itemize}
\item Maximum Error (MaxAE) ($L^\infty$): The maximum absolute error between the predicted and true outputs.
\item MSE (Mean Squared Error).
\item L2 Error, we report the median across the set of functions and the range between $5\%$ and $95\%$ percentiles.
\end{itemize}
For the other computations, we have selected the best performing RandONets, highlighted by a green square point in the figures.

In Supplementary Figure~\ref{fig:bratu_training_single_neurons}, we report the results for the dataset with parameter $\lambda=1$ fixed. We also report the corresponding execution times in Supplementary Figure~\ref{fig:bratu_training_single_time}. Furthermore, in Supplementary Figure~\ref{fig:bratu_training_multi_neurons} we report the convergence error in terms of MaxAE, L2 and MSE across the test dataset. We also report the corresponding execution times in Supplementary Figure~\ref{fig:bratu_training_multi_time}.

\subsection{Data generation for the Gap-Tooth Liouville-Bratu-Gelfand PDE}
\label{App:Bratu_gap_tooth_data_generation}
Analogous to the entire domain case, we solve the Liouville-Bratu-Gelfand PDE numerically to generate training and testing datasets for different values of the bifurcation parameter $\lambda$. 

We need to pay special attention to the spatial discretization because there are two grid sizes involved: the size of the teeth and gaps, as well as the number of grid points in each tooth. We discretize the spatial domain $[0, 1]$ into $N_{teeth}=21$ and $N_{gaps}=20$, where each tooth has $N_x = 21$ grid points, leading to a total of 821 equidistant points. The temporal domain $t \in [0, T]$ is discretized into $N_t = 41$ time steps, corresponding to a step size of $\Delta t = 0.0001$ (10 times smaller than the full domain simulation due to stability constraints). The parameter $\lambda$ is varied between 0 and 3.8 across $N_{\text{par}} = 39$ evenly spaced values.
To generate data we use \texttt{chebfun}~\cite{platte2010chebfun, driscoll2014chebfun} package in \texttt{Matlab}, analogously as before.  The initial conditions are drawn from the same family of functions, but with slightly different parameter distributions,  
\begin{equation}
\begin{split}
    n_c, \, n_s, s_g \sim \mathcal{U}(0,20), \quad A_c, \, A_s \sim \mathcal{U}(0,0.33), \\ l \sim \mathcal{U}(0,3), \quad c_g \sim \mathcal{U}(1/6,5/6).
\end{split}
\end{equation}  
This approach incorporates greater variation in the dataset because we use more discretization points and local functions. These local functions exhibit smoother behavior on each tooth compared to the previous full-domain case. For each $\lambda$ and initial condition, the solution $u(x, t)$ is computed and stored at all time steps. We again use a healing time of $0.0005 s$ to remove sharp transient effects from the initial conditions.

\subsection{Training of the RandONet Neural Operators for the Gap-Tooth Liouville-Bratu-Gelfand PDE}
\label{App:Bratu_gap_tooth_training}
The input to this local neural operator is the current profile $u(x,t)$ in a small spatial region, often called a `tooth', and the output is the profile at the next point in time. In the classical gap-tooth scheme, one uses Neumann boundary conditions coming from a global interpolating spline to link all the teeth together over the (possibly) large gaps. Our local NO must therefore take two extra branch inputs to represent these Neumann boundary conditions. However, instead of constructing the interpolating spline after every time step to compute its derivative, we select the mean values in the neighboring teeth as extra inputs because they contain the same information.
Finally, in order to handle the homogeneous Dirichlet boundary conditions at $x=0$ and $x=1$, we introduce two ghost cells with mean values set to zero. These ghost values are then passed to the local and global NOs at the left-most and right-most ``teeth".

In \texttt{Matlab}, we train different RandONet architectures for the single-parameter case. In particular, the number of neurons in the trunk layer is set to $100$. We also vary the number of neurons in the branch hidden layer in the set $\{10, 20, 40, 80, 100, 150, 300, 500, 750, 1000\}$. Our code investigates the convergence of the method with respect to these neuron counts in the branch net.

The training process is repeated over $10$ iterations, and we compute the $L^\infty$-error to select the best RandONet model among all iterations and different branch net size. The code supports GPU acceleration, which speeds up the training process, using \texttt{gpuArray} of the \texttt{parallel computing toolbox} of \texttt{Matlab}.

For the other computations, we have selected the best performing RandONets, highlighted by a green square point in the figures. In Supplementary Figure~\ref{fig:bratu_gap_tooth_training_single_neurons}, we report the single-parameter results. We also report the corresponding execution times in the Supplementary Figure~\ref{fig:bratu_gap_tooth_training_single_time}.

%

\begin{figure*}[ht!]
\subfloat[]{
    \includegraphics[width=0.4\textwidth]{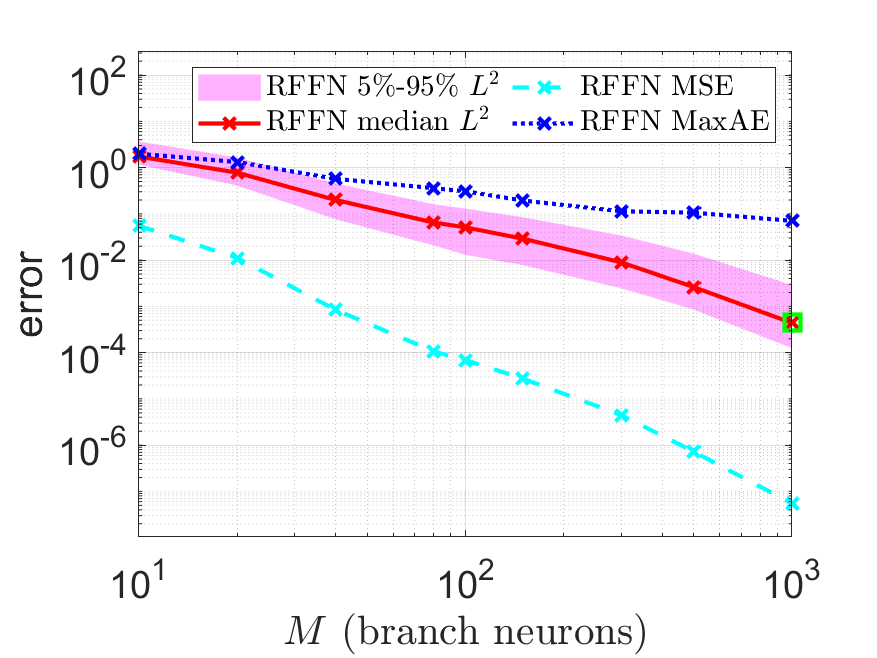}
    \label{fig:FHN_training_multi_neuron}
    }
    \subfloat[]{
    \includegraphics[width=0.4\textwidth]{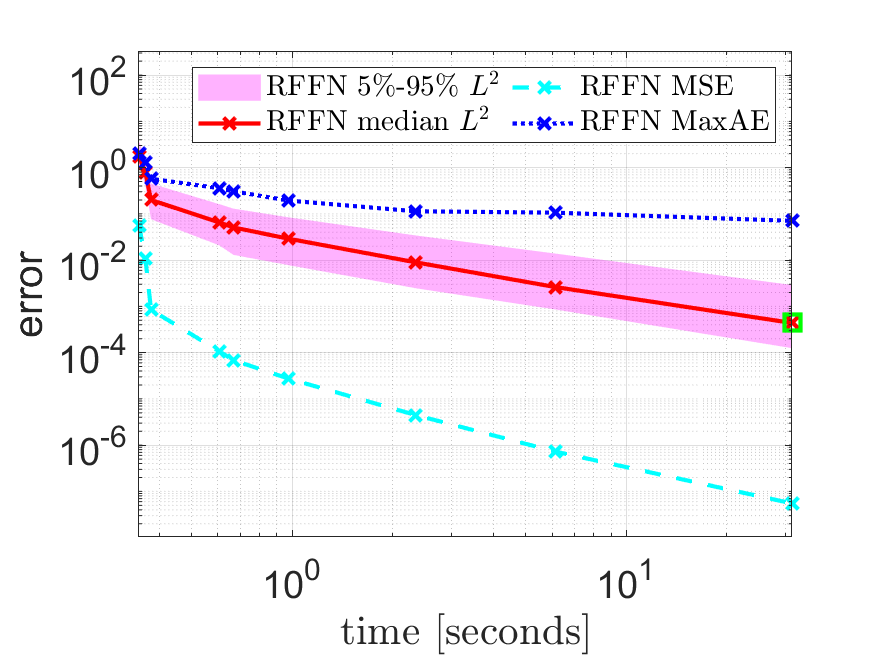}
    \label{fig:FHN_training_multi_time}
    }
    \caption{Training convergence of the RandONets for the FHN PDE in Eq.~\eqref{eq:FHN_PDE}, simulated with Lattice Boltzmann method~\cite{galaris2022numerical,theodoropoulos2000coarse}.
    (a)-(b) Multi paramater $\epsilon \in (0.005,0.995)$ with Homotopy-based RandONets.
    Convergence of the RandONets w.r.t (a) the number of neurons in the branch and (b) the computational time. We report the Maximum absolute error (MaxAE), the Mean Square Error (MSE) and the median of the $L^2$ error and the range between $5\%$ and $95\%$ percentiles.
    }
\end{figure*}
\section{Further Details on the FHN PDEs}
In this section, we describe data generation for training the POD-RandONet on the Fitzhugh-Nagumo equations using Lattice Boltzmann method.
\subsection{The D1Q3 Lattice Boltzmann Model}
\label{App:FHN_LBM}
We employ a Lattice Boltzmann Model (LBM) as a fine-scale simulator for the FitzHugh-Nagumo system. The mesoscopic system is described by the probability density function $ f(\vec{r},\vec{c},t) $ which evolves according to the Boltzmann transport equation
\begin{equation}
\frac{\partial f}{\partial t }+\vec{c}\cdot \nabla f=\mathcal{R}(f),
\end{equation}
where $ \mathcal{R}(f) $ represents the collision term. Using the BGK approximation,
\begin{equation}
\mathcal{R}(f)=\frac{1}{\tau}(f^{eq}-f),
\end{equation}
we obtain the Lattice Boltzmann method with discrete velocities $ c_i $:
\begin{equation}
f_i(\vec{r}+\vec{c}_i\Delta t,t+\Delta t)=f_i(\vec{r},t)+\frac{\Delta t}{\tau}(f_i^{eq}-f_i).
\end{equation}
We adopt a $ D1Q3 $ scheme that uses three velocities $ c_i $ per function: $ c_1 = \frac{\Delta x}{\Delta t} $, $ c_{-1} = -\frac{\Delta x}{\Delta t} $, and $ c_0 = 0 $. Each node in the lattice hosts six dummy particles: three for the activator $ u $ and three for the inhibitor $ v $. The macroscopic functions are recovered as
\begin{equation}
u(x_j,t_k)= \sum^1_{i=-1}f^u_i(x_j,t_k), \quad v(x_j,t_k)= \sum^1_{i=-1}f^v_i(x_j,t_k).
\end{equation}
The equilibrium distributions for the purely diffusive FHN system are:  
\begin{equation}
f^{u,eq}_i = \omega_i u, \quad f^{v,eq}_i = \omega_i v,
\end{equation}
where the weights are $ \omega_{\pm1} = \frac{1}{6} $, $ \omega_0 = \frac{4}{6} $. The time evolution of the system further incorporates reaction terms $ R^l_i $ via
\begin{equation}
f_i^l(x_{j+i},t_{k+1})=f_i^l(x_j,t_k)+\frac{\Delta t}{\tau^l}(f_i^{l,eq}-f_i^l) + \Delta t R^l_i,
\end{equation}
with $l \in \{u,v\}$. Here, the reaction terms are:
\begin{equation}
R^u_i = \omega_i (u - u^3 - v), \quad R^v_i = \omega_i \varepsilon (u - \alpha_1 v - \alpha_0).
\end{equation}
Finally, the relaxation parameter $\tau^l$ relates to the macroscopic diffusivity $ D^l $ as
\begin{equation}
\frac{\Delta t}{\tau^l} = \frac{2}{1 + 6 D^l \frac{\Delta t}{\Delta x^2}}.
\end{equation}

\paragraph{Data Generation}
In our simulations, we vary the bifurcation parameter $ \varepsilon $ between $0.005$ and $0.955$~\cite{theodoropoulos2000coarse}. We discretize the spatial domain using $ \Delta x = 0.2 $ (41 points) and the time domain with step size $ \Delta t = 0.01 $ over $ t \in [0, 450] $. The initial conditions for $ u $ and $ v $ are random elements from the function family  
\begin{equation}
\begin{aligned}
    &u_0(x) = w \tanh{\big(\alpha (x - c)\big)} + \beta, \quad v_0(x) = 0.12  u_0(x), \\
    & w \sim \mathcal{U}(0.8,1.2), \quad \alpha \sim \mathcal{U}(0.5,1), \\
    & c \sim \mathcal{U}(2,18), \quad \beta \sim \mathcal{U}(-0.4,0).
\end{aligned}
\end{equation}
In the above, $\mathcal{U}(a,b)$ denotes the uniform distribution in the interval $[a,b]$.

\subsection{Training Neural Operators: POD-RandONets}
\label{App:FHN_training}
We initially attempted to use vanilla RandONets without POD modes, but the results were not sufficiently accurate. The issue likely arises from the two coupled equations, which have solutions with different amplitude scales, leading to increased dimensionality. Additionally, the long timescale of the system (requiring $\sim 100$ seconds for one limit cycle) causes large error accumulation over many iterations. To address these challenges, we introduced POD modes, which improved both accuracy and stability by reducing dimensionality and filtering out high-frequency artifacts.
The data we collected from the LBM simulation are two functions $[u(x);v(t)]$, evaluated at $41$ grid points, and we record these solutions across the entire time horizon $[0.450]$. Let us denote the training data by $X=[U;V]\in \mathbb{R}^{82\times s}$, where $s$ is the number of data points.

Computing the POD modes remains efficient, even with approximately 400,000 function-data points, since it only requires the eigen-decomposition of the covariance matrix $C = X X^T$, which has dimension $82 \times 82$. We denote the eigen-decomposition of the covariance matrix by $C = V \Lambda V^{-1}$. From this, we retain the leading $k = 12$ components, capturing $99.9\%$ of the total POD energy. The projection matrix $V_k^T \in \mathbb{R}^{12 \times 82}$ is then used to obtain the reduced representation $\tilde{X} = V_k^T X$, which serves as input to the RandONet. We refer to such models as \emph{POD-RandONets}.

Afterward, we train homotopy-based RandONets and evaluate their performance. We only consider the multi-parameter case. In particular, the number of neurons in the trunk layer is $250$, and the number of neurons in the hidden branch layer varies in $[10, 20, 40, 80, 100, 150, 300, 500, 750, 1000]$. Our code investigates the convergence of the method with respect to these neuron counts in the branch net.

We repeat the training process over $10$ iterations and calculate the validation $L^\infty$-error to select the best RandONet among all iterations and different branch net sizes. The code supports GPU acceleration, which speeds up the training process, using \texttt{gpuArray} of the \texttt{parallel computing toolbox} of \texttt{Matlab}.

After training, we evaluate the model on the test dataset. We have selected the best-performing RandONets, highlighted by a green square point in the figures. In Supplementary Figure~\ref{fig:FHN_training_multi_neuron} we report the convergence error in terms of MaxAE, $L^2$ and MSE across the test dataset. We also report the corresponding execution times in the Supplementary Figure~\ref{fig:FHN_training_multi_time}.

\section{Pseudo-Arclength Continuation}
\label{App:numerical_analysis}
In this appendix, we describe the pseudo-arclength continuation algorithm for computing the bifurcation diagram of an operator as a function of the parameter $\lambda$. We focus on continuation past saddle-node bifurcations without going into details of bifurcation point detection, types of bifurcation points and branch switching. We refer to a number of published works~\cite{doedel2007lecture, doedel2012numerical, allgower2012numerical, fabiani2021numerical} for more details on these concepts. 

Consider a parameter-dependent function
\begin{equation}
    \psi(u;\lambda), \qquad \psi:\mathbb{R}^{n+1}\rightarrow\mathbb{R}^n
    \label{eq:ODEs}
\end{equation}
coming from the NO in Eq.~\eqref{eq:psi}, where $u\in\mathbb{R}^n$ is the $n$-dimensional state variable vector, and $\lambda \in \mathbb{R}$ is a scalar parameter. The goal is to construct a \emph{branch} $\Gamma$
\begin{equation} 
    \Gamma =\{(u;\lambda) \in \mathbb{R}^{n+1} \text{ such that } \psi(u; \lambda)=0\},
    \label{eq:equilibria}
\end{equation}
corresponding to the steady states of the system in Eq.~\eqref{eq:ODEs} for various values of the parameter $\lambda$.
The main concept underlying pseudo-arclength continuation~\cite{allgower2012numerical} is to follow the tangent vector $\tau$ of Eq.~\eqref{eq:equilibria} to construct a curve containing all steady-state solutions. This curve is parametrized using the (pseudo-)arclength $s$.  

A typical implementation involves a \emph{predictor-corrector} process. We start from a known point $(u(s), \lambda(s)) \in \Gamma$ and its tangent vector $\tau(s)$ to the curve, calculated as the unit vector in the null space of the extended matrix
\begin{equation} \label{eq:extended_jacobian}
\left[\nabla \psi(u(s);\lambda(s)); \frac{\partial \psi}{\partial \lambda}(u(s);\lambda(s))\right].
\end{equation}
We then compute a new point $(u(s+\Delta s);\lambda(s+\Delta s))$ in two steps: (a) \emph{predictor} compute an initial guess along the tangent $\tau(s)$; 
(b) \emph{corrector} iteratively refines the guess through Jacobian-Free Newton-Krylov to a new point on the curve $\Gamma$ in Eq.~\eqref{eq:equilibria}. Note that this new point is not always a distance $\Delta s$ from the starting point due to the nonlinear nature of the Newton method; this is why it is called \emph{pseudo}-arclength continuation.

This method fails at bifurcation points, that is, where the extended Jacobian in Eq.~\eqref{eq:extended_jacobian} becomes singular. However, it is possible to trace solution branches past saddle-node bifurcations because the extended Jacobian does not become singular at such points.
\end{document}